\documentclass[10pt,twocolumn,letterpaper]{article}

\usepackage{3dv}
\usepackage{times}
\usepackage{epsfig}
\usepackage{graphicx}
\usepackage{amsmath}
\usepackage{amssymb}
\usepackage{subfigure}
\usepackage{comment}
\usepackage{caption}
\usepackage{multirow}
\usepackage{float}

\threedvfinalcopy 

\def\threedvPaperID{****} 

\ifthreedvfinal\fi
\begin{document}
\title{SCFusion: Real-time Incremental Scene Reconstruction \\with Semantic Completion}

\author{Shun-Cheng Wu\textsuperscript{1}, Keisuke Tateno\textsuperscript{2}, Nassir Navab\textsuperscript{1}, Federico Tombari\textsuperscript{1,2}\\
\textsuperscript{1}Technische Universitat M\"{u}nchen, Germany \hspace{1.5cm} \textsuperscript{2}Google Inc., Switzerland\\
{\tt\small shuncheng.wu@tum.de, ktateno@google.com, nassir.navab@tum.de, tombari@google.com}
}

%
%
\maketitle
%
%
%
%
%
\begin{abstract}
Real-time scene reconstruction from depth data inevitably suffers from occlusion, thus leading to incomplete 3D models. 
Partial reconstructions, in turn, limit the performance of algorithms that leverage them for applications in the context of, e.g., augmented reality, robotic navigation, and 3D mapping. 
Most methods address this issue by predicting the missing geometry as an offline optimization, thus being incompatible with real-time applications. 
We propose a framework that ameliorates this issue by performing scene reconstruction and semantic scene completion jointly in an incremental and real-time manner, based on an input sequence of depth maps.
Our framework relies on a novel neural architecture designed to process occupancy maps and leverages voxel states to accurately and efficiently fuse semantic completion with the 3D global model.
We evaluate the proposed approach quantitatively and qualitatively, demonstrating that our method can obtain accurate 3D semantic scene completion in real-time.
\end{abstract}

\begin{figure}[th]
    \centering
    \includegraphics[width=0.95\columnwidth]{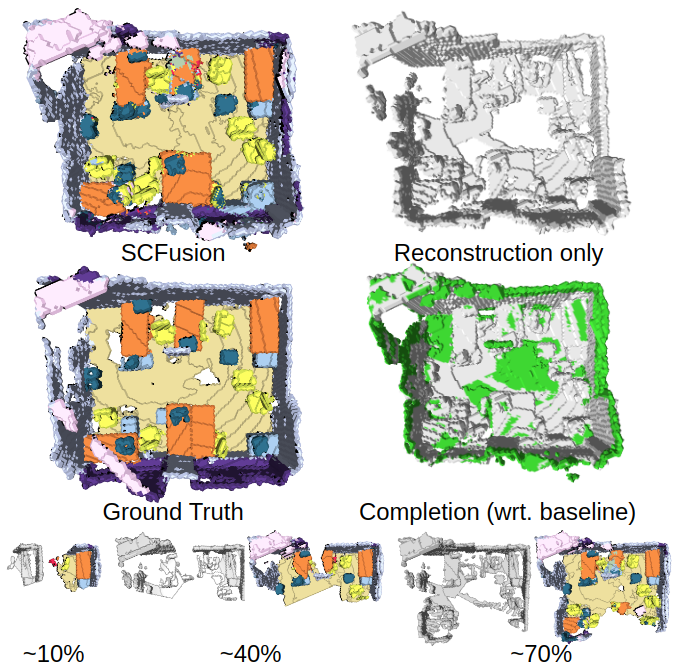}
    \caption{Incremental 3D reconstruction suffers from occlusions, leading to incomplete scenes. Unlike other approaches, which treat semantic scene completion as an offline optimization, SCFusion carries it out incrementally and in real-time along with scene reconstruction, yielding an accuracy comparable to state-of-the-art offline methods.}
    \label{fig:teaser}
\end{figure}

\section{Introduction}
The development of consumer depth cameras has fostered impressive advancements in research fields related to 3D geometry. Thanks to the availability of a stream of dense depth maps as input data, 3D computer vision algorithms are now capable of accurately localizing objects in the surrounding scene and reconstructing the 3D map of the environment. 
%
However, like any other camera, a depth camera is a viewpoint-dependent sensor which can estimate depth information only for the surface visible from the current vantage point. Since the foreground objects create occlusions for background objects and structure, this will result in missing depth information across multiple different objects. Hence, when depth maps are fused together via Simultaneous Localization and Mapping (SLAM) or Structure-from-Motion (SfM) \cite{newcombe2011kinectfusion, dai2017bundlefusion}, the obtained 3D reconstructions are geometrically incomplete. Importantly, the incompleteness of the reconstructed scene challenges, in turn, those methods that leverage scene reconstruction for tasks related to, \eg, augmented reality, robotic navigation and scene understanding, as they need to be robust to handle partial shapes.

Several methods have been proposed to recover the missing information in a scene from a given single depth image \cite{sscnet, yang20173d, wang2018adversarial, li2019rgbd, wang2019forknet} or a reconstructed 3D scan \cite{dai2018scancomplete, cherabier2018learning, sgnn}. These methods demonstrate the possibility of using partial observations to reasonably estimate the full geometry and even the semantic representations. 
%
Despite the important steps forward of research in scene completion, a gap remains to bring completion methods from either a single frame or the entire scan to real-time contexts. %
Existing single-frame methods \cite{sscnet,wang2019forknet,yang20173d} treat each input frame individually without exploiting the availability of additional viewpoints, this leading to inaccuracies in the extrapolated shapes, which rely mostly on priors learned by the network from the training set. On the other hand, approaches designed to process an entire scan are off-line methods which treat this task as a post-processing optimization, and consequently cannot be used in real-time applications \cite{sgnn, dai2018scancomplete}. In addition, the inconsistency of input and output formats limits the possibility of a direct integration within a unified model. Indeed, most methods take a (truncated) signed distance function (SDF) or one of its variations as input, while predicting occupancy probability \cite{sscnet, yang20173d, wang2019forknet} or an unsigned distance function (DF) \cite{dai2018scancomplete}. The only method \cite{sgnn} that has consistent input and output formats does not predict semantic information. All these difficulties hinder the use of these approaches for real-time applications.

We fill this gap by proposing a framework consisting of a novel network and fusion scheme that enables real-time scene reconstruction with incremental semantic scene completion, which we dub \emph{Scene Completion Fusion} (SCFusion). 
%
The two main pipelines that compose our proposed approach, \ie scene reconstruction and scene completion, are carried out in parallel over two threads. The reconstruction pipeline continuously fuses input depth maps into a global volumetric map, 
while the completion pipeline semantically completes the global map by regularizing its sub-maps via a fully-connected conditional random field (CRF).
%
The key component of our method is to use an occupancy representation over the entire system. The explicit voxel states and probability distribution on its estimation from the occupancy mapping \cite{moravec1985high, hornung2013octomap} provide more information than the use of TSDF fusion. 
We leverage this advantage and design a network which takes occupancy probability and voxels with unknown state as inputs to predict occupancy and semantic labels. The use of voxels with unknown state represents an additional input for the network, to guide it towards regions where completion might be necessary. The consistent input and output formats allow a direct integration between the predicted occupancy from the neural network and the global occupancy map built by the SLAM engine. 
%
In addition, we utilize the explicit known and unknown voxel state from occupancy mapping in our sub-map extraction and fusion process to define a set of integration policies which result in better performance in the task of semantic scene completion.  

Notably, the currently available datasets do not have a complete ground truth for scenes since the one they include is obtained from real scans of the environment \cite{dai2017scannet, wald2019rio, Matterport3D, nyudataset, InteriorNet} that inevitably suffer from camera occlusions, while what has been a standard synthetic benchmark (SunCG \cite{sscnet}) is no longer available.
Hence, to evaluate the performance of our method we have built a dataset with complete scenes using the alignments available in Scan2CAD \cite{scan2cad}, where the full 3D models from ShapeNet \cite{shapenet} are fused into the scenes of ScanNet \cite{dai2017scannet}. We refer to this new dataset as CompleteScanNet. 

Our contributions can be summarized as follows: 
(i) A framework for real-time incremental semantic scene completion, to the best of our knowledge the first of this kind. 
(ii) A novel neural architecture which leverages the voxel states in occupancy maps to predict scene completion and appropriately fuse it with the global 3D model, 
based on the idea of leveraging known/unknown states for 3D semantic scene completion.
(iii) A benchmark dataset based on ShapeNet \cite{shapenet} 3D models and ScanNet \cite{dai2017scannet} scenes that can be used to evaluate semantic scene completion algorithms based on RGB-D sequences. This fills an important gap given the absence of such benchmarks in literature.  


\begin{figure}[t]
    \centering
\includegraphics[width=0.9\linewidth]{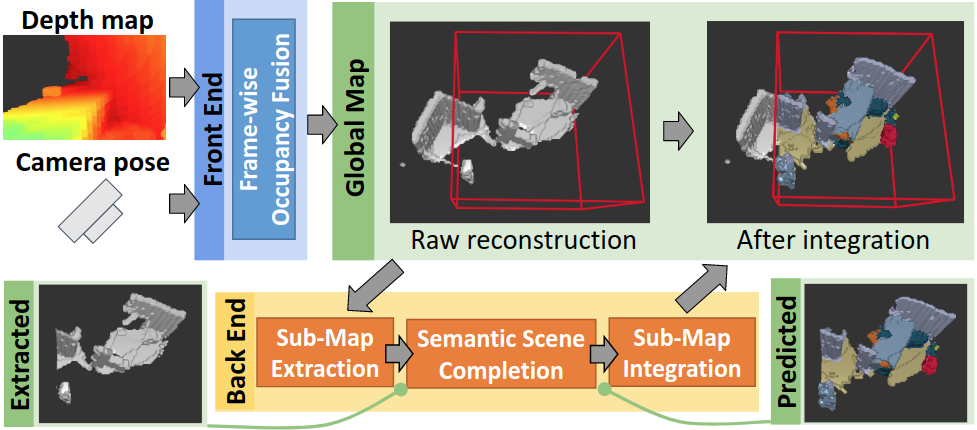}
    \caption{The overall pipeline of SCFusion for real-time scene reconstruction and semantic scene completion. The front-end reconstructs a 3D scene using occupancy maps. The back-end semantically completes the geometry and fuses it back to the global map.}
    \label{fig:pipeline}
\end{figure} 

\section{Related Work}


\subsection{Real-time dense 3D reconstruction}
Research activities in real-time dense 3D reconstruction can be subdivided between depth-based \cite{newcombe2011kinectfusion, keller2011pbf} and monocular \cite{newcombe2011dtam, schoeps20153dv,  yokozuka2019vitamine}. From the point of view of scene representation, the methods can be classified into three types: TSDF-based \cite{newcombe2011kinectfusion, voxelhashing, Kahler2015infinitam}, occupancy-based \cite{hornung2013octomap, moravec1985high} and surfel-based \cite{keller2011pbf, whelan2015elastic}. Given the scope of this paper, we will only review depth camera-based methods with TSDF and occupancy map representation.

For TSDF-based methods, Kinect Fusion \cite{newcombe2011kinectfusion} relies on a 3D volume which stores a Truncated Signed Distance Field (TSDF) value at each voxel. The input depth maps are back-projected into a 3D volume, which consists of voxel grids, and each depth observation is converted into a TSDF value and averaged together over the voxel grids. The surface is determined by finding the interpolated zero-crossing point via ray-casting. 
For occupancy-based methods, unlike TSDF which is a surface estimation approach, occupancy-based methods estimate occupancy in each voxel grid. This restricts the reconstructed resolution to the defined voxel size, and the reconstructed surface is not well-defined. Loop et al. \cite{loop2016closed} proposed to use a quadratic b-spline instead of a Gaussian noise model which covers the gap of occupancy reconstruction and allows occupancy mapping to have equivalent accuracy as the TSDF method. 
Vespa et al. \cite{supereight} propose to set the standard deviation value to be proportional to the measured depth distance to be more realistic to the triangulation-based depth camera noise model.


Since 3D volumetric representations are memory consuming and are difficult to extend to a large scale, Voxel Hashing \cite{voxelhashing} proposed a reconstruction framework based on a hash-grid 3D representation. Allocating small voxel blocks around observed surfaces instead of doing it on the entire space to reduce memory consumption and use memory more efficiently. \cite{hornung2013octomap} uses an octree to efficiently divide space into cubes of different size. 


\subsection{3D shape completion}
The completion of 3D shapes starts from an input partial 3D shape to provide an occlusion-free and complete 3D shape. Existing methods can be divided into two categories: object completion and scene completion.

\textbf{Object completion} 
focuses on a single object which includes recovering local surface primitives by using a continuous energy minimization \cite{sorkine2004least, nealen2006laplacian, zhao2007robust}, completing shapes by leveraging symmetric information or prior information from a database \cite{thrun2005shape, mitra2006partial, speciale2016symmetry} and replacing partial shape with an aligned CAD model retrieved from a database \cite{avetisyan2019scan2cad, kim2012acquiring, nan2012search}. However, to apply single object shape completion to scenes, additional object detection and segmentation are required. The completion quality thus additionally relies on how well the detection and segmentation methods perform.

\textbf{Scene completion} 
focuses on completing the entire scene with or without predicting the semantics. This can be done by using purely geometric approaches \cite{firman2016structured, sgnn} or by also considering semantic information. Recent trends target joint prediction of semantics and geometry by leveraging deep learning \cite{sscnet, guo2018VVN, wang2018adversarial}. Most recent work focus on single viewpoint semantic completion, either based on a single depth image \cite{sscnet, wang2019forknet} or with RGB information \cite{guo2018VVN}. Only a few works target completion of an entire scan \cite{dai2018scancomplete}. 
Although these methods have shown promising results, they target single frame prediction and do not handle sequences of depth maps or temporal information.





\subsection{Semantic 3D reconstruction}
This line of work focuses on joint optimization for 3D reconstruction and semantic segmentation. Initially, \cite{ladicky2012joint} proposed to jointly optimize semantic segmentation and stereo matching by using a random field. Later works tried to solve this problem by using a conditional random field on either single view \cite{kim20133dcrf} or multi-view \cite{hane2013joint, hane2016dense}. This idea was also extended to object-centric reconstruction \cite{karimi2015segment, hane2014class}. 
A common drawback of the above approaches is that they are not able to capture complex relationships in 3D space. Recently  \cite{cherabier2018learning} suggested an end-to-end trainable way to capture more complex information between the semantic labels and the 3D geometry. 
However, these methods are computationally expensive and require extensive use of memory. Finally, a different corpus of work focuses on incremental construction of a semantic 3D map by fusing 2D semantic predictions from images \cite{mccormac2017semanticfusion, panopticfusion}.


\section{Proposed incremental semantic scene sompletion framework}
In this section, we propose our framework for real-time incremental semantic scene completion. The flow diagram sketching the algorithm pipeline deployed at each input frame is shown in figure \ref{fig:pipeline}. 
The stages regarding front-end scene reconstruction are shown as blue boxes there, while those regarding back-end semantic scene completion are depicted with orange boxes. The outputs from front- and back-end operations are marked as green boxes.
We assume a stream of depth maps acquired from a moving RGB-D sensor, and the corresponding poses for each depth image acquired from an external visual-inertial odometry (VIO) sensor or any pose estimation methods. 
Our front-end pipeline fuses each paired input pose and depth map into a gravity aligned dense volumetric occupancy map (section \ref{sec:src_pipeline}). The back-end pipeline predicts the occupancy and semantic label incrementally with our sub-maps strategy along with online map regularization (section \ref{sec:sscc_pipeline}).

%
%

\subsection{Frame-wise occupancy map fusion}  \label{sec:src_pipeline}
This section outlines our volumetric occupancy mapping method used in the front-end scene reconstruction pipeline. 
Our reconstruction system uses occupancy mapping instead of TSDF fusion to provide a consistent format with the prediction from our network. In addition, occupancy mapping provides three explicit states at each voxel: occupied, empty and unknown \cite{hornung2013octomap,loop2016closed}, that can provide additional information to our network architecture as well as to our fusion scheme. 
As proposed in \cite{voxelhashing}, we use a hashed voxel map representation which stores voxel cubes only when a depth measurement is present. Each voxel cube consists of the same number of voxels storing fused depth values from input measurements and, in our implementation, the predicted label from the back-end pipeline. For every input pair of a depth map and its pose, valid depth measurements are projected and fused into the global map. Each depth measurement is converted to the occupancy probability using a log-odd representation and the noise model proposed in \cite{supereight}. 


Given a sensor measurement \(z_{t}\) at time \(t\), each voxel in the global volumetric map stores a fused occupancy probability \(P(\mathbf{v}|z_{1:t}) \in \mathbb{R}\) at each voxel location \(\mathbf{v} \in \mathbb{R}^3\). 
During occupancy fusion, the occupancy probability \(P(\mathbf{v}|z_{1:t})\), \ie the probability of a voxel being occupied or empty given the sensor measurements \(z_t\), is estimated according to
\begin{equation} \label{eq:ofu_probability}
    P(\mathbf{v}|z_{1:t}) = \frac{P(\mathbf{v}|z_t)}{1-P(\mathbf{v}|z_t)} \frac{P(\mathbf{v}|z_{1:t-1})}{1-P(\mathbf{v}|z_{1:t-1})} \frac{1-P(\mathbf{v})}{P(\mathbf{v})}
\end{equation}
based on the measured occupancy probability \(P(\mathbf{v}|z_t)\) at time \(t\), the previous probability \(P(\mathbf{v}|z_{1:t-1})\) and a prior probability \(P(\mathbf{v})\). The measured occupancy probability \(P(\mathbf{v}|z_t)\) can be defined based on the given sensor model. Here, we use a quadratic B-spline sensor noise model as proposed in Vespa et al. \cite{supereight} to compute \(P(\mathbf{v}|z_t)\).  

Then, a projected depth measurement corresponding to voxel location \(\mathbf{v}\) in the global map coordinate system is computed as 
\begin{equation} 
 z_t(\mathbf{v}) = T_t^{-1} K^{-1} \dot{\mathbf{u}} D_t(\mathbf{u}),
\end{equation}
\begin{equation} 
\mathbf{u} = \pi(K T_t \mathbf{v}) \in \mathbb{R}^2
\end{equation}
where \(D_t(\cdot)\) is the input depth image at time \(t\), \(T_t = [R_t, t_t] \in \mathbb{SE}(3)\) is the current camera pose, composed of a 3\(\times\)3 rotation matrix \(R_t \in \mathbb{SO}(3)\) and a 3D translation vector \(t_t \in \mathbb{R}^3 \), and \(\mathbf{u}\) is the pixel location in the input depth image projected from voxel location \(\mathbf{v}\), with its homogeneous representation \(\dot{\mathbf{u}}\).

As in Loop et al. \cite{loop2016closed}, by using the log-odds notation and the common assumption of an uniform prior probability \(P(\mathbf{v}) = \frac{1}{2}\), equation (\ref{eq:ofu_probability}) can be re-written as
\begin{equation} \label{eq:ofu_logodd}
    l(\mathbf{v}|z_{1:t}) = l(\mathbf{v}|z_{1:t-1}) + l(\mathbf{v}|z_{t})
\end{equation}
with \(l(\mathbf{v}) = log\left(\frac{P(\mathbf{v})}{1-P(\mathbf{v}) }\right )\)

%
In addition to the occupancy probability, in each voxel we stores a semantic label \(L_t(\mathbf{v}) \in L\) where \(L\) denotes all class labels, its confidence \(W^{L}_{t}(\mathbf{v}) \in \mathbb{R}\) and a time stamp \(T_{t}(\mathbf{v})\). The time $t$ in our system represents the number of observations being fused to the global map.

\subsection{Incremental semantic scene completion}  \label{sec:sscc_pipeline}
In this section, we describe the stages of the proposed pipeline carrying out incremental sub-map semantic scene completion, which represents the core of our proposal.
Given the incremental fashion of scene reconstruction, we propose to also perform scene completion incrementally during reconstruction. 
Our approach uses the view frustum from the given pose to search for candidate regions, referred to as sub-maps, which will be used to extract the input for the proposed network, as described in section \ref{sec:ssc_extraction}.
Then, the input scenes are semantically completed by our network (see section \ref{sec:ssc_prediction}).

The use of sub-maps carries out completion at a local level and, thanks to its modularity, has the benefit of scaling up to large scenes. Nevertheless, 
by discretizing the scene geometry into sub-maps, this might create ambiguities on overlapping parts across neighboring sub-maps, \eg in case of different predictions. Hence, we propose a novel fusion scheme (see section \ref{sec:ssc_integration}) with a map regularization method (see section \ref{sec:ssc_regularization}) to handle this issue, aimed at fusing both temporal and spatial information. 
%
For the sake of efficiency and thread parallelization, all operations are performed in a separate GPU stream and CPU thread which 
are isolated from the main front-end pipeline.

\subsubsection{Sub-map extraction} \label{sec:ssc_extraction}
A sub-map is defined as an axis-aligned bounding box in the global map coordinate system, consisting of an anchor point and its enclosing bounding box.
The anchor point is calculated based on the view frustum. In our implementation the bounding box has a fixed size of \(64\times 64\times 64\). The view frustum is computed from the camera pose and its intrinsic parameters, with a distance range of [0.01m, 5m].
Given a view frustum, a minimum number of sub-maps are selected to cover the view frustum region.
Specifically, by defining the y-axis as the elevation axis, the first sub-map is selected as the view frustum region with minimum x and z values. Then adjacent sub-maps are iteratively added until the view frustum is fully covered.

%
%
Next, we want to discard the extracted sub-maps whose voxels did not undergo major changes based on the last depth map update. This is important for efficiency reasons, since we want to avoid to run completion on regions that were not modified in time. For this we adopt a specific criterion: a sub-map is discarded 
if the percentage of the out-dated voxels within its bounding box is above a threshold \(\tau_{s}\).
A voxel is considered \emph{outdated} when the difference between the voxel time stamp \(T_{t}(v)\) and the current time t is larger than \(\tau_{t}\). In turn, the time stamp of a voxel is updated when a predicted information is fused with this voxel, being set to 0. 
With that, the system is able to prevent over completion on the same local map and also to increase the efficiency of the system. In our implementation, we set \(\tau_{s}\) to 0.3 and \(\tau_{t}\) to 30.

\subsubsection{Semantic scene completion} \label{sec:ssc_prediction}
A key characteristic of the proposed network is that it takes, as input, occupancy probabilities and a binary mask to predict a semantically completed scene. 
Similarly to \cite{sscnet}, a voxel is consider occupied if it is assigned a non-empty label. We treat this state as an occupied observation in the map, which enables a direct integration of the associated voxel in the global map.
The use of a binary mask is inspired by the research field in image inpainting, where it was shown that using a mask to indicate missing regions improves the prediction accuracy from a neural network \cite{kohler2014mask}. Also, preventing mask vanishing during the recursive convolutional operations can further improve network performance \cite{pconv2018ECCV, FreeGated2019ICCV}. Inspired by this, we treat semantic scene completion as a 3D inpainting task, and propose a network architecture that leverages the benefits of occupancy maps and combines them with 3D inpainting. 

Our network is built on the semantic prediction branch of ForkNet \cite{wang2019forknet} with some modifications. 
First, the network takes a voxel grid of normalized occupancy probabilities and a voxel grid of binary masks as input. A binary mask is generated using the unknown state encoded with occupancy mapping. 
Second, we replace all convolutional layers with gated convolutional layers to prevent mask vanishing, as shown in  \cite{FreeGated2019ICCV}. 
Third, instance normalization is applied after each layer except for the final one. 
Last, a discriminator with spectral normalization is added during training \cite{miyato2018spectral}. 
For more details on the network architecture we kindly refer to the supplementary materials.

\subsubsection{Sub-map integration}  \label{sec:ssc_integration}
We leverage the three explicit states in occupancy mapping, i.e. \emph{occupied}, \emph{empty} and \emph{unknown}, to design a fusion policy that carefully fuses the predicted results into the global map. Given a voxel prediction from the network, we define the following rules. 
First, the prediction is discarded if classified as empty or if its corresponding voxel in the global map is in the empty state.
Second, the predicted semantic label is instead fused in the global map if the corresponding voxel is in either the unknown or occupied state. 
Analogously for the completion part, a voxel predicted as occupied by the network is fused only if the corresponding voxel in the global map is in the unknown state.
Based on these rules, our method is able to add semantic and geometric information to the scene while being guaranteed to maintain the same reconstruction accuracy for the visible surface as the mapping approach that we employ as backbone, \ie \cite{supereight}.
%
Remarkably, when fusing occupied voxels into the global map, we consider them as low confidence observations, with an assigned probability of 0.51, and follow the same approach described in section \ref{sec:src_pipeline} to fuse the predictions. 
This has the benefit of improving wrongly predicted geometry coming from an individual observation (\ie, a depth map).
%
%
%
%

As for merging labels, unlike occupancy estimation which deals with a continuous space, label prediction is categorical, hence requires a different integration strategy able to handle probability distributions. A simple solution is to save the entire probability distribution, however this would be impractical with a large number of labels, since the memory footprint in this case would grow linearly with the number of labels. Instead, we follow the approach in \cite{tateno2015real}, which stores a single label and a confidence value per voxel. Unlike their method, which applies decrements and increments of the confidence depending on the number of similar observations, we propose to integrate the softmax value of the predicted label as representative of the label confidence. 
Specifically, for a given voxel \(\mathbf{v}\) and its label \(l_{t}(v)\) is predicted with a confidence value \(w_{t}(v)\). If the confidence of the voxel is higher than the predicted confidence, we keep the label of the voxel from the previous state \(L_{t}(v) = L_{t-1}(v)\), and the confidence weight of the voxel is updated as: 
\begin{equation}
   W^{L}_{t}(v) = 
\begin{cases}
  W^{L}_{t-1}(v) + w_{t}(v), & \text{if}\ \left(L_{t-1}(v) = l_{t}(v)\right) \\
  W^{L}_{t-1}(v) - w_{t}(v), & \text{otherwise}
\end{cases}
\end{equation}
If \(l_{t}(v)\) is different from the voxel label and the confidence of the voxel is below the predicted confidence, we replace the voxel label and reduce its weight as:
\begin{equation}
L_{t}(v) = l_{t}(v), W^{L}_{t}(v) = w_{t}(v) - W^{L}_{t-1}(v)
\end{equation}
The label weight value is clamped with a maximum label confidence \(W^{L}_{max}\).

\begin{table*}[t]
\resizebox{\textwidth}{!}{%
    \begin{tabular}{c|c|c|c|c|c|c|c|c|c|c|c|c|c}
    Metric & Method & Ceiling & Floor & Wall & Window & Chair & Bed & Sofa & Table & TV & Furni & Object & Mean \\
    \hline\hline
\multirow{2}{*}{IoU}&ForkNet& \textbf{0.360} &          0.500 &          0.272 & \textbf{0.255} &          0.181 &          0.148 & \textbf{0.335} &          0.244 &          0.508 &          0.118 &          0.085 &          0.273\\ \cline{2-14} 
                    &Ours   &          0.197 & \textbf{0.541} & \textbf{0.379} &          0.108 & \textbf{0.310} & \textbf{0.194} &          0.266 & \textbf{0.322} & \textbf{0.659} & \textbf{0.219} & \textbf{0.148} & \textbf{0.304}\\
    \hline\hline
\multirow{2}{*}{Precision}&ForkNet& \textbf{0.735} & \textbf{0.739} &          0.466 & \textbf{0.460} &          0.421 &          0.278 & \textbf{0.452} &          0.375 &          0.554 &          0.245 &          0.244 &          0.452\\ \cline{2-14} 
                          &Ours   &          0.432 &          0.680 & \textbf{0.591} &          0.248 & \textbf{0.528} & \textbf{0.304} &          0.378 & \textbf{0.505} & \textbf{0.771} & \textbf{0.410} & \textbf{0.300} & \textbf{0.468}\\ 
  \hline\hline
\multirow{2}{*}{Recall}&ForkNet& \textbf{0.463} &          0.611 &          0.414 & \textbf{0.564} &          0.339 & \textbf{0.697} & \textbf{0.772} &          0.494 & \textbf{0.913} &          0.338 &          0.212 & \textbf{0.529}\\ \cline{2-14} 
                       &Ours   &          0.343 & \textbf{0.722} & \textbf{0.517} &          0.391 & \textbf{0.450} &          0.667 &          0.714 & \textbf{0.497} &          0.842 & \textbf{0.347} & \textbf{0.253} &          0.522\\ 
   \hline\hline
    \end{tabular}
}
\caption{Comparison between ForkNet \cite{wang2019forknet} and SCFusion on the test set of CompleteScanNet. Our method outperforms ForkNet in IoU and precision, while reporting slightly lower recall.}
\label{tab:quantitative_networks}
\end{table*}

\begin{figure*}[t]
\centering
\begin{minipage}[b]{.18\textwidth}
  \centering
  Input
  \includegraphics[width=\textwidth, height=0.08\textheight]{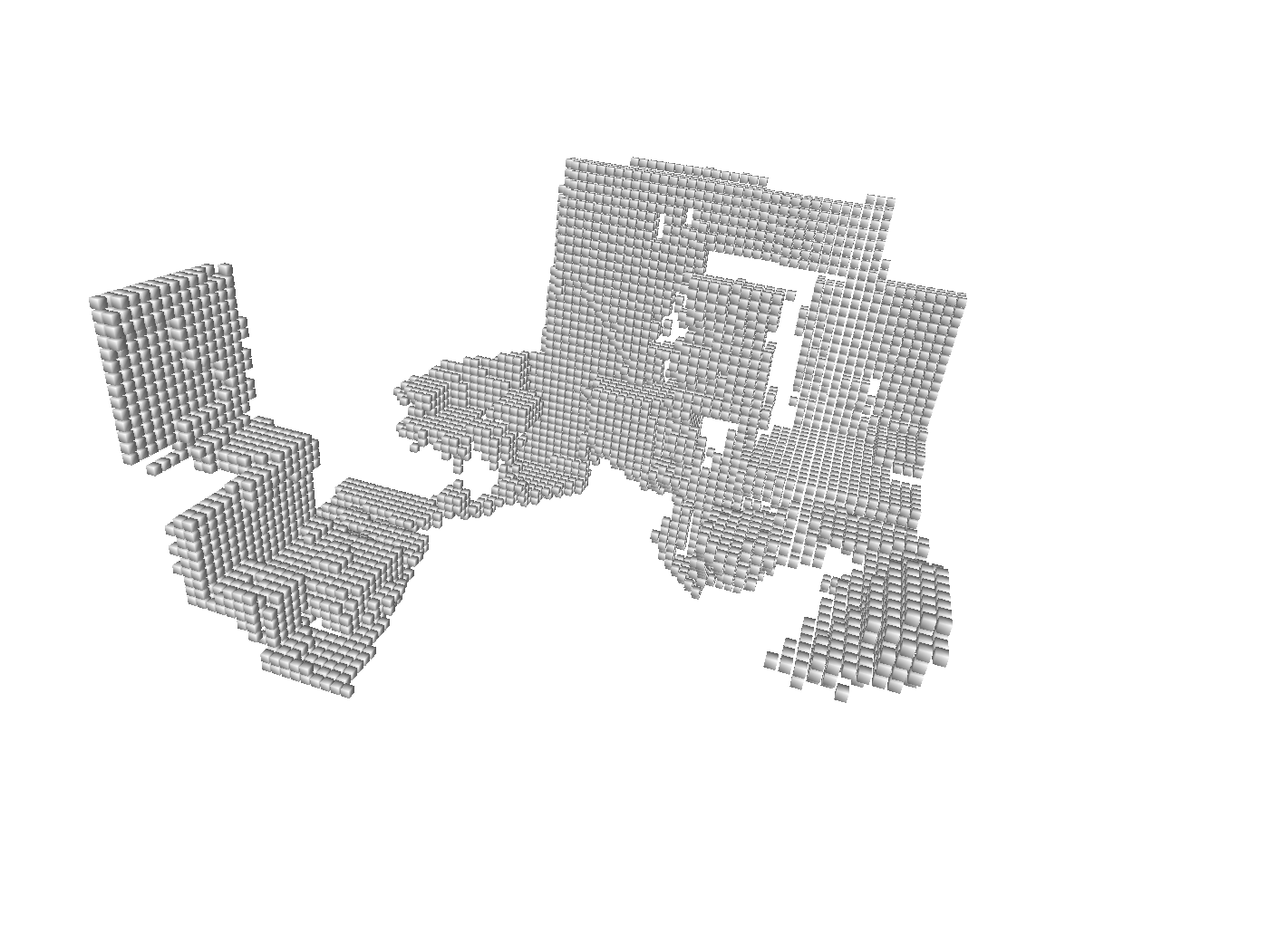}
  \includegraphics[width=\textwidth, height=0.08\textheight]{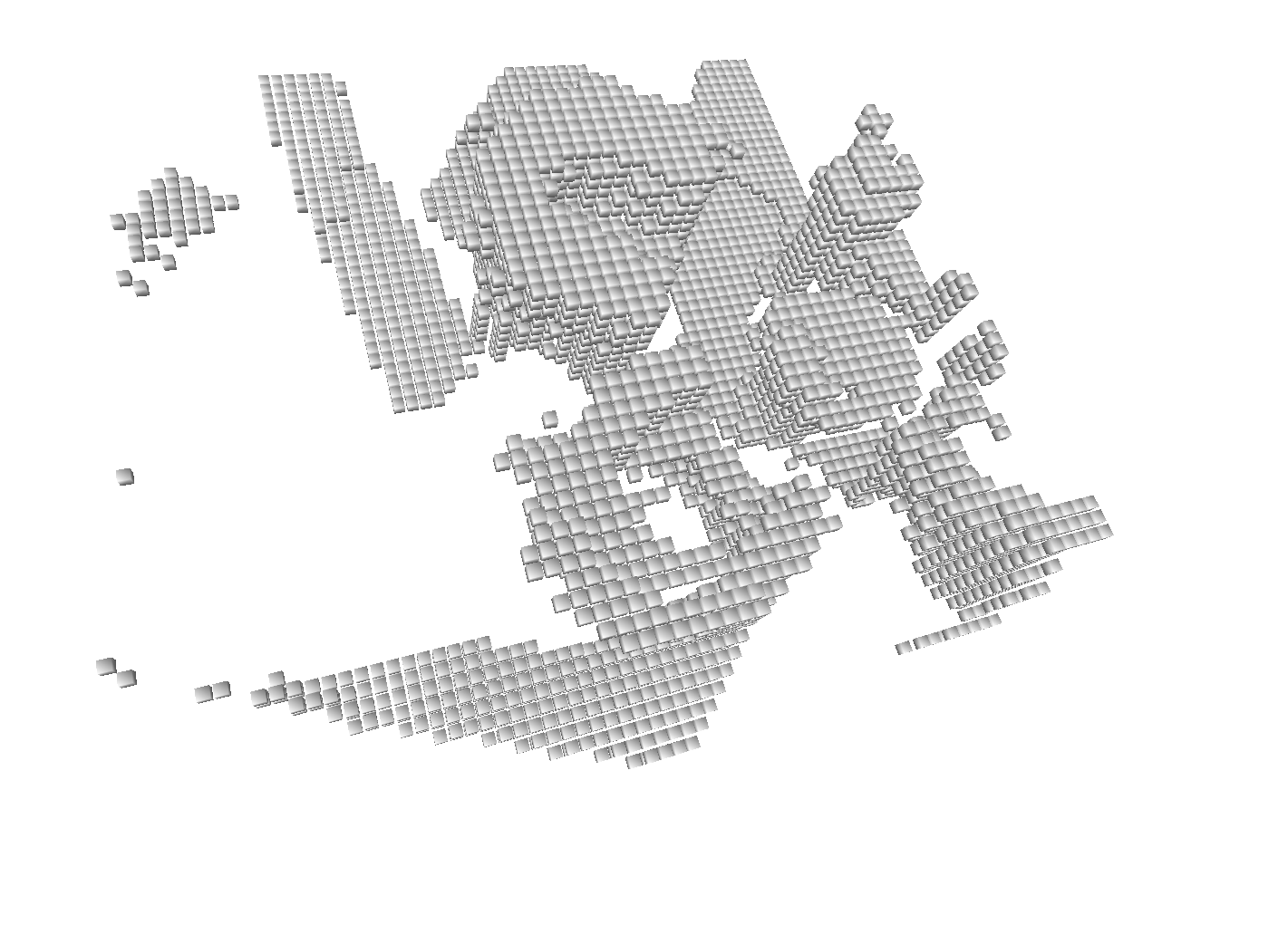}
  \includegraphics[width=\textwidth, height=0.08\textheight]{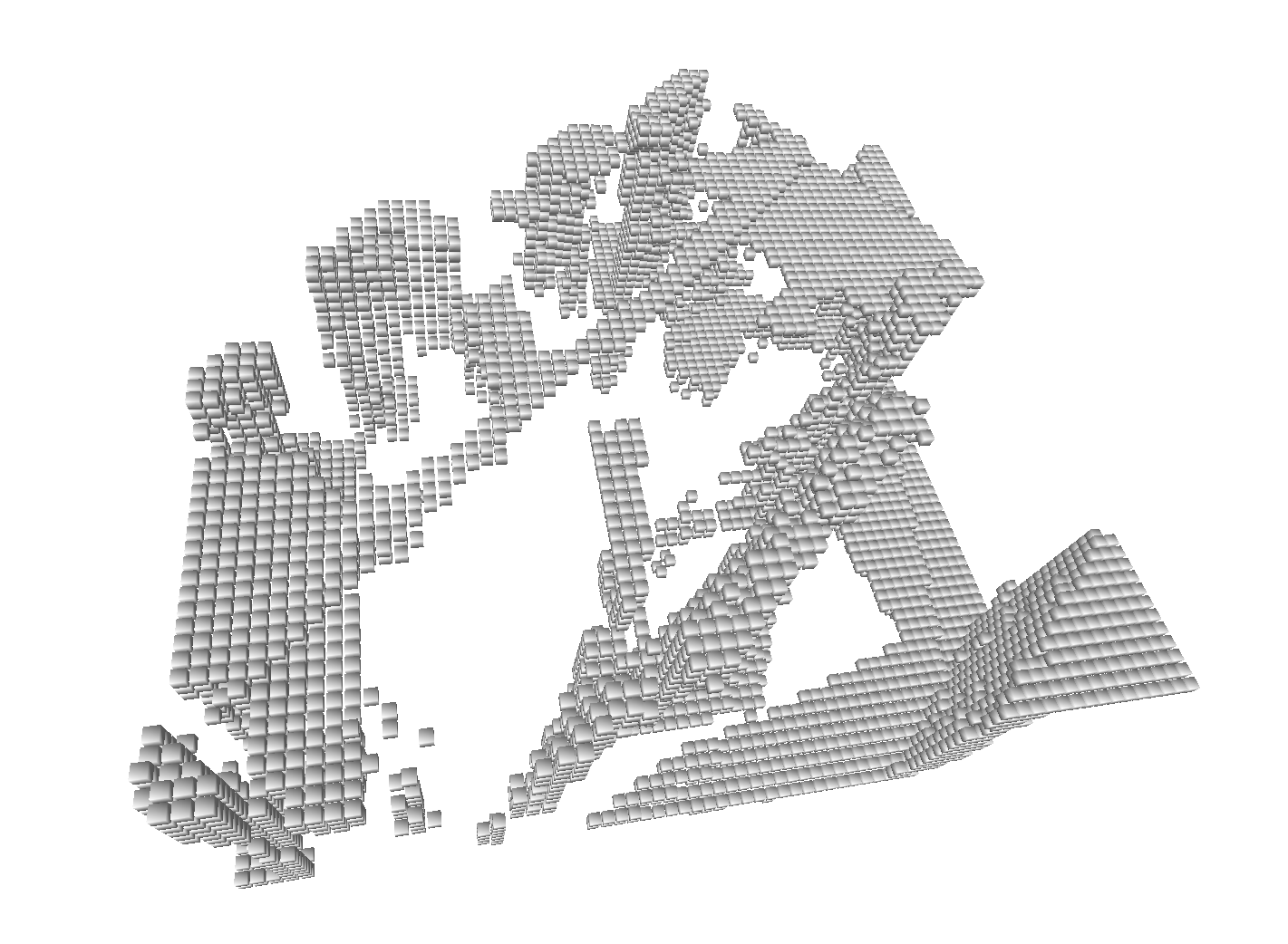}
\end{minipage}%
\begin{minipage}[b]{.18\textwidth}
  \centering
  ForkNet \cite{wang2019forknet}
  \includegraphics[width=\textwidth, height=0.08\textheight]{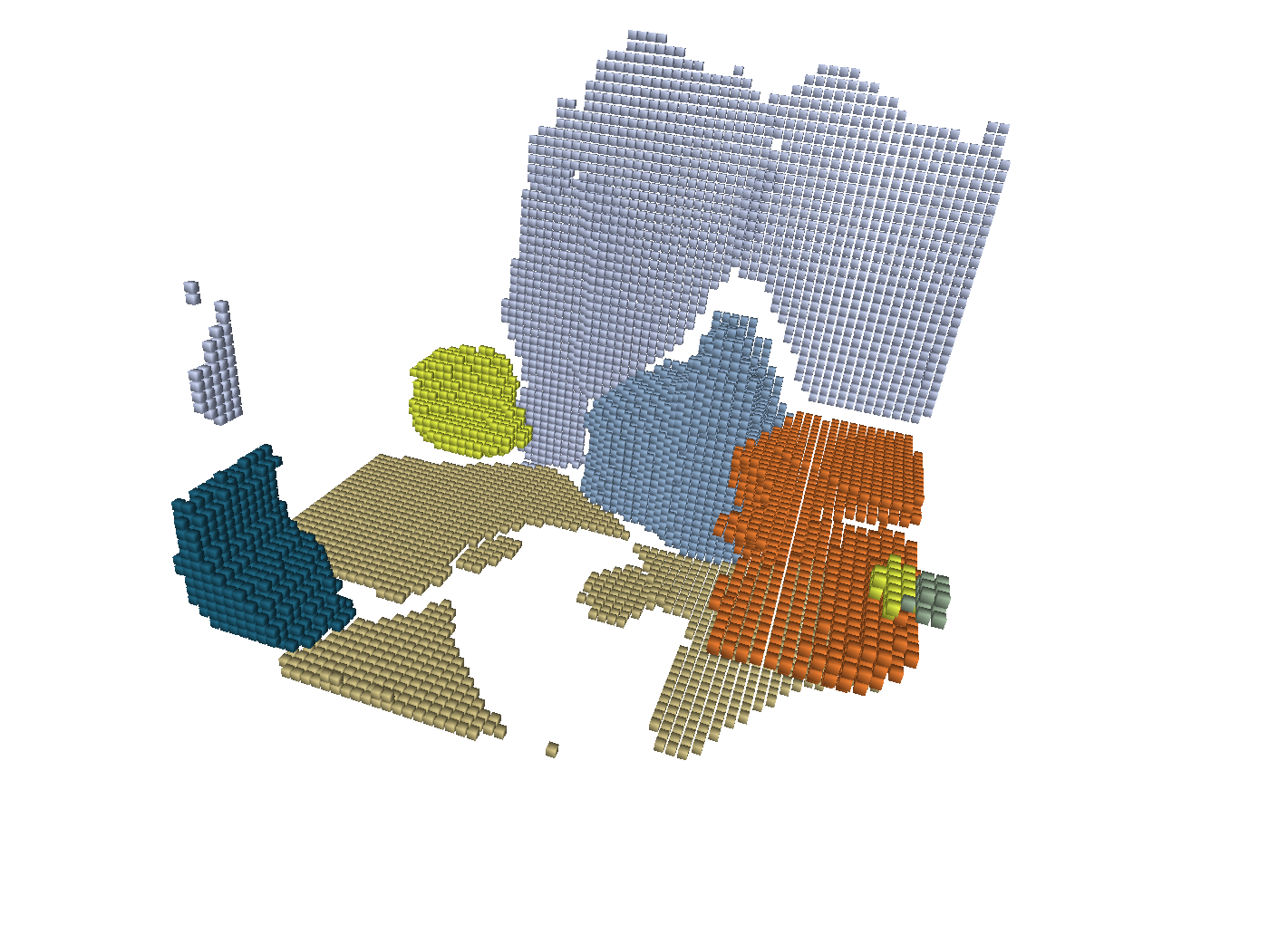}
  \includegraphics[width=\textwidth, height=0.08\textheight]{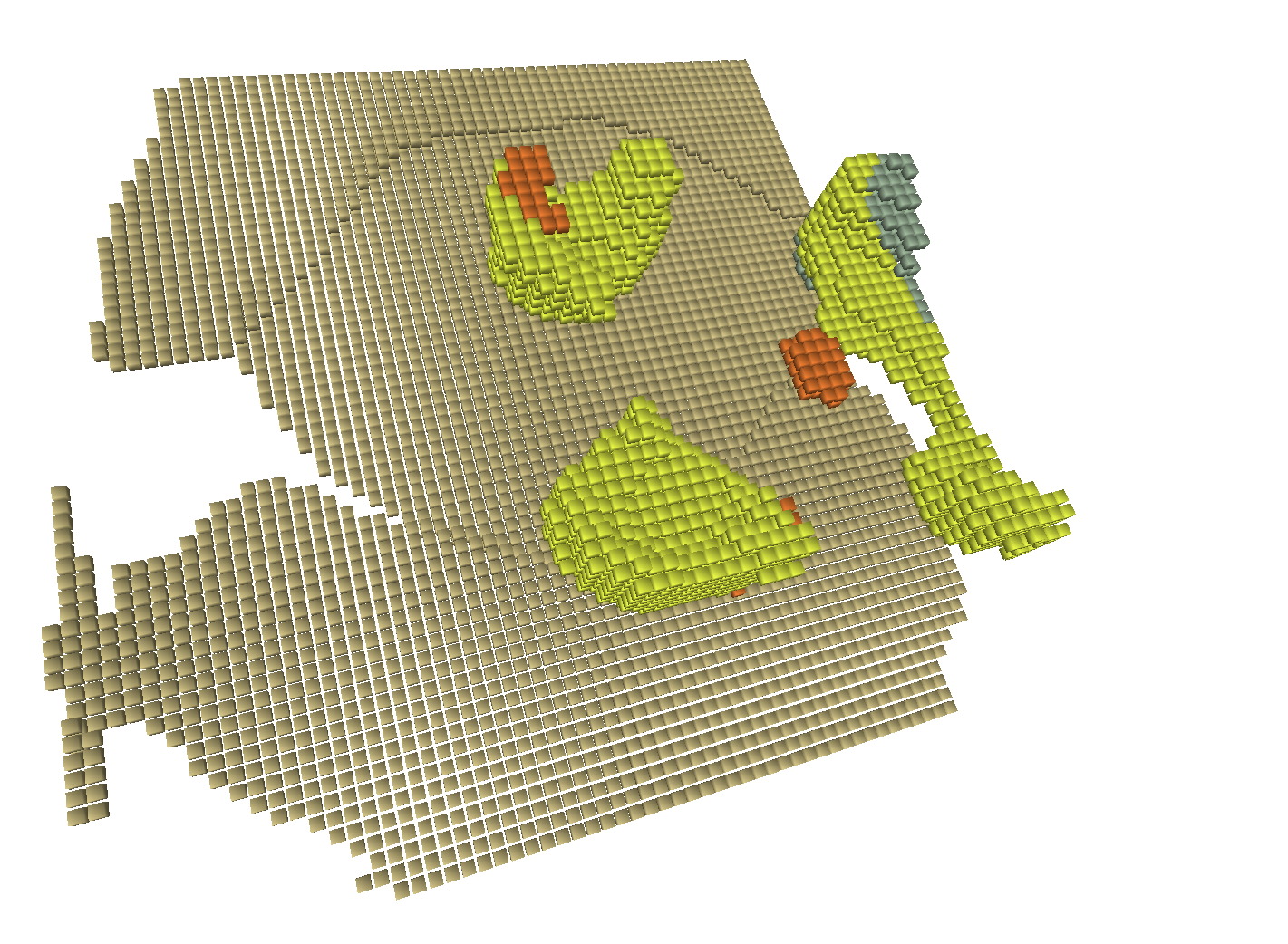}
  \includegraphics[width=\textwidth, height=0.08\textheight]{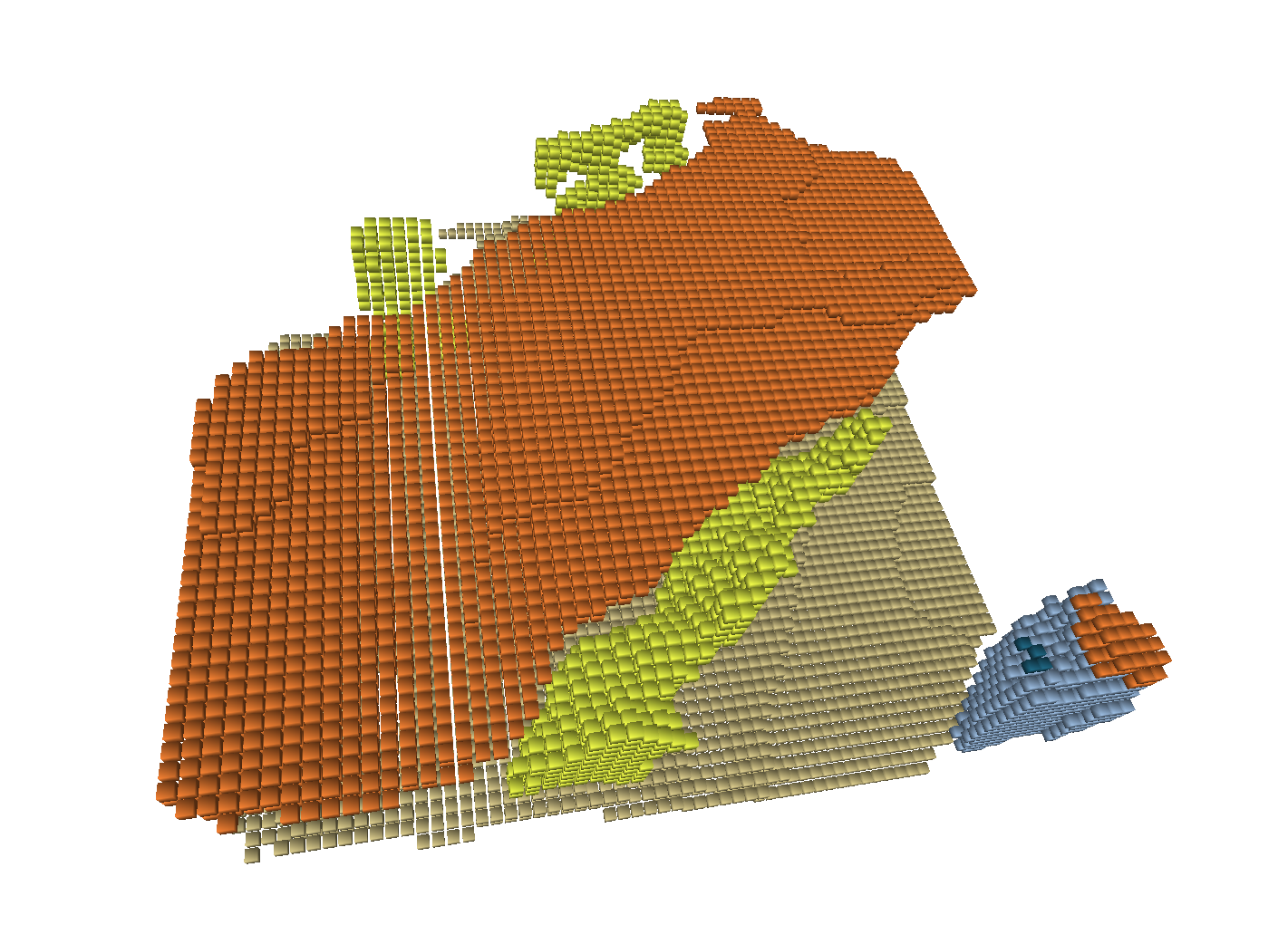}
\end{minipage}%
\begin{minipage}[b]{.18\textwidth}
  \centering
  Ours
  \includegraphics[width=\textwidth, height=0.08\textheight]{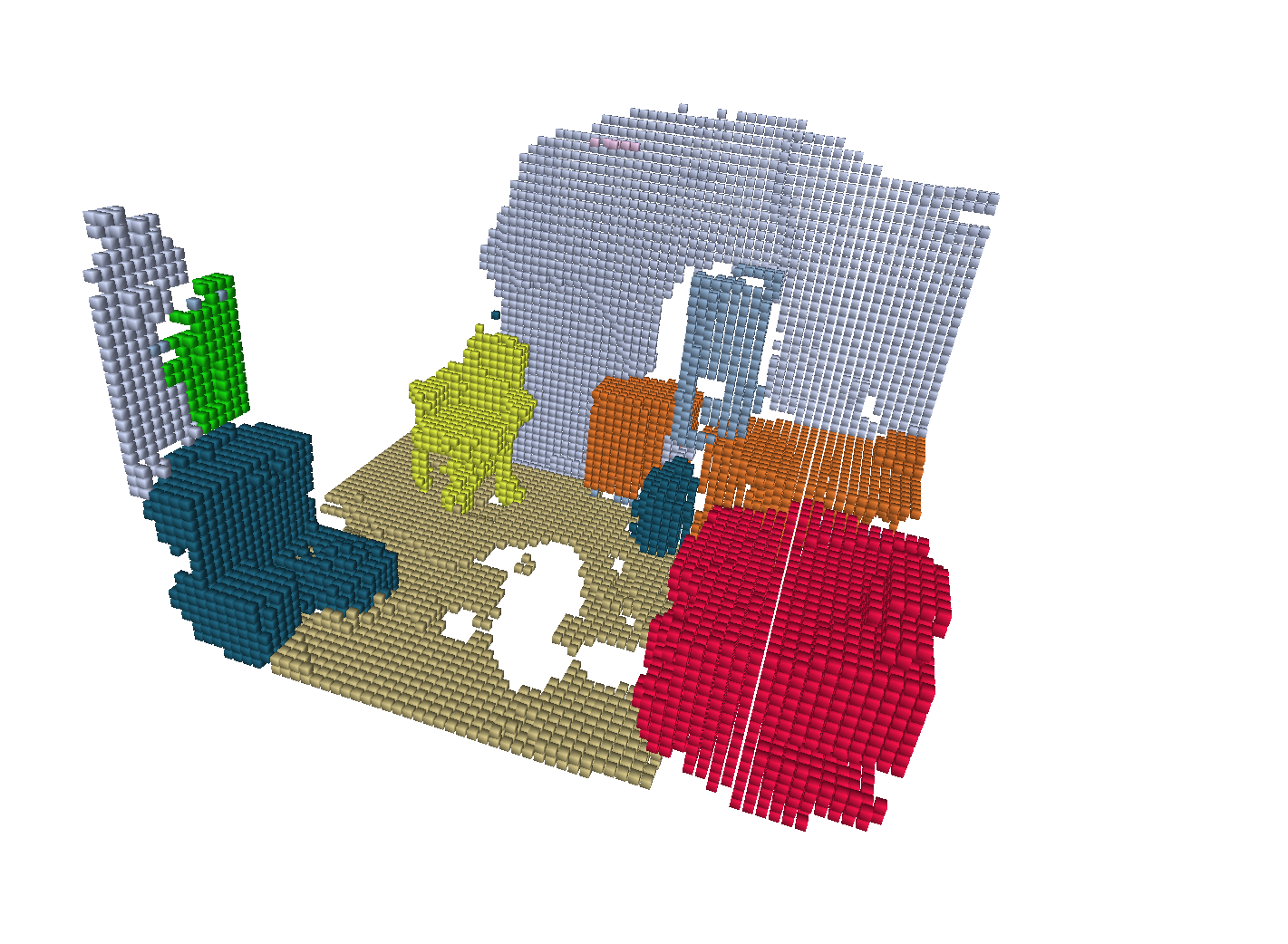}
  \includegraphics[width=\textwidth, height=0.08\textheight]{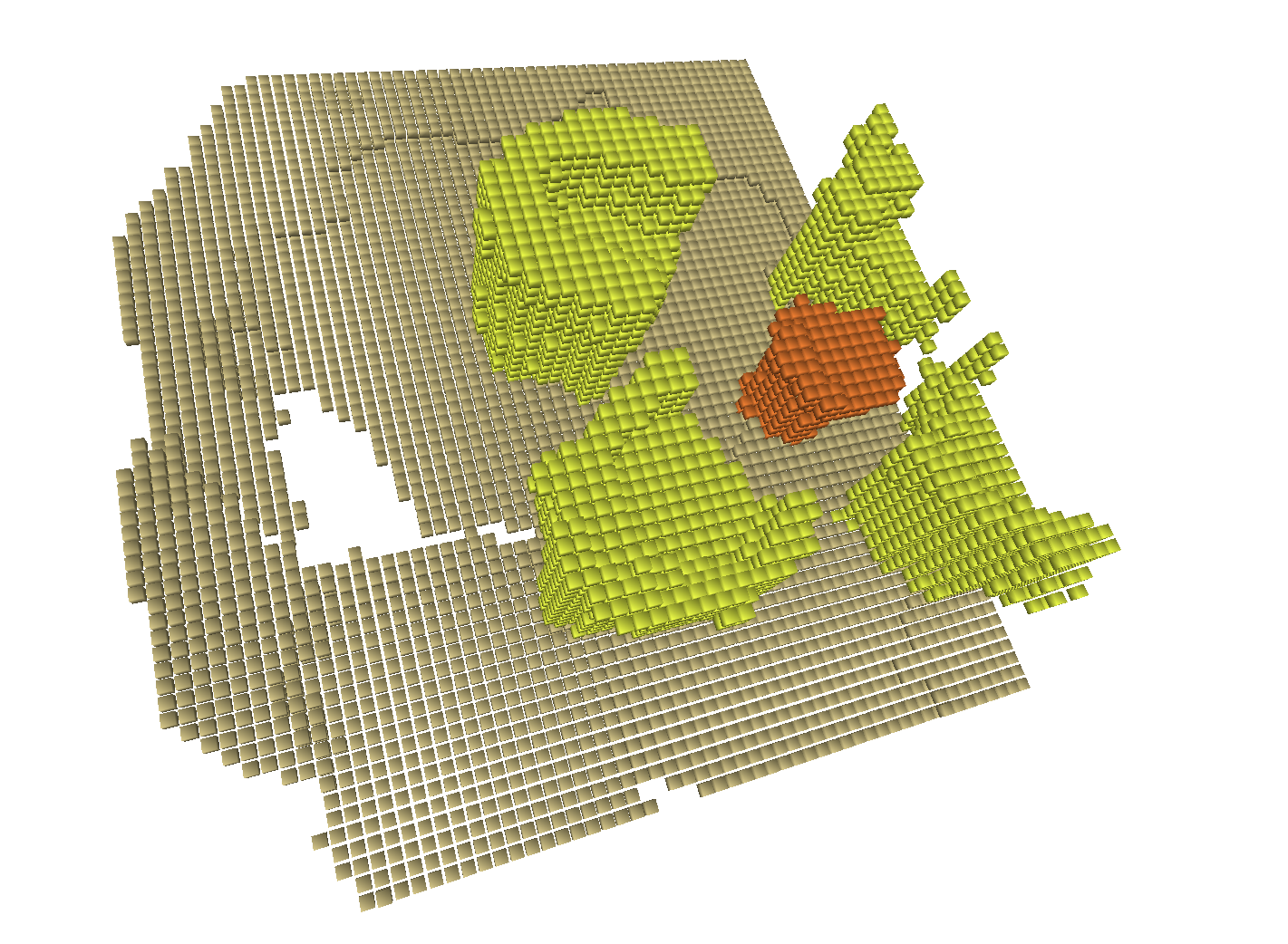}
  \includegraphics[width=\textwidth, height=0.08\textheight]{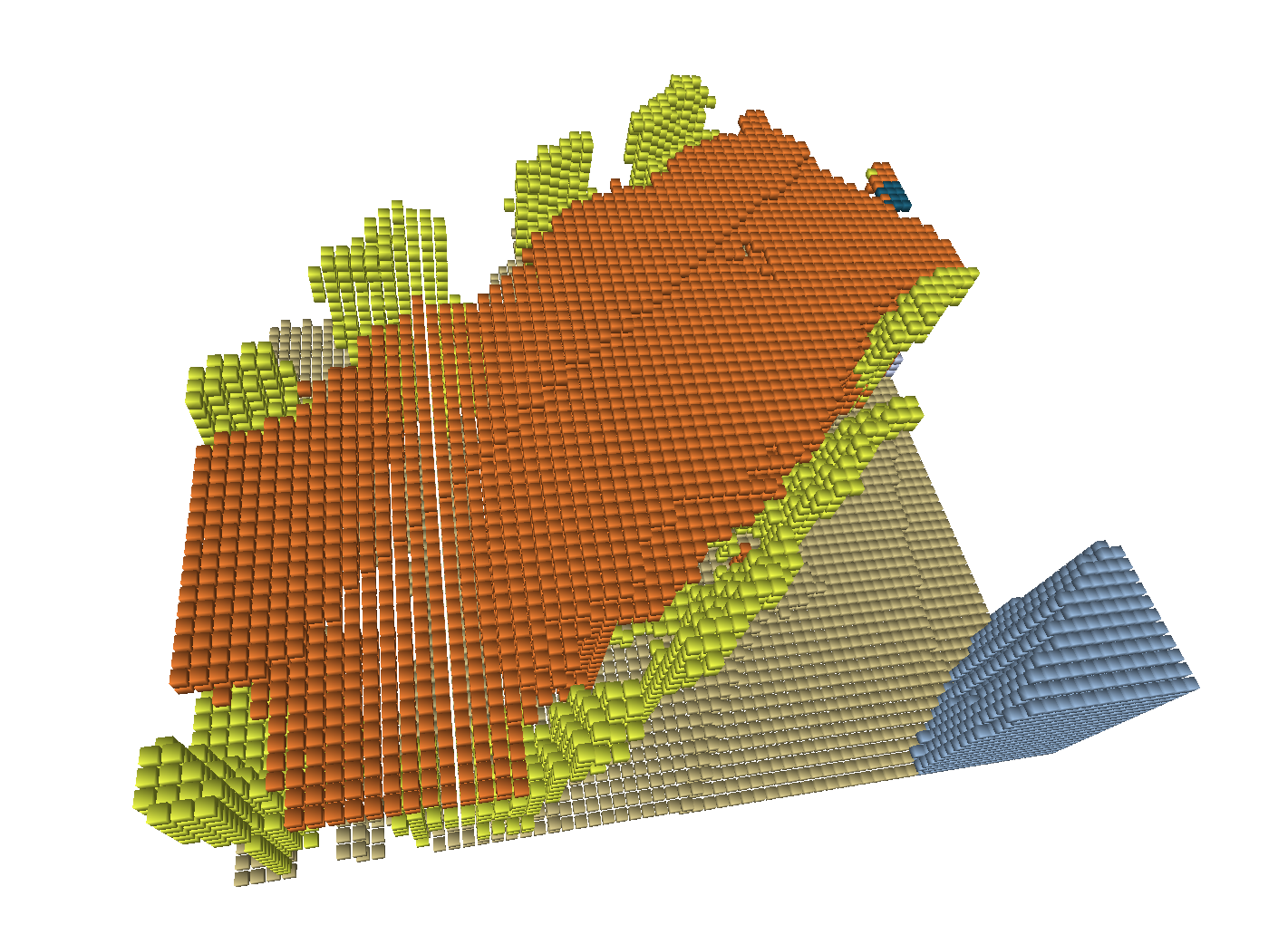}
\end{minipage}%
\begin{minipage}[b]{.18\textwidth}
  \centering
  Ground Truth
  \includegraphics[width=\textwidth, height=0.08\textheight]{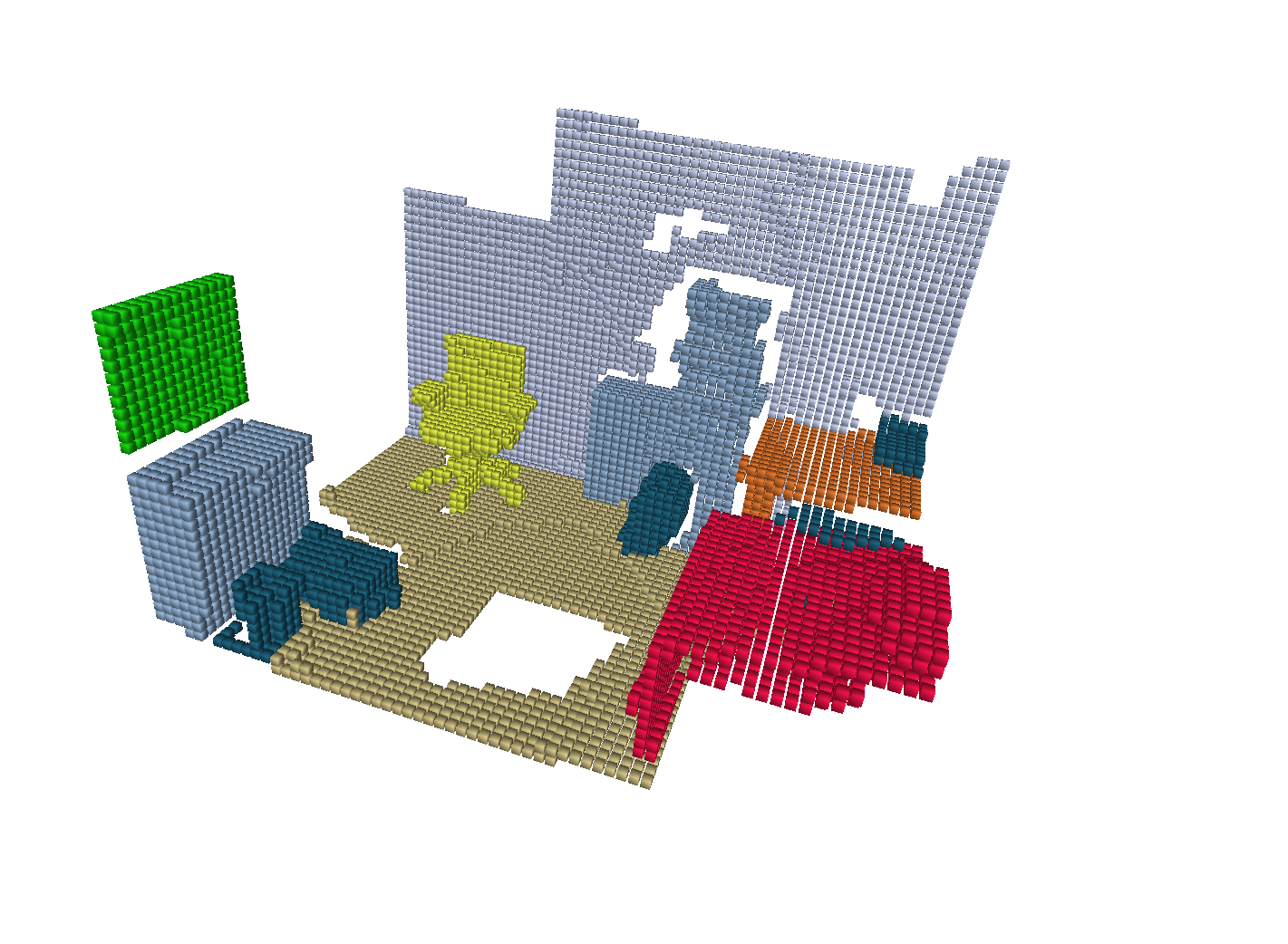}
  \includegraphics[width=\textwidth, height=0.08\textheight]{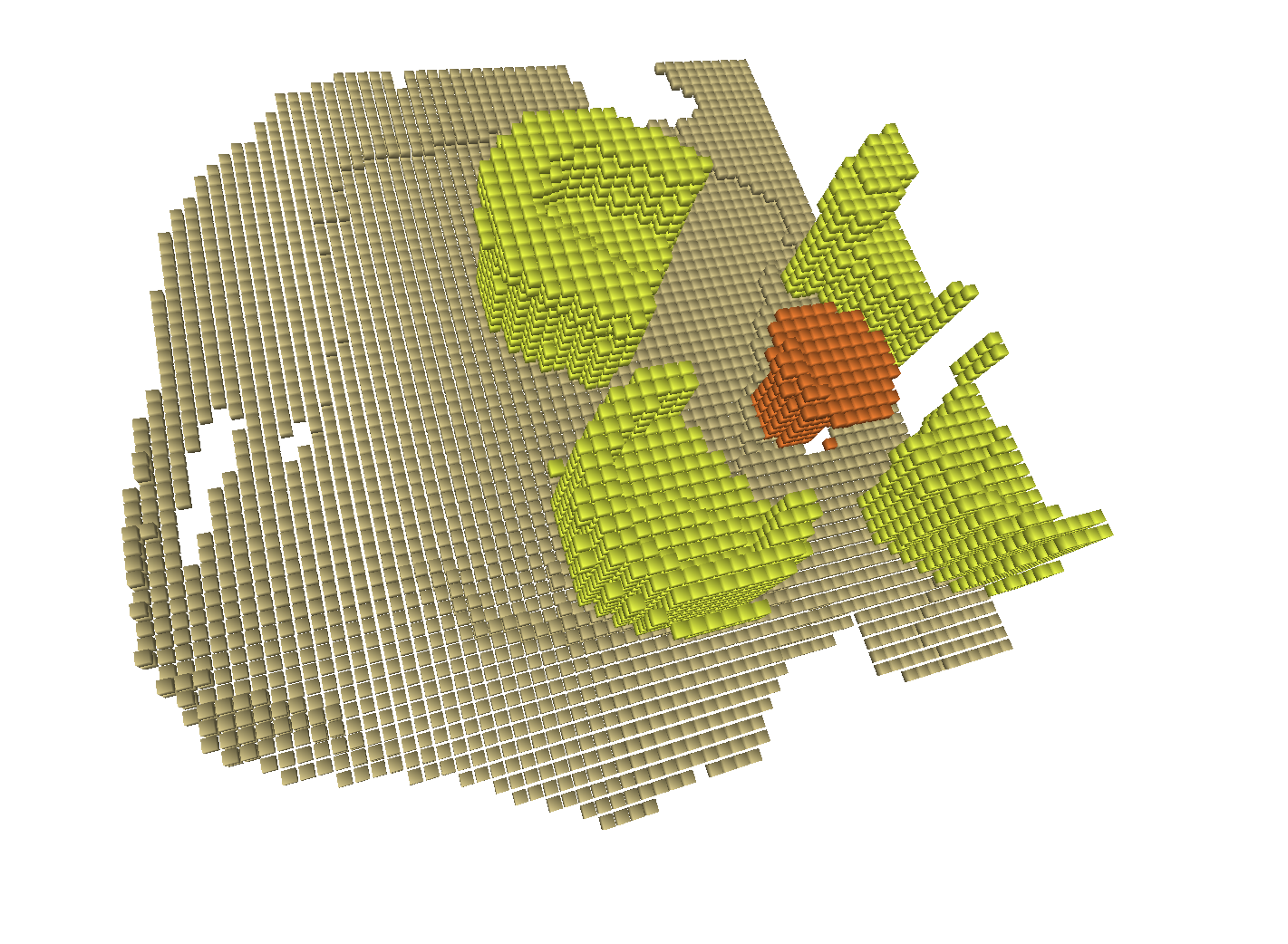}
  \includegraphics[width=\textwidth, height=0.08\textheight]{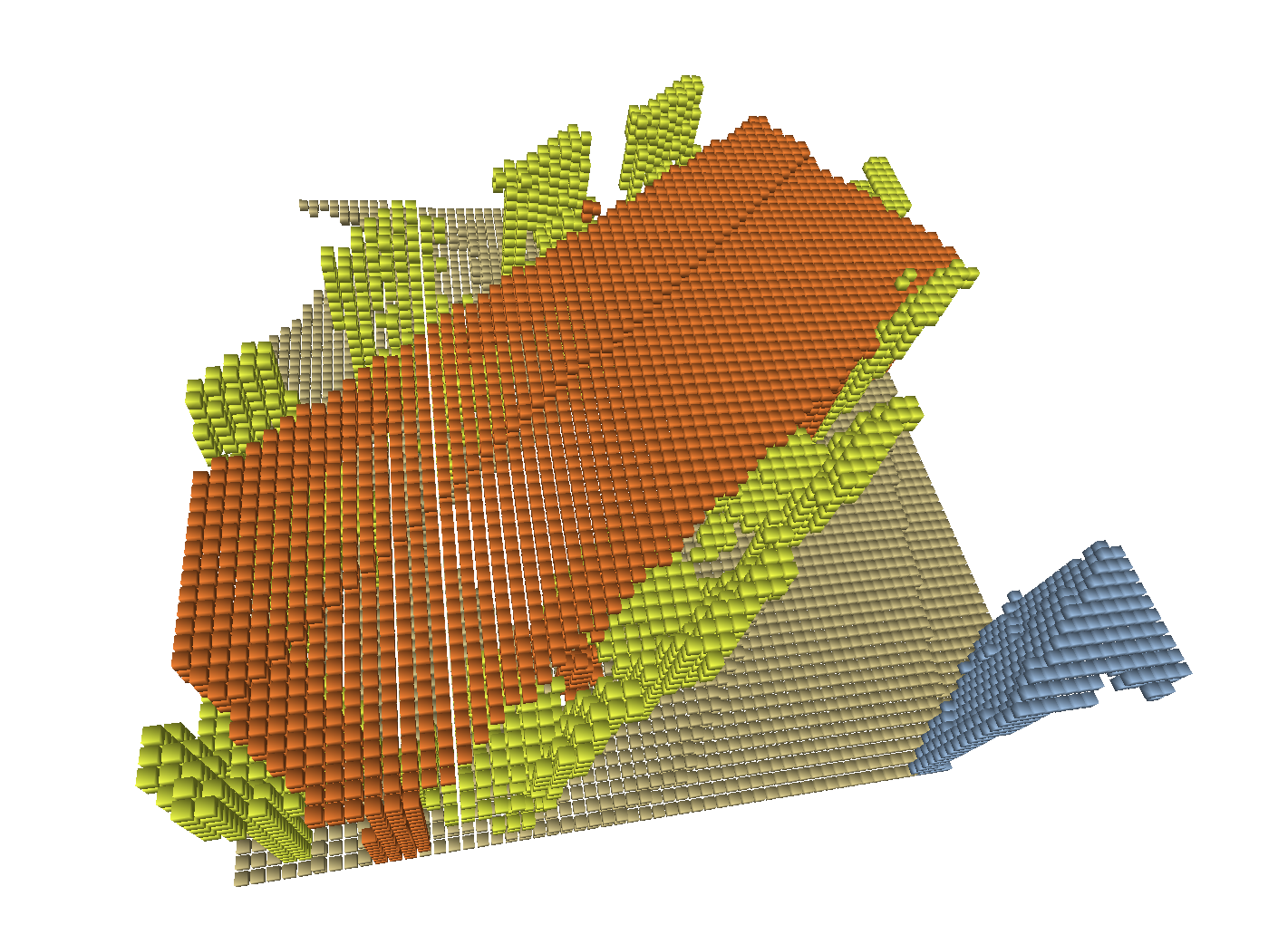}
\end{minipage}%
\begin{minipage}[b]{.13\textwidth}
  \centering
  \includegraphics[width=\textwidth, height=0.24\textheight]{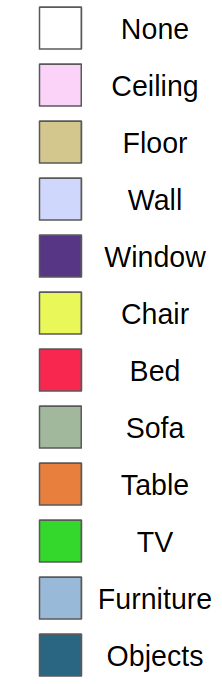}
\end{minipage}%
\\
\caption{Semantic scene completion comparison on some CompleteScanNet test scenes. While both ForkNet and SCFusion can accurately complete geometry and predict semantics, our method outperforms ForkNet if compared to the ground truth.}
\label{fig:qulitative_networks}
\end{figure*}

\subsubsection{Online map regularization} \label{sec:ssc_regularization}
As mentioned, the use of sub-maps enables our framework to be real-time, but it also brings in potential inconsistencies nearby the borders of the sub-maps due to the discretization of the global map. 
To improve 3D semantic reconstruction accuracy under this aspect, we apply a regularization approach for the global map based on a fully connected CRF model, whose use for 3D maps has been explored in \cite{panopticfusion, mccormac2017semanticfusion}, showing promising results. 

As explained in \ref{sec:ssc_extraction}, our map stores, at each voxel, only a label with an associated weight rather than the whole probability distribution. Hence, we use and modify the approach from \cite{panopticfusion} since it also relies on storing individual labels. We use the only voxel locations as regularization term. Differently from \cite{panopticfusion}, which uses the frequency for a certain label as probability estimate, we calculate the probability of a voxel for a certain label \(L\) as:
\begin{equation}
    p^{L}(v) = \max(W^{L}(v) / W^{L}_{max}, p^{L}_{min})
\end{equation}
where \(p^{L}_{min}\) is set to be slightly above the average label probability (in our case 0.1).
Experimental results will be presented showing how our map regularization method is able to improve the accuracy of 3D semantic reconstruction - in particular, quantitatively in table \ref{tab:quantitative_full}), as well as qualitatively in the supplementary material.



\section{Data generation} \label{sec:data_gen}
Due to the lack of a dataset including both completed 3D scene reconstructions and depth map sequences, it is hard to evaluate the performance of our approach, as well as to train our semantic scene completion network so that it can be applied on real data. 
Scene completion approaches \cite{dai2018scancomplete,wang2019forknet} relied on the synthetic SunCG dataset \cite{sscnet}, which is currently no longer available. 
Recently, \cite{sgnn} showed how a scene completion network can be trained with incomplete ground truth in a self-supervised manner. However, this approach does not predict semantic labels, this limiting its use on many applications of interest. 

\begin{table*}[!t]
    \resizebox{\textwidth}{!}{%
    \begin{tabular}{c|c|c|c|c|c|c|c|c|c|c|c|c}
         & Ceili. & Floor & Wall & Window & Chair & Bed & Sofa & Table & TV & Furni & Object & Mean \\
        \hline
        ForkNet+Fusion (-s)\cite{wang2019forknet} & 
        0.131 & 0.390 & 0.241 & 0.000 & 0.147 & 0.061 & 0.233 & 0.291 & 0.003 & 0.212 & 0.033 & 0.158 \\
        ScanComplete (-s)\cite{dai2018scancomplete}& 
        0.225 & \textbf{0.541} & \textbf{0.440} & 0.020 & 0.312 & 0.055 & 0.177 & \textbf{0.371} & 0.007 & 0.195 & 0.107 & 0.222 \\
        Proposed w/o CRF (-s) & 
        \textbf{0.250} & 0.532 & 0.413 & 0.121 & 0.348 & 0.277 & 0.339 & 0.350 & 0.084 & 0.287 & 0.130 & 0.284 \\
        Proposed (-s) & 
        0.236 & 0.537 & 0.437 & \textbf{0.134} & \textbf{0.362} & \textbf{0.282} & \textbf{0.346} & 0.356 & \textbf{0.086} & \textbf{0.299} & \textbf{0.136} & \textbf{0.292} \\
        \hline
        ForkNet+Fusion (-f)\cite{wang2019forknet} & 
        0.052 & 0.225 & 0.123 & 0.000 & 0.092 & 0.057 & 0.182 & 0.149 & 0.001 & 0.126 & 0.018 & 0.093\\
        ScanComplete (-f)\cite{dai2018scancomplete}& 
        \textbf{0.164} & \textbf{0.393} & \textbf{0.350} & 0.018 & 0.204 & 0.038 & 0.112 & \textbf{0.277} & 0.006 & 0.132 & 0.078 & 0.161\\
        Proposed (-f) & 
        0.128 & 0.329 & 0.265 & \textbf{0.096} & \textbf{0.225} & \textbf{0.207} & \textbf{0.264} & 0.210 & \textbf{0.074} & \textbf{0.192} & \textbf{0.086} & \textbf{0.189}\\
        \end{tabular}
    }
\caption{Comparison in IoU for semantic scene completion on CompleteScanNet.
IoU is measured on both visible surfaces only (-s) and on the entire scan (-f). }
\label{tab:quantitative_full}
\end{table*}

\begin{figure*}[t]
\centering
\begin{minipage}[b]{.20\textwidth}
  \centering
  Reconstruction
  \includegraphics[width=\textwidth, height=0.09\textheight]{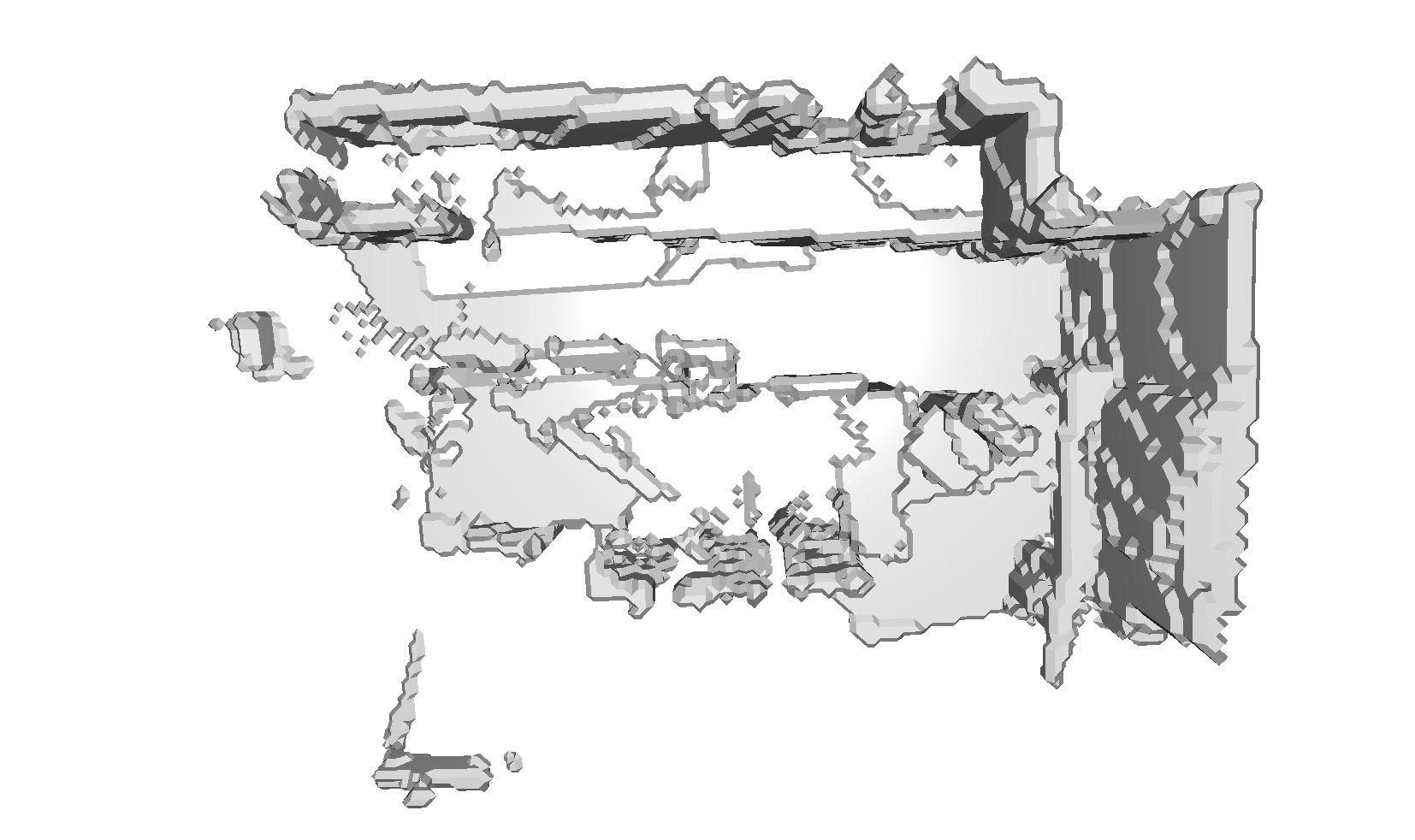}
  \includegraphics[width=\textwidth, height=0.09\textheight]{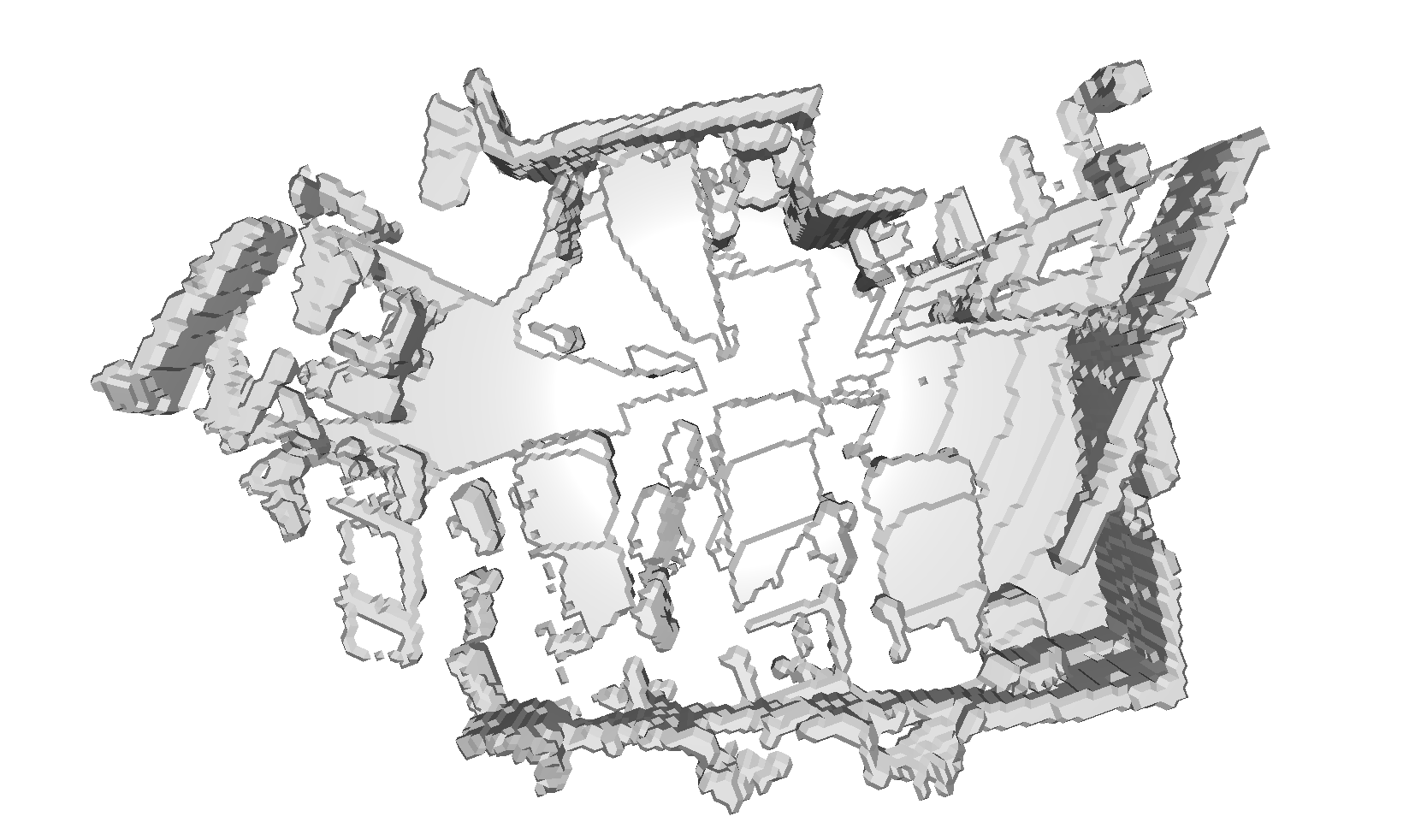}
  \includegraphics[width=\textwidth, height=0.09\textheight]{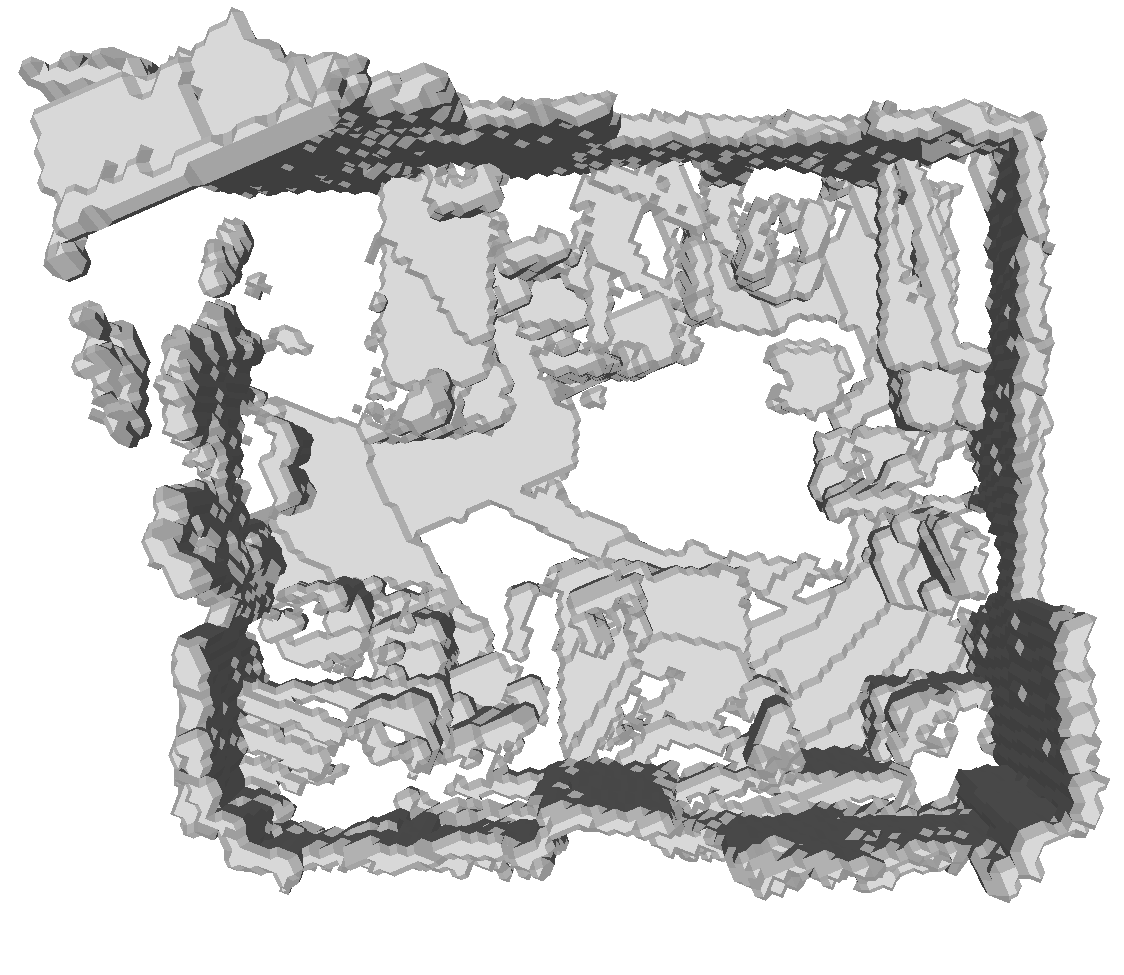}
  \includegraphics[width=\textwidth, height=0.09\textheight]{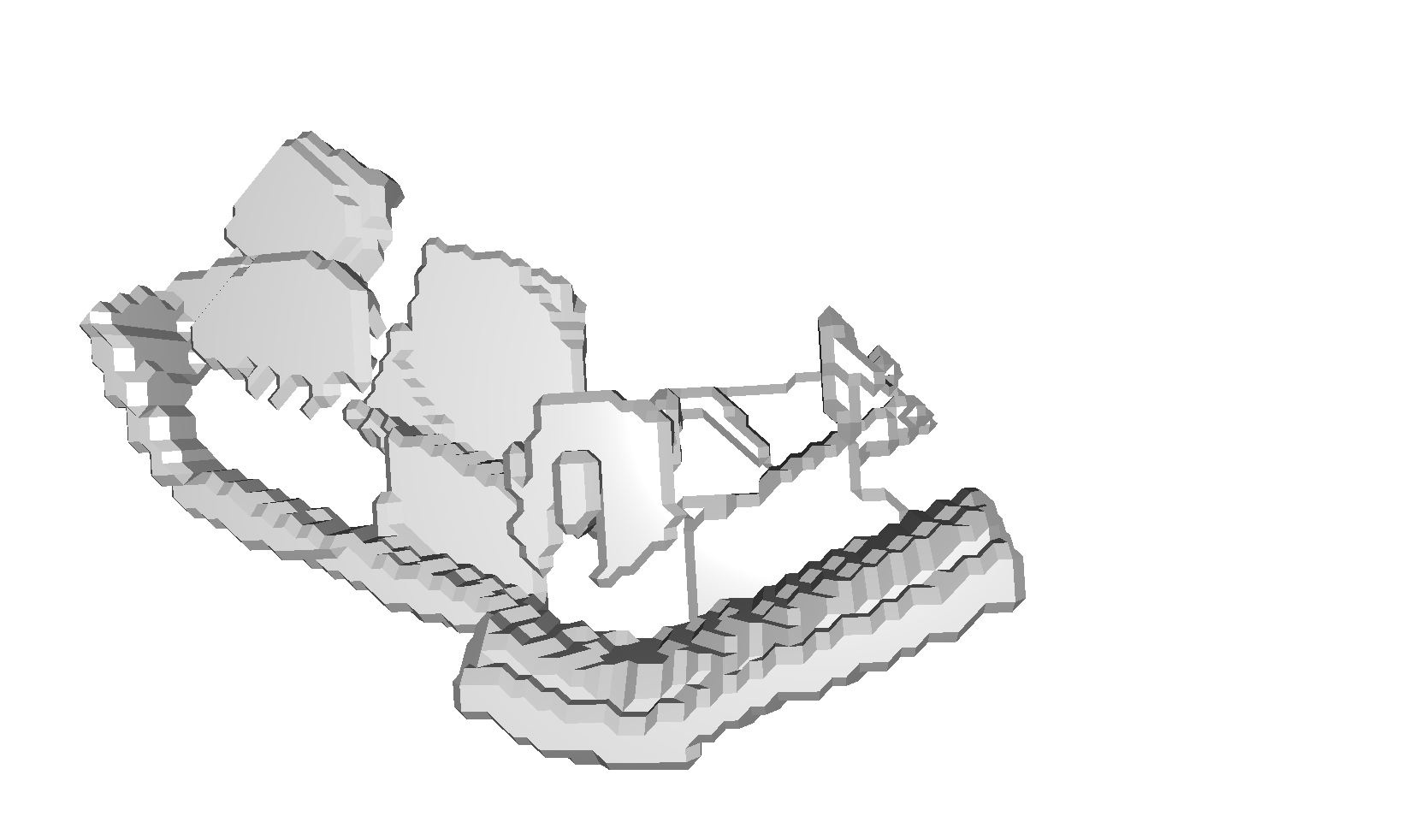}
\end{minipage}%
\begin{minipage}[b]{.20\textwidth}
  \centering
  ForkNet+Fusion \cite{wang2019forknet}
  \includegraphics[width=\textwidth, height=0.09\textheight]{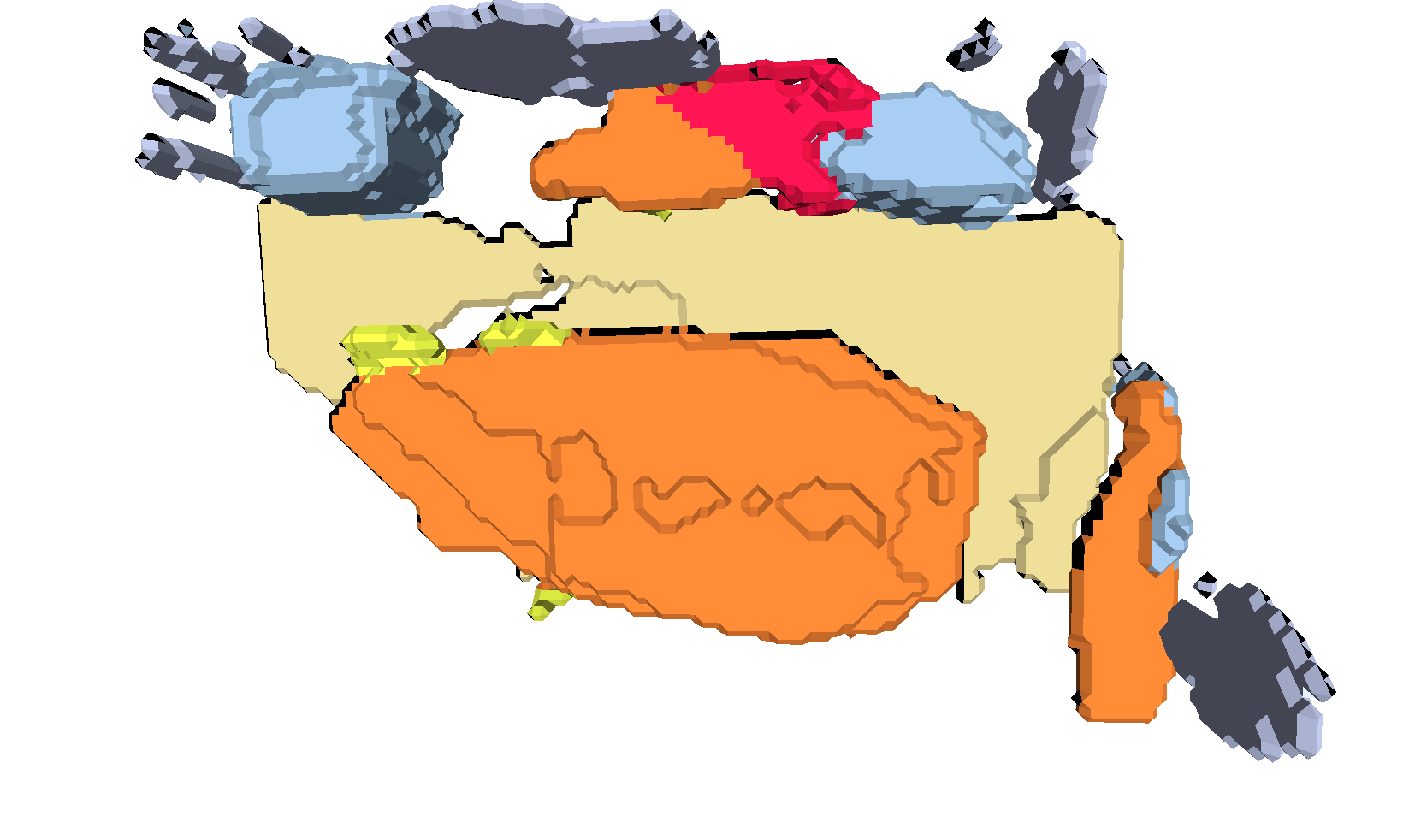}
  \includegraphics[width=\textwidth, height=0.09\textheight]{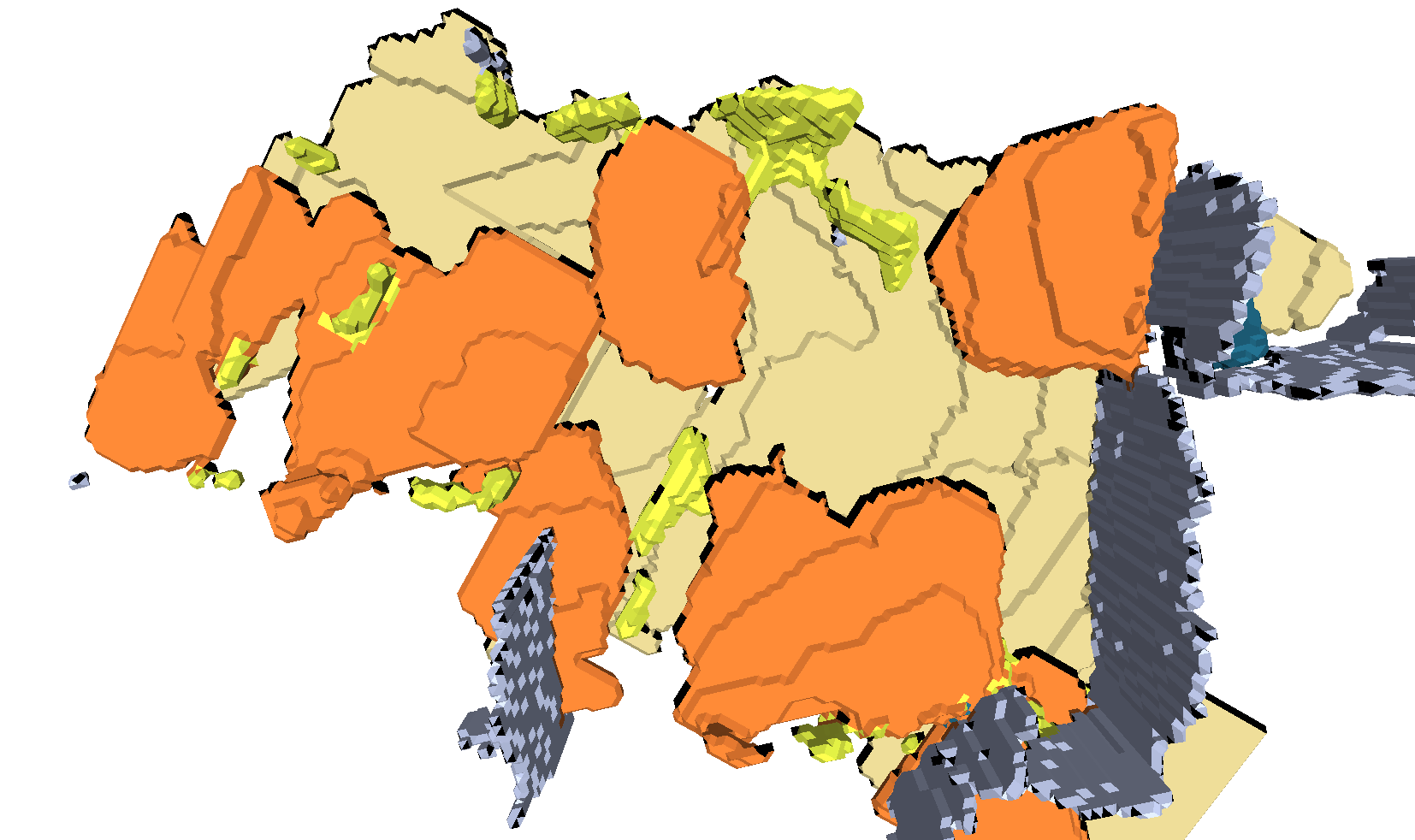}
  \includegraphics[width=\textwidth, height=0.09\textheight]{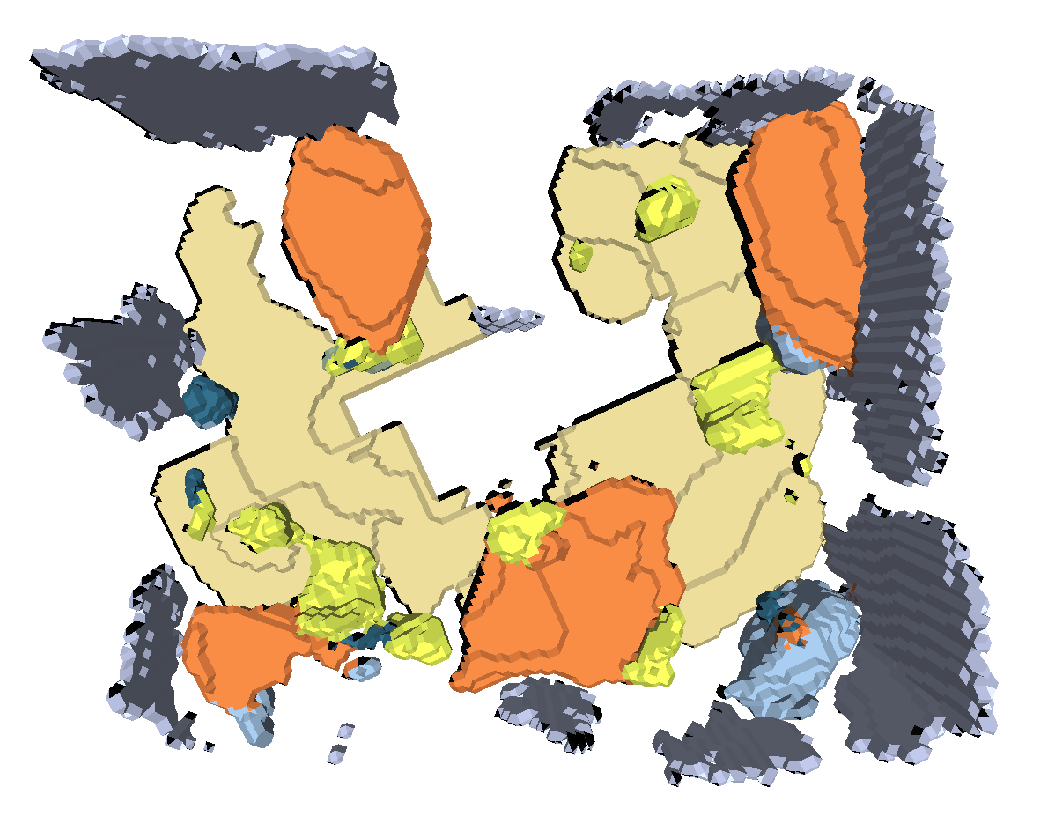}
  \includegraphics[width=\textwidth, height=0.09\textheight]{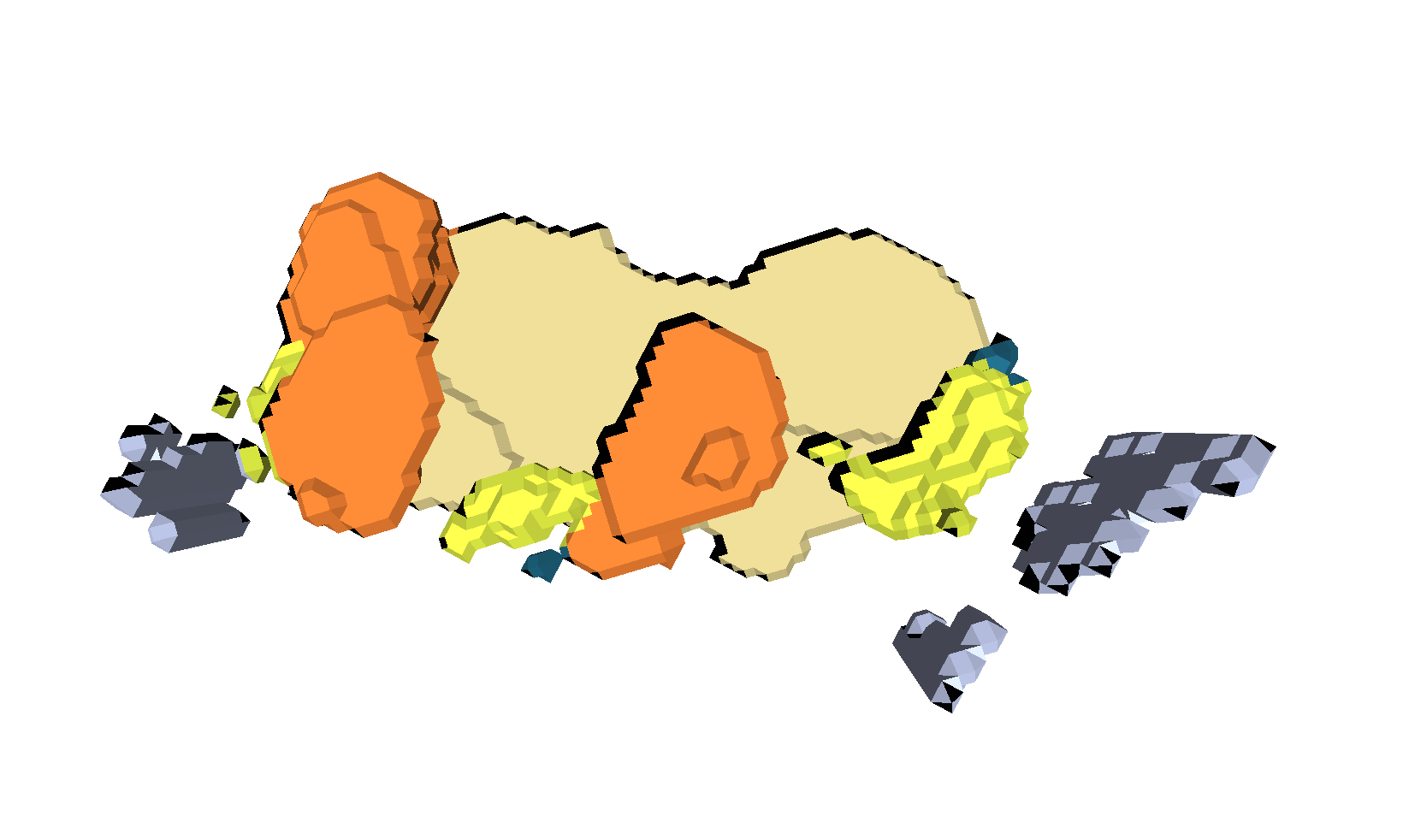}
\end{minipage}%
\begin{minipage}[b]{.20\textwidth}
  \centering
  ScanComplete \cite{dai2018scancomplete}
  \includegraphics[width=\textwidth, height=0.09\textheight]{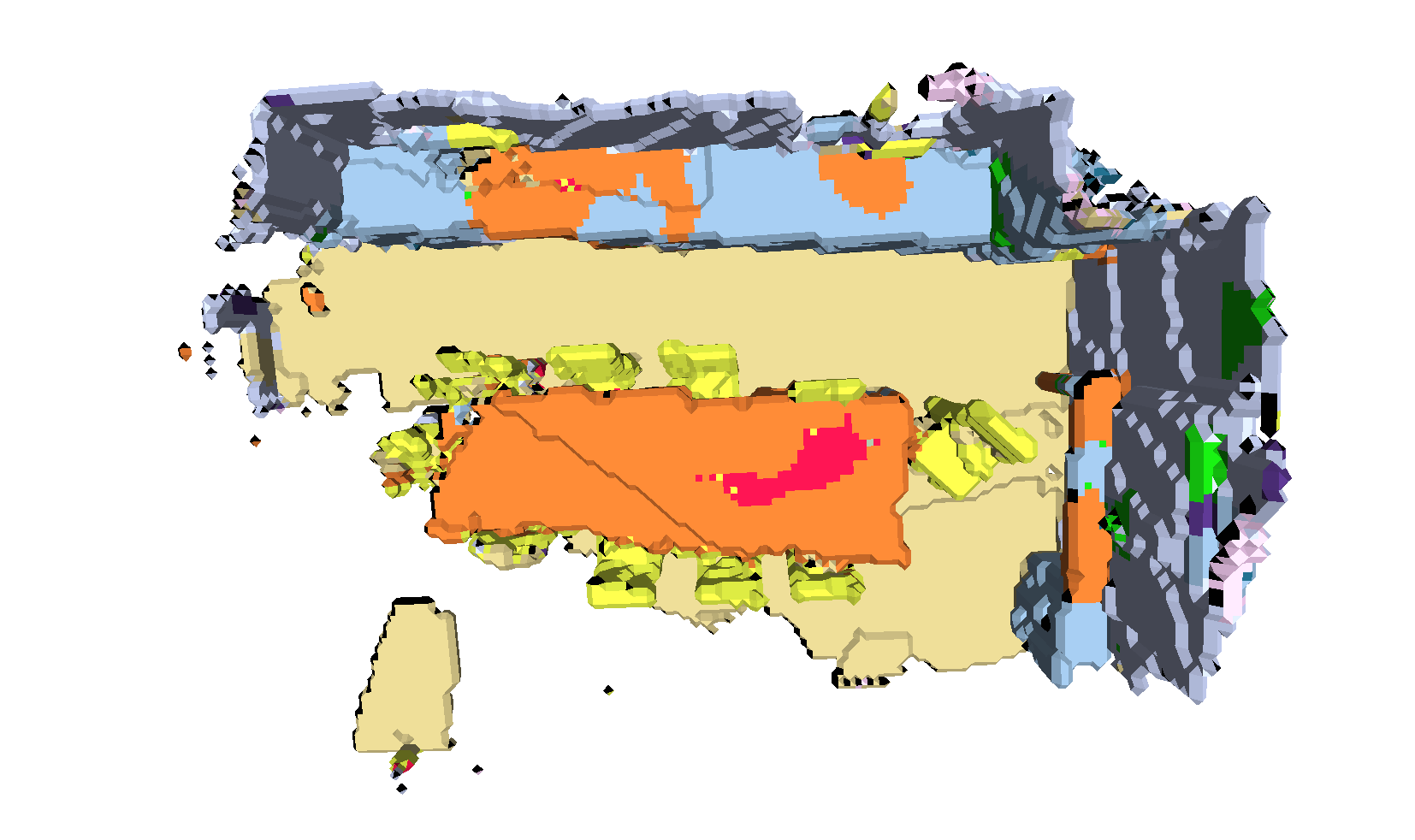}
  \includegraphics[width=\textwidth, height=0.09\textheight]{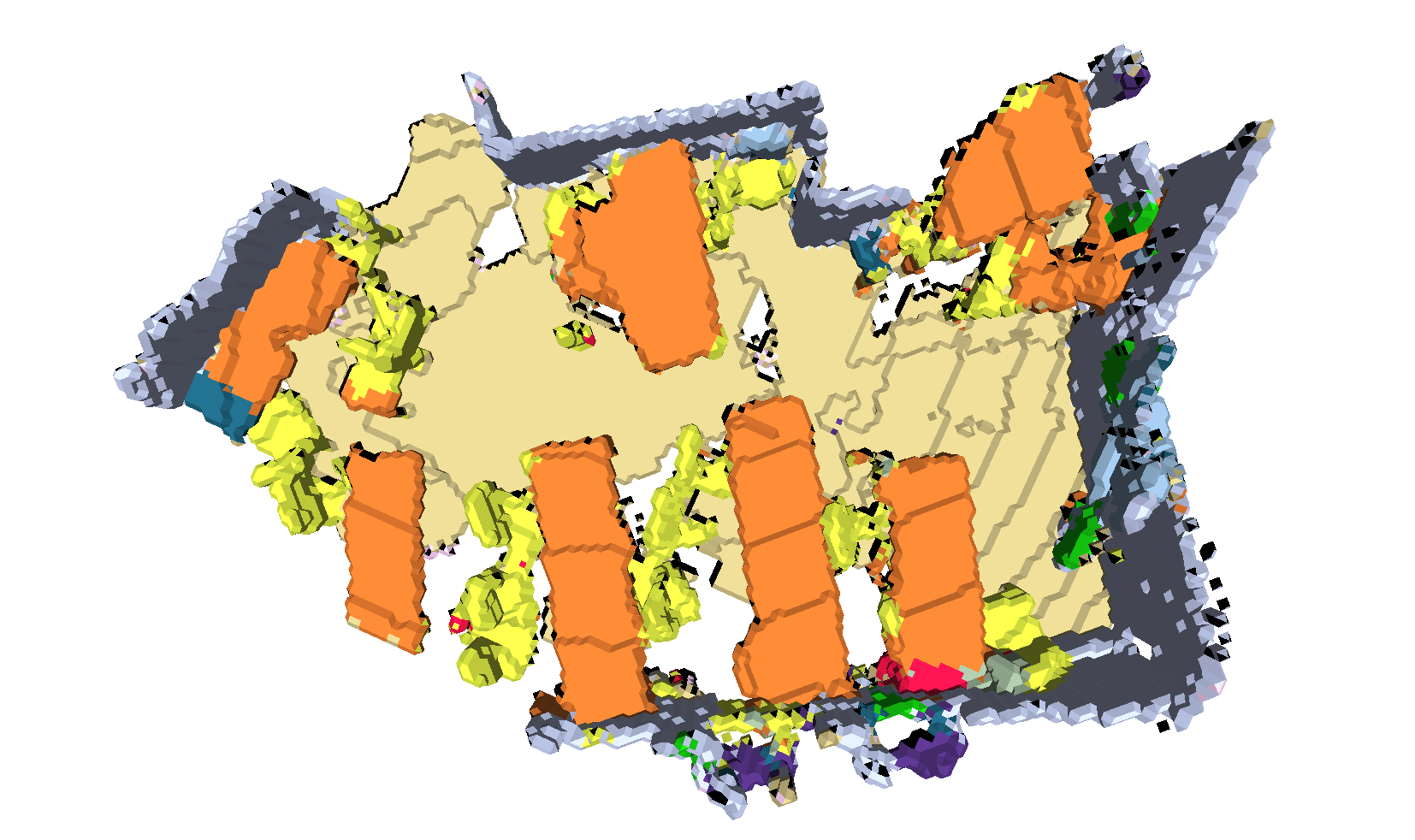}
  \includegraphics[width=\textwidth, height=0.09\textheight]{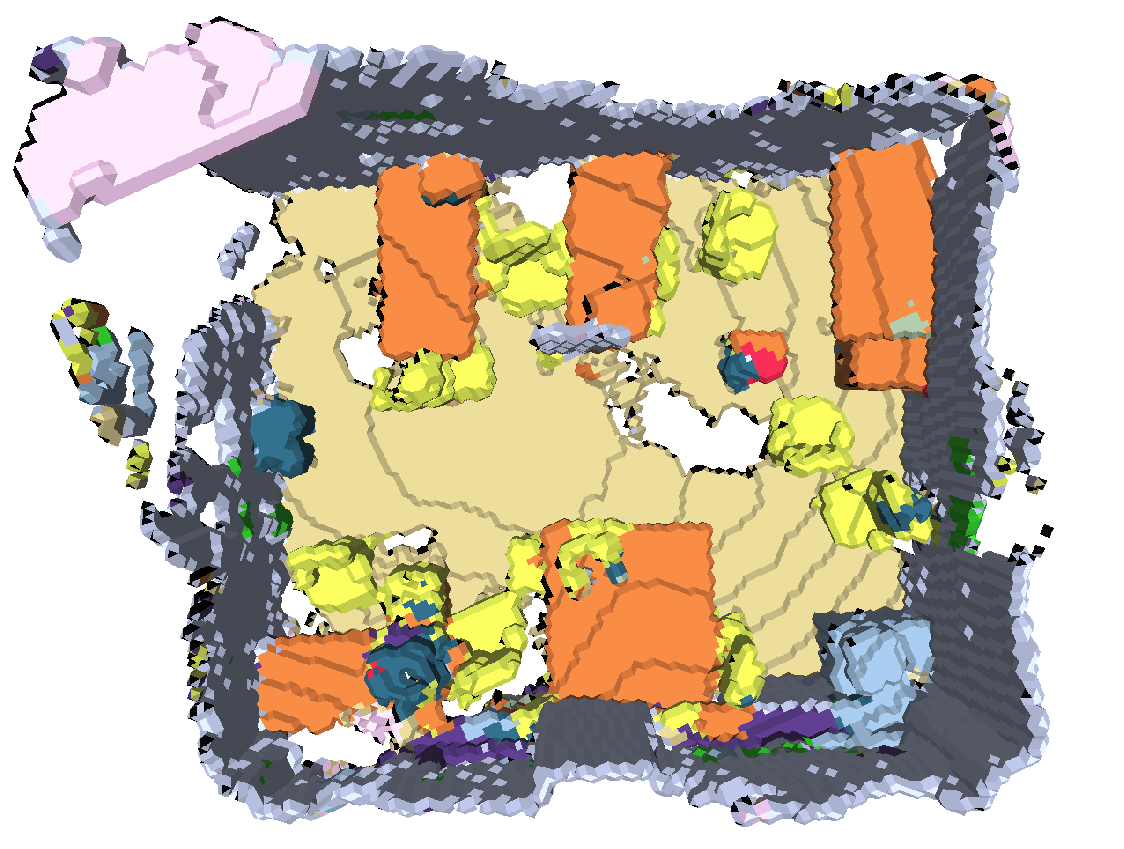}
  \includegraphics[width=\textwidth, height=0.09\textheight]{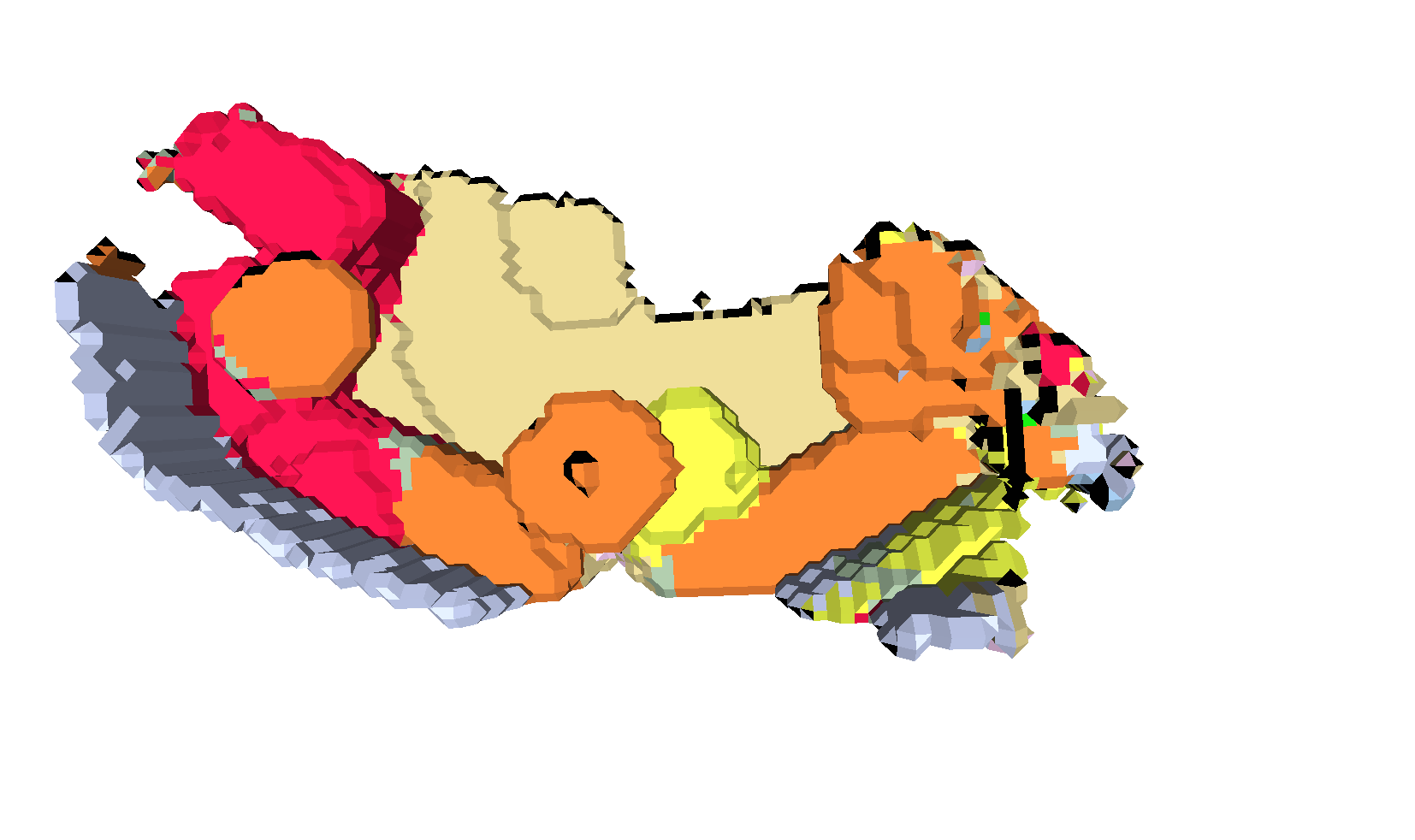}
\end{minipage}%
\begin{minipage}[b]{.20\textwidth}
  \centering
  Ours
  \includegraphics[width=\textwidth, height=0.09\textheight]{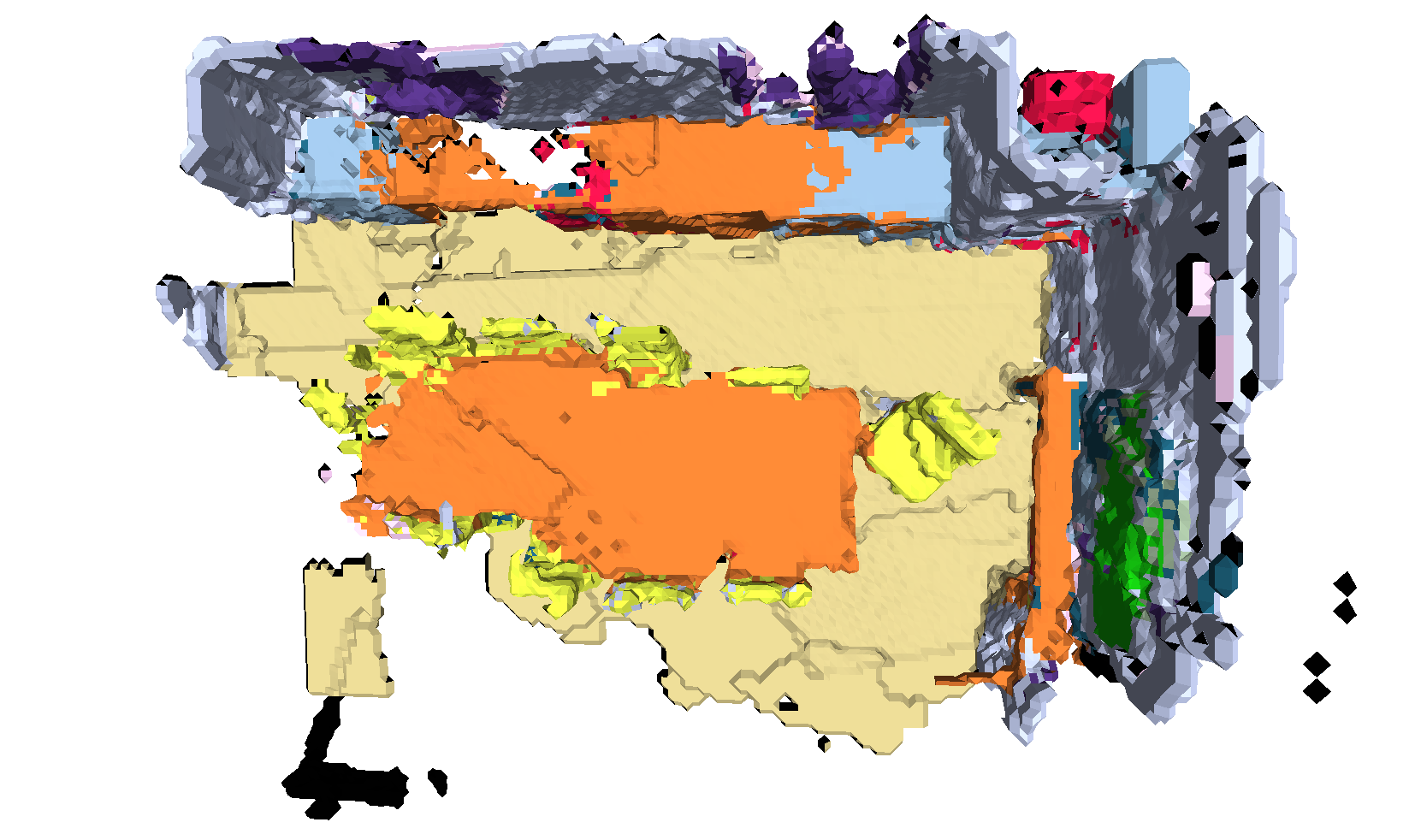}
  \includegraphics[width=\textwidth, height=0.09\textheight]{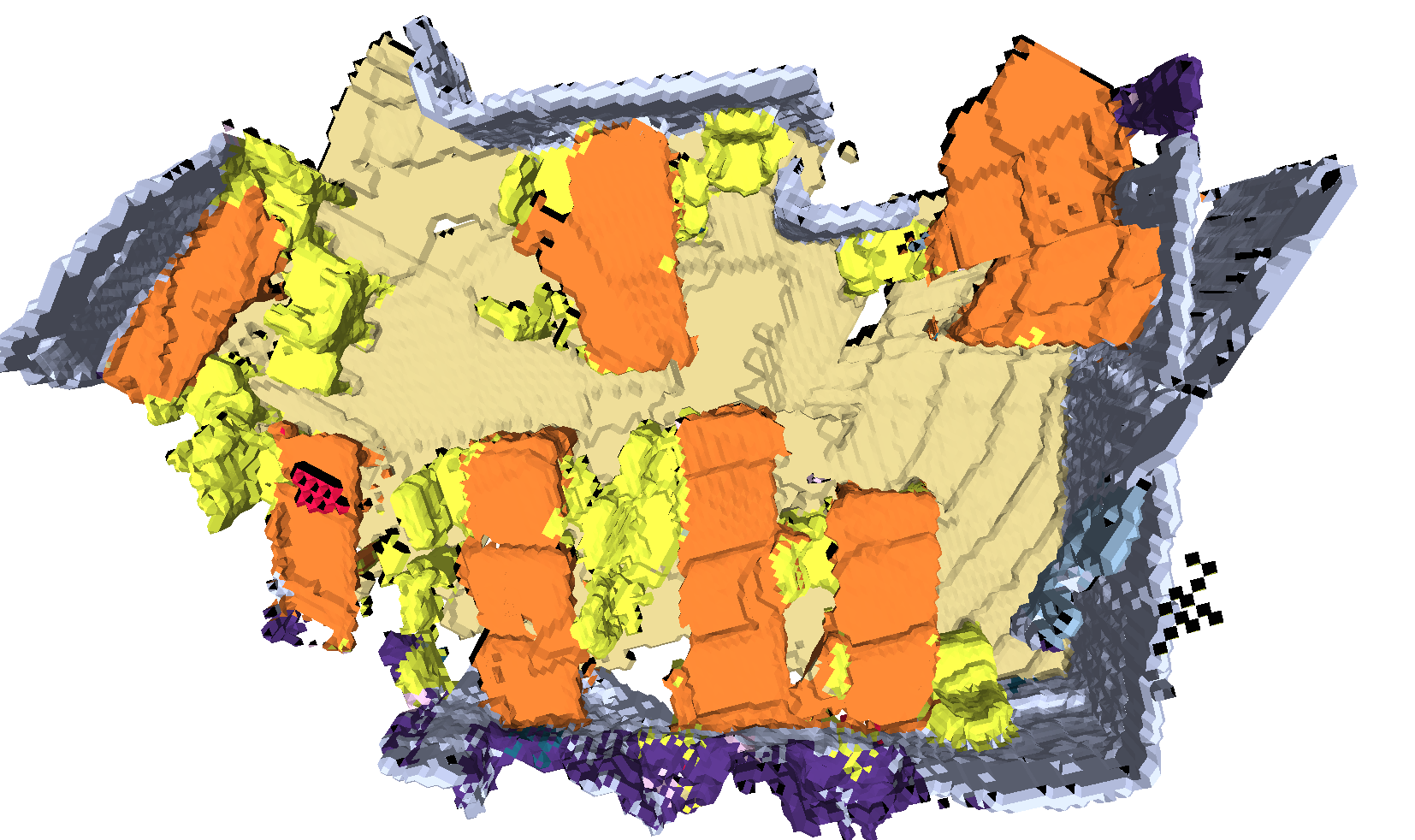}
  \includegraphics[width=\textwidth, height=0.09\textheight]{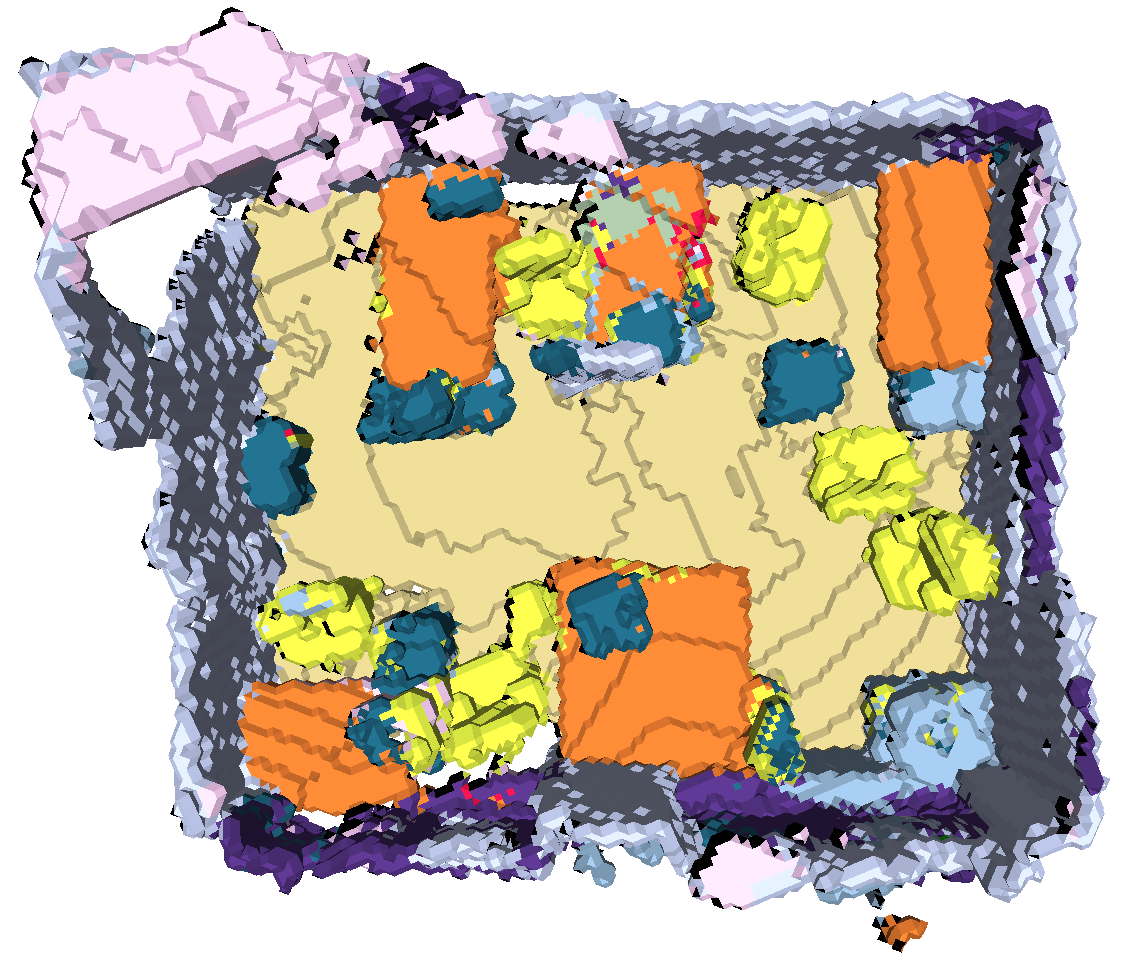}
  \includegraphics[width=\textwidth, height=0.09\textheight]{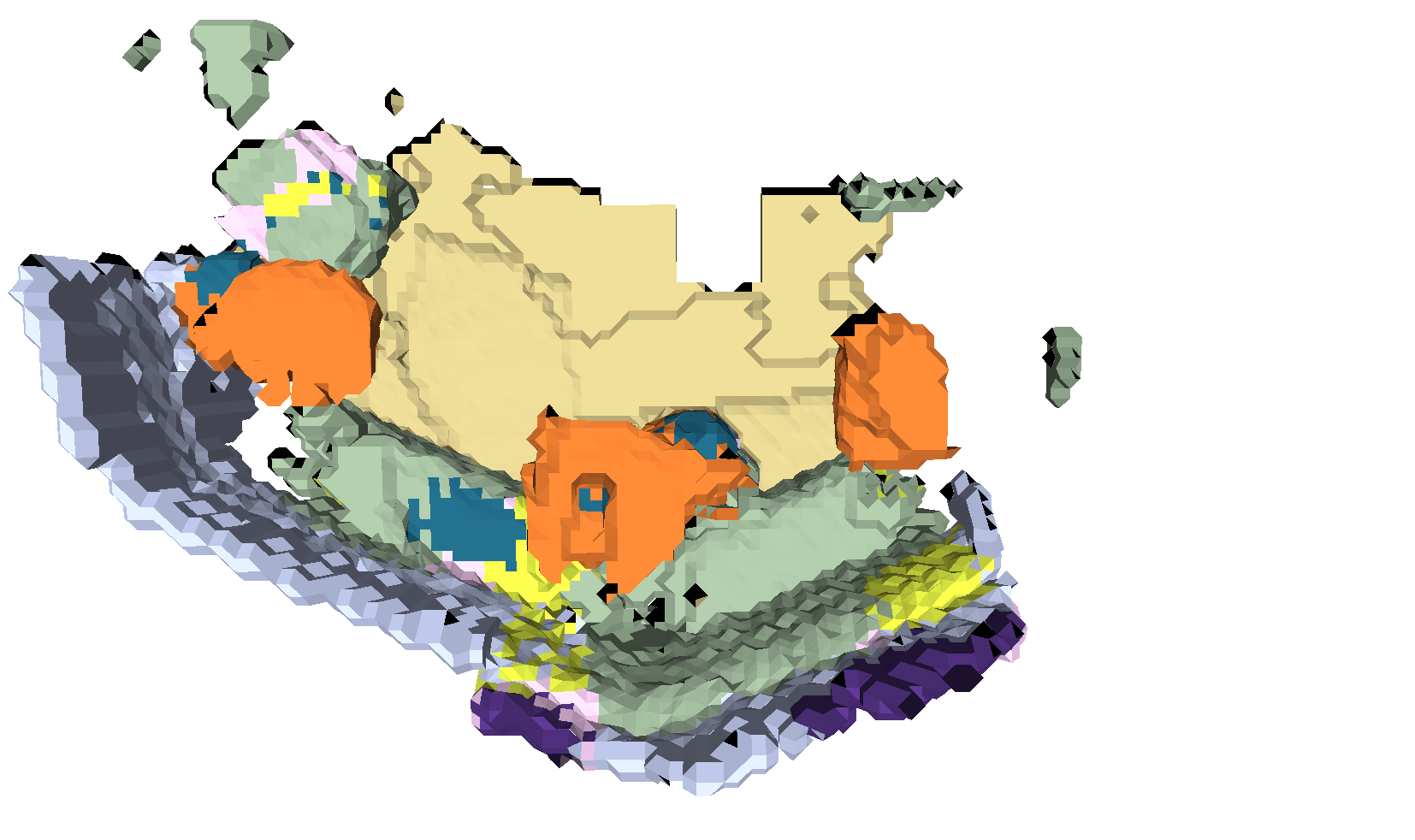}
\end{minipage}%
\begin{minipage}[b]{.20\textwidth}
  \centering
  Ground Truth
  \includegraphics[width=\textwidth, height=0.09\textheight]{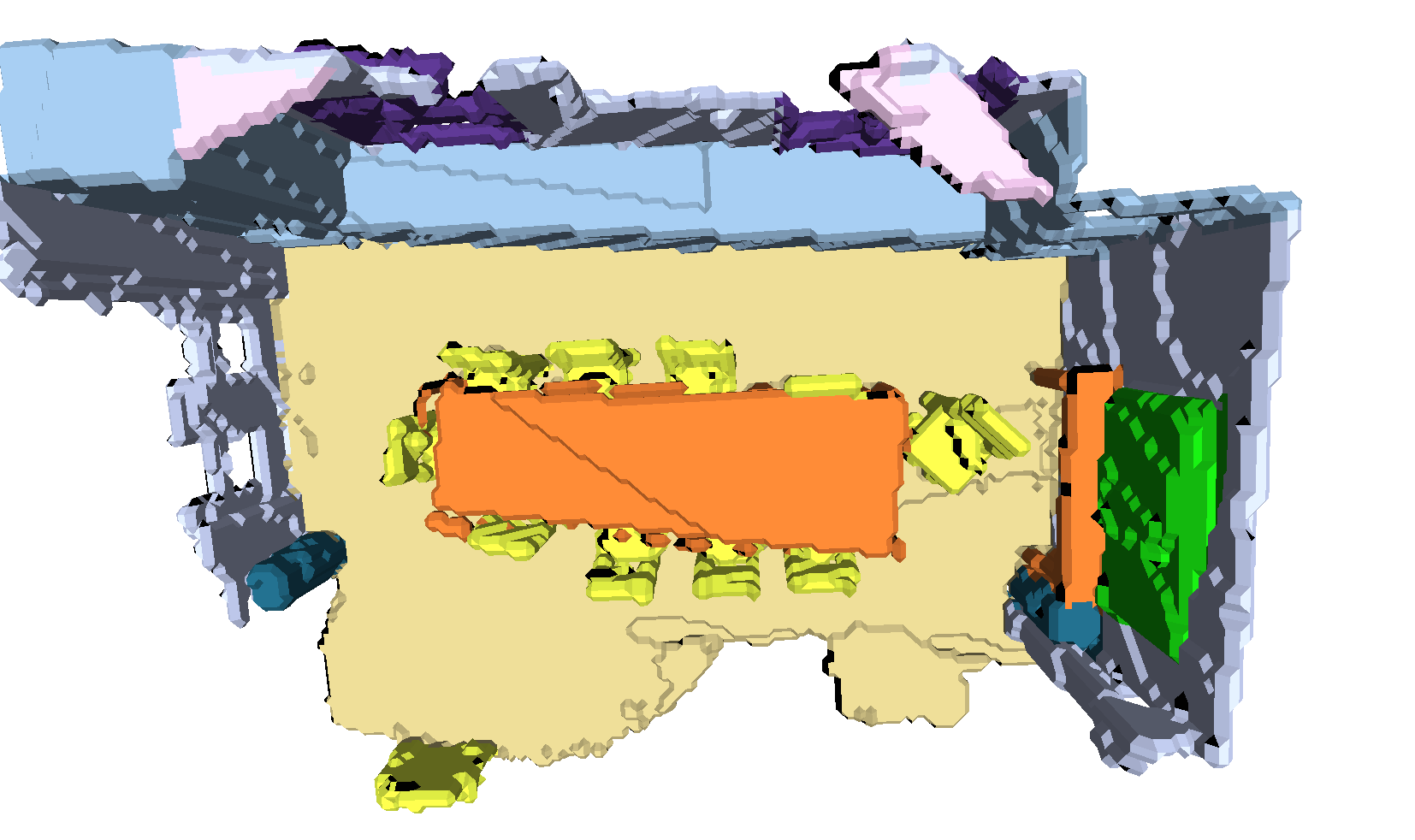}
  \includegraphics[width=\textwidth, height=0.09\textheight]{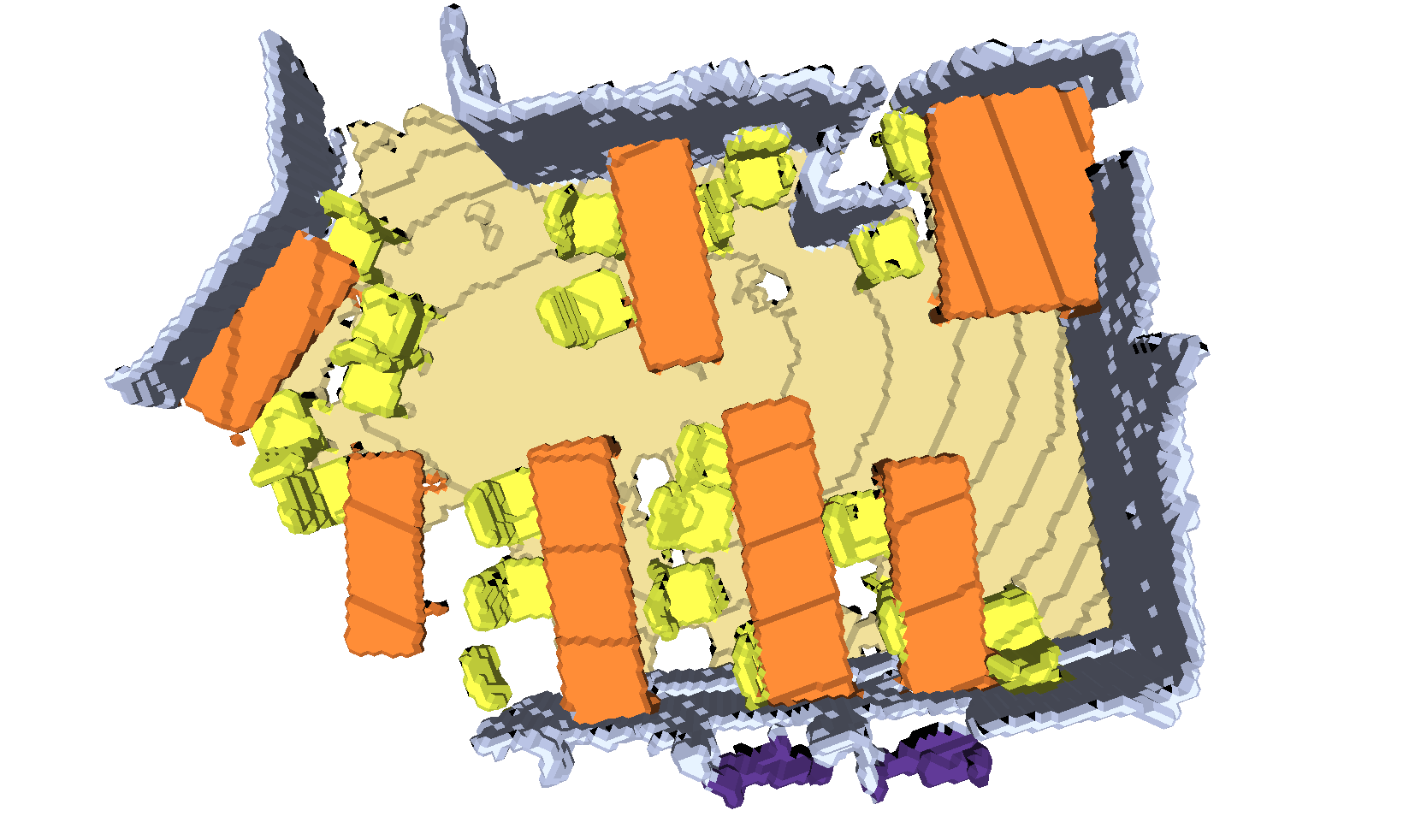}
  \includegraphics[width=\textwidth, height=0.09\textheight]{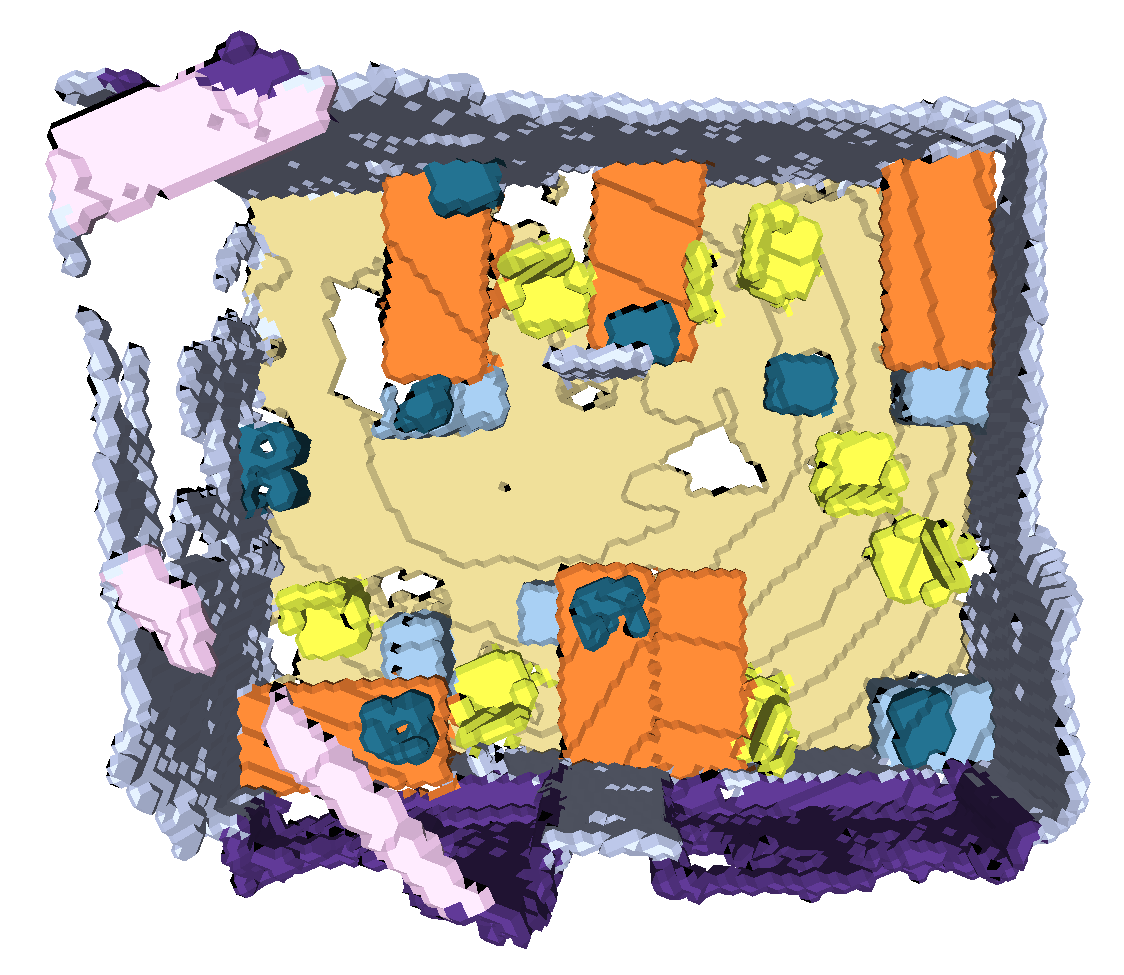}
  \includegraphics[width=\textwidth, height=0.09\textheight]{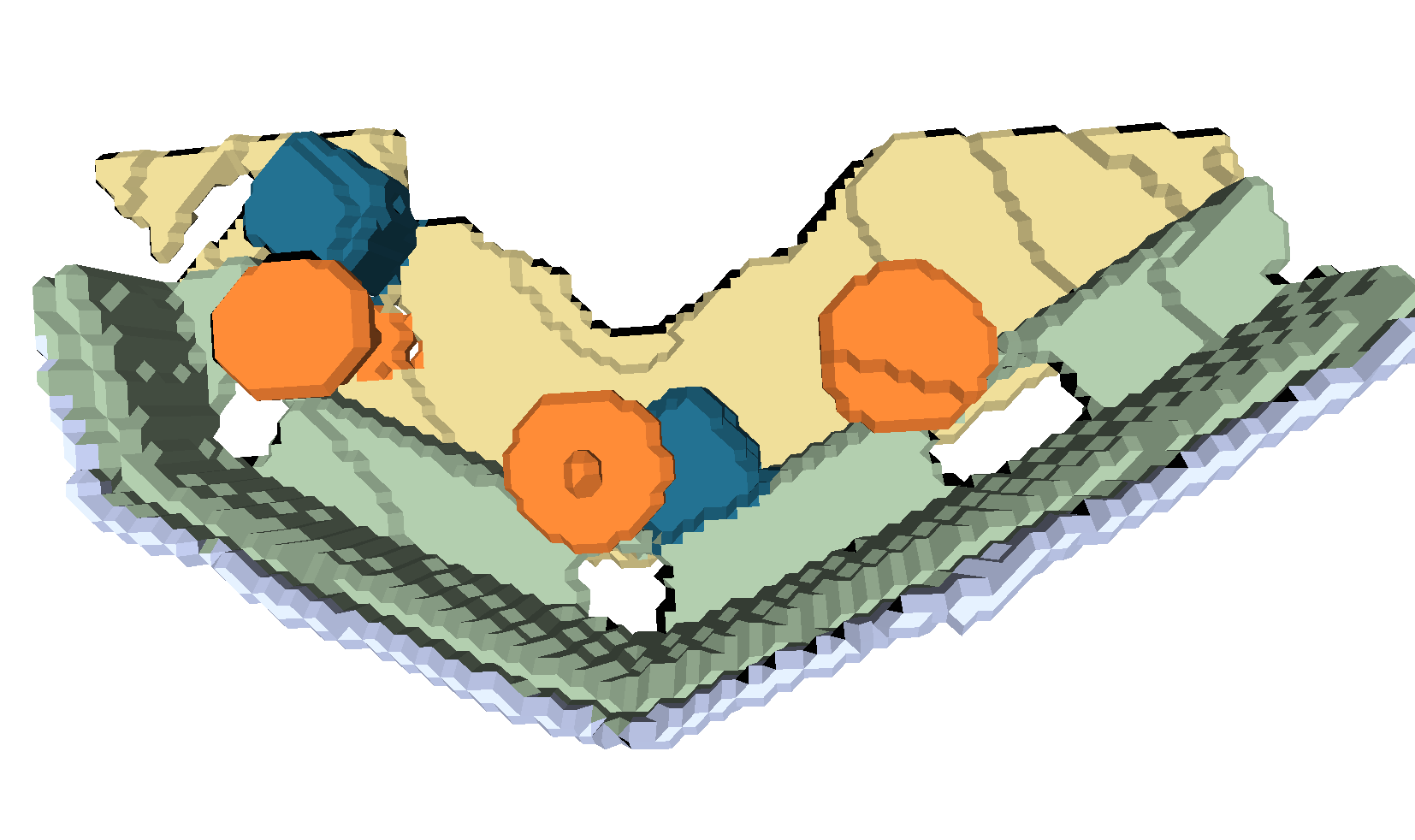}
\end{minipage}%
\\
\begin{minipage}[b]{\textwidth}
\centering
\includegraphics[width=1.0\textwidth]{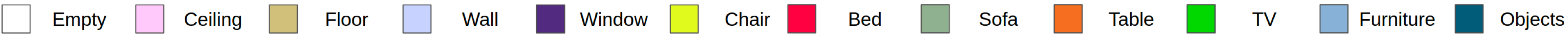}
\end{minipage}

\caption{Semantic scene completion comparison on some CompleteScanNet test scenes. 
}
\label{fig:qulitative_full_scannet}
\end{figure*}

To handle this issue, we developed a new benchmark where 3D object models are added to semantically annotated real-world scenes via model fitting. 
We use the model alignment annotations provided in Scan2CAD dataset \cite{scan2cad} to fit 3D models from ShapeNet \cite{shapenet} in the 3D scenes of ScanNet \cite{dai2017scannet}. We refer to this new dataset as \textbf{CompleteScanNet}. 
During model fitting, we replace the original object instances with their corresponding fully 3D object models to prevent geometry misalignments among object shapes. 
Then, we generate a depth sequence for each scene with full 3D objects by rendering depth maps. 
%
Note that the generated scenes still include incomplete parts due to the presence of incomplete background parts, e.g. floor, wall, ceiling. To help the network to better learn completion on incomplete parts, we devised a skip-frame training approach, where only one frame every 200 is processed, while the rest is discarded. This generates relatively more incomplete input scenes for training. Experimental results show that a network trained with this technique is able to predict a more complete and accurate scene than when trained on sequences that include all frames (see figures \ref{fig:qulitative_networks}, \ref{fig:qulitative_full_scannet}). 

Following this data generation pipeline, we use occupancy mapping with ground truth poses from ScanNet and rendered depth sequences to reconstruct a scene with voxel size of 5 \(cm^3\). Then we use uniform sampling to sample sub-maps with a constant size of [\(64\times 64\times 64\)]. 
To ensure that the extracted sub-maps are not empty and have enough object variation, a sub-map is discarded if the percentage of empty voxels exceeds 95\% and the number of different labels appearing in the scene is less than 2.
After sub-map extraction and filtering , 45448 training and 11238 testing samples are generated from the initial 1201 training and 312 validation scenes. 
Finally, ScanNet labels are mapped to the SunCG \cite{sscnet} ones to prevent confusion during training caused by objects having different labels but similar shapes.
%
%
%
\section{Experimental results}
Experiments are conducted to validate the performance of, respectively, the proposed network and the overall SCFusion framework on the CompleteScanNet dataset. 
First we validate the design of our network by comparing the performance on single prediction to its baseline method, \ie ForkNet \cite{wang2019forknet} with the use of intersection-over-union (IoU) as the metric on the CompleteScanNet sub-map test set. 
Then, we evaluate the performance of SCFusion on entire scenes by comparing it against the state of the art in offline scene completion, i.e. ScanComplete \cite{dai2018scancomplete}. Finally, we show the effectiveness of the proposed fusion method by comparing it against fusion of ForkNet predictions.

\subsection{Parameters and experimental environment}
Training and testing is done on an Intel(R) Core i7-8700 CPU @ 3.20GHz, 64GB DDR4 RAM, 2 x NVIDIA GeForce RTX 2080ti. Our network is trained with Adam optimizer with a learning rate of 0.001, batch size of 4 with an accumulated gradient factor of 4. The whole network is trained for ~40 hours until convergence.

\subsection{Network evaluation}  \label{sec:eval_network_performance}
We evaluate the performance of our network against ForkNet \cite{wang2019forknet} on the 11238 test sub-maps of CompleteScanNet. Both networks are trained from scratch until convergence on the CompleteScanNet training set. Since ForkNet takes inverted TSDFs as input, we generate its training data using TSDF fusion.
Results in figure \ref{fig:qulitative_networks} and table \ref{tab:quantitative_networks} show that our method tends to predict more precise object shapes than ForkNet, which instead predicts coarser geometries. Moreover, the predicted scenes of our method tend to be more complete than ForkNet. Although the dataset still partially present incomplete ground truth surfaces, both ForkNet and our method trained with the proposed skip-frame approach are able to predict reasonably complete scenes. 

We provide an ablation study on the changes we made on the ForkNet architecture in the Section 2 of the supplementary material. 
\begin{table}[t]
\resizebox{\columnwidth}{!}{%
\begin{tabular}{|l|l l l|}
\hline
Thread & Operation & \begin{tabular}[c]{@{}l@{}}Mean \\ (ms)\end{tabular}  & \begin{tabular}[c]{@{}l@{}}Std\\ (ms)\end{tabular} \\ \hline
\multirow{2}{*}{Scene Reconstruction} 
 & Input Processing & 0.039 & 0.097 \\ \cline{2-4} 
 & Mapping & 5.956 & 23.484   \\ \hline
\multirow{4}{*}{Semantic Scene Completion} 
 & Sub-Map Extraction & 0.620 & 0.467 \\ \cline{2-4} 
 & Semantic Scene Completion  & 123.067  & 8.898  \\ \cline{2-4}
 & Sub-Map Fusion             & 1.124  & 0.232 \\ \cline{2-4} 
 & CRF Regularization         & 33.261 & 12.092 \\ \hline
                                           \hline
Average run-time per-frame  & & 10.863 & 25.486 \\
\hline
\end{tabular}
}
\caption{Runtime analysis of the SCFusion stages averaged on the scene0645\_00 sequence from CompleteScanNet.}
\label{tab:runtime}
\end{table}
\subsection{Full framework evaluation}  \label{sec:eval_full_framework} 
We compare our scene reconstruction approach against the state of the art for scene completion on all 312 test scenes of the CompleteScanNet based on the IoU metric.
Since our method predicts occupancies while ScanComplete predicts surfaces, we evaluate IoU on both the visible surface (referred as -s) and the entire occupancy (referred as -f).
We use the same trained network as the one in the experiments of section \ref{sec:eval_network_performance}. 
As for ScanComplete \cite{dai2018scancomplete}, we use the three hierarchy levels of training data generated by our pipeline described in section \ref{sec:data_gen}, and trained it from scratch until all levels converged. 
In addition, we include as baseline a framework where ForkNet \cite{wang2019forknet} replaces our back-end and its completions are fused into a separated map (ForkNet+Fusion).
Finally, as ablation, we include our method without CRF regularization. 

All the test scenes are reconstructed with the rendered depths and the skip-frame method as described in our data generation section in order to evaluate both semantic labelling and scene completion.
The scene prediction results are obtained using the proposed incremental pipeline, apart from ScanComplete which are predicted hierarchically from three-level of scans.

Results are illustrated in table \ref{tab:quantitative_full} and figure \ref{fig:qulitative_full_scannet}. 
Our method has the highest IoU score in both visible surface (-s) and completed regions (-f). The naive integration of single predictions has the worst performance both qualitatively and quantitatively, since the predicted geometry often does not match the ground truth geometry. While ScanComplete is able to accurately complete the scene and predict semantic labels, our method obtains a higher accuracy for both reconstruction and semantics. It is noteworthy that our method runs in real-time with a single voxel level input while outperforming all other methods.

\begin{figure}[t]
    \centering
    \includegraphics[width=0.8\linewidth]{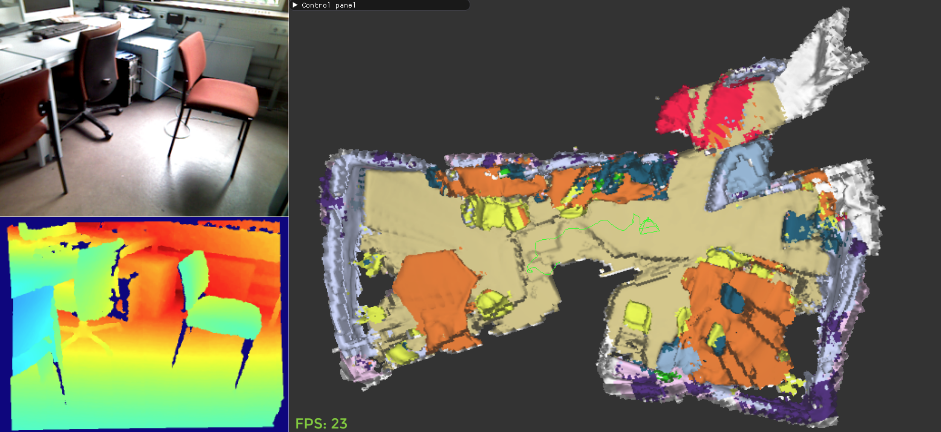}
    \caption{Output of SCFusion on a self-recorded sequence.}
    \label{fig:qua_live}
\end{figure}

\subsection{Run-time analysis} 
We present in table \ref{tab:runtime} a runtime analysis on one sequence (scene0645\_00) of the CompleteScanNet dataset. The sub-map extraction and fusion operations will block the front-end pipeline for a short time due to the asynchronous access to the global map in the back-end pipeline. Since our method assumes a known pose, the time related to tracking is not included in the analysis.

\subsection{Real-world scenario}
Finally, we show qualitative results of SCFusion on a recorded office sequence using a Xtion PRO LIVE sensor with the frame-to-model pose estimation method implemented in InfiniTAM \cite{Kahler2015infinitam}. We use the same trained network in section \ref{sec:eval_full_framework}. The reconstruction results are shown in figure \ref{fig:qua_live}. Our method can successfully obtain semantically complete scenes on the recorded sequence. Notably, the network was still trained on CompleteScanNet, which presents remarkable differences with respect to this test sequence. Also noteworthy, SCFusion runs in real-time also when including the tracking process. Please refer to the supplementary material for the full video.
%


\section{Conclusions}
We have proposed SCFusion, the first framework for incremental real-time semantic scene completion.  
Key ideas for our proposal are the design of a network that processes occupancy maps for efficient 3D semantic scene completion, as well as a specific fusion policy that can integrate completion with a global map incrementally built by a SLAM front-end. 
This fills a gap in the scene completion literature, arguably being the first approach to run semantic scene completion incrementally and in real-time. 
Experimental results show how SCFusion obtains accurate results both in terms of geometry and semantics, en par or even outperforming the state of the art for both offline completion of entire scans and single depth maps. 
\section*{Acknowledgment}%
\label{sec:Acknowledgment}%
This work is supported by the \emph{German Research Foundation} (DFG, project number 407378162).

{\small
\bibliographystyle{ieee}
\bibliography{ms}
}

\newpage

\section{Supplementary material}
\subsection{Network architecture}
\begin{figure*}
    \centering
    \includegraphics[width=0.85\linewidth]{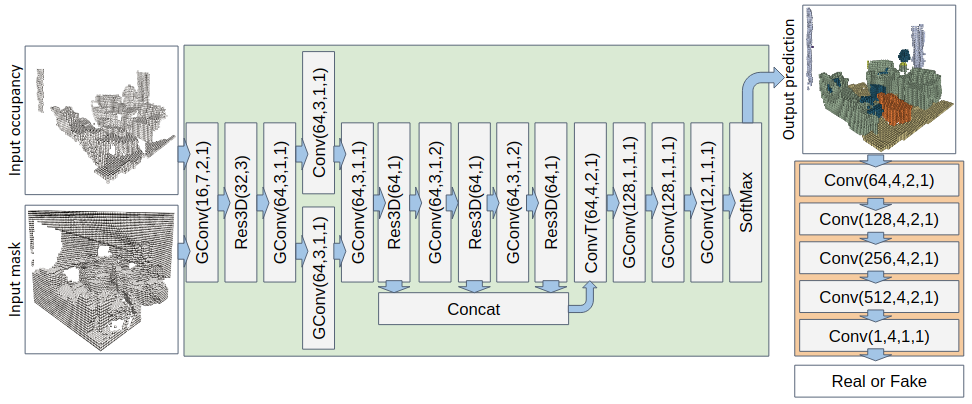}
    \caption{The proposed network architecture. The main network operations is included within the green box, while the operations of discriminator is shown within the orange box. }
    \label{fig:network_architecture}
\end{figure*}

Our network architecture consists of a generator and a discriminator, which are highlighted in a green and an orange box, respectively (Figure \ref{fig:network_architecture}).
The generator consists of gated convolutional layers, denoted as $GConv(c,k,s,d)$, 3D ResNet blocks \cite{hara2018can}, denoted as $Res3D(\cdot,\cdot)$, 3D transpose convolution layers, denoted as $ConvT(c,k,s,d)$ and a SoftMax layer. Where c is the output channels, k is the number of kernels, s is the stride, d is the dilation. 
Note that all the convolutional layers, including inside the $Res3D(\cdot,\cdot)$ are replaced with gated convolutional layers. Apart from the final $GConv(\cdot)$ layer and Softmax, all operations in the generator are followed with an instance normalization layer and a LeakyRelu activation function with a negative slope ratio of 0.2.
As for the discriminator, it consists of five convolutional layers, denoted as $Conv(c,k,s,d)$, all followed by a spectral normalization operation and a leakyReLu operation, except the final convolutional layer.

\subsection{Ablation study}
\subsubsection{The effect of map regularization}
The effect of our map regularization method is illustrated in figure \ref{fig:abla_crf}. We highlighted the regions which failed to predict labels that are corrected by the regularization.

\subsubsection{Different network designs}
\begin{figure*}[h]
\centering
\begin{minipage}[b]{.19\textwidth}
  \centering
  Input
  \includegraphics[width=\textwidth, height=0.1\textheight]{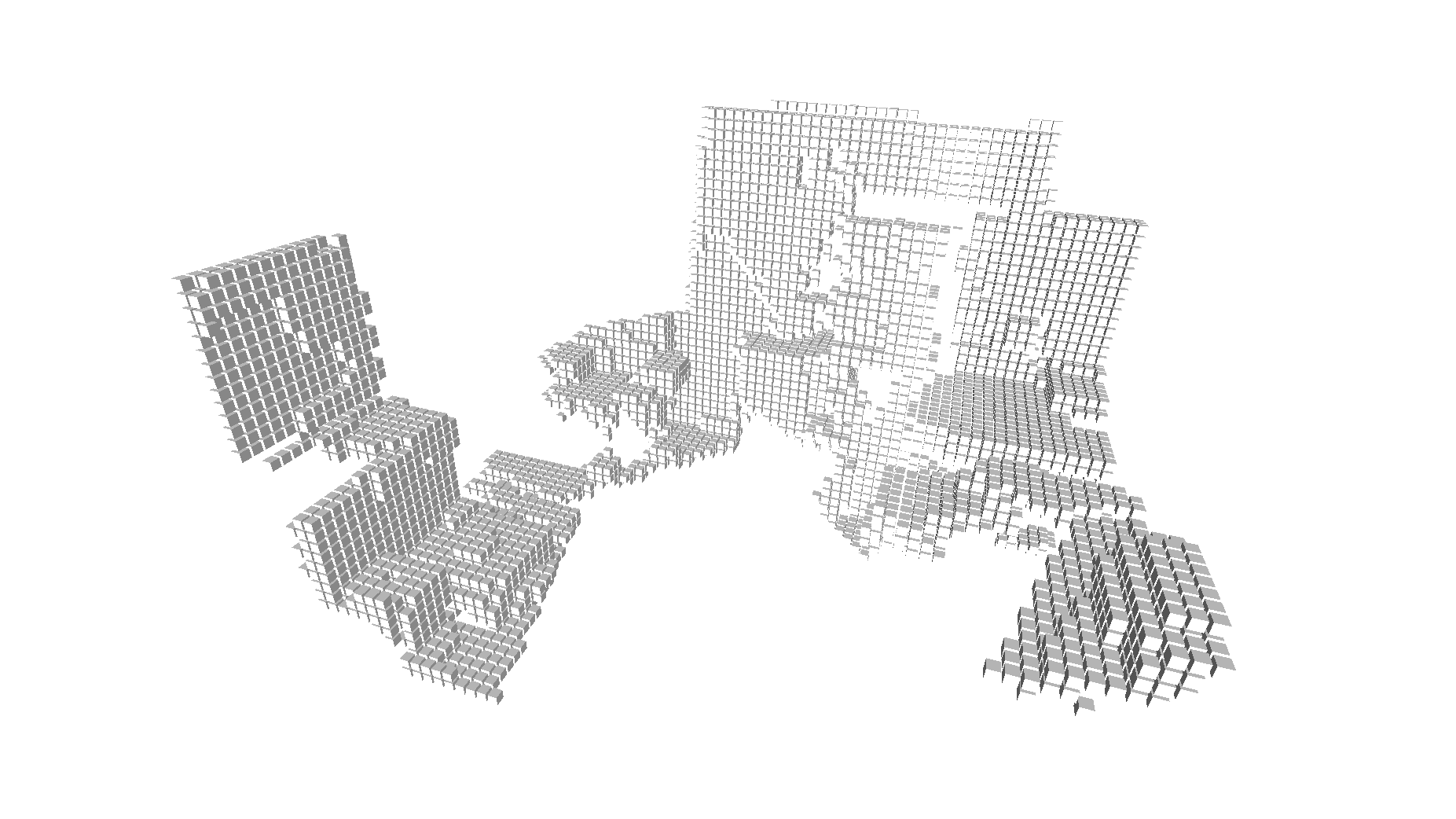}
  \includegraphics[width=\textwidth, height=0.1\textheight]{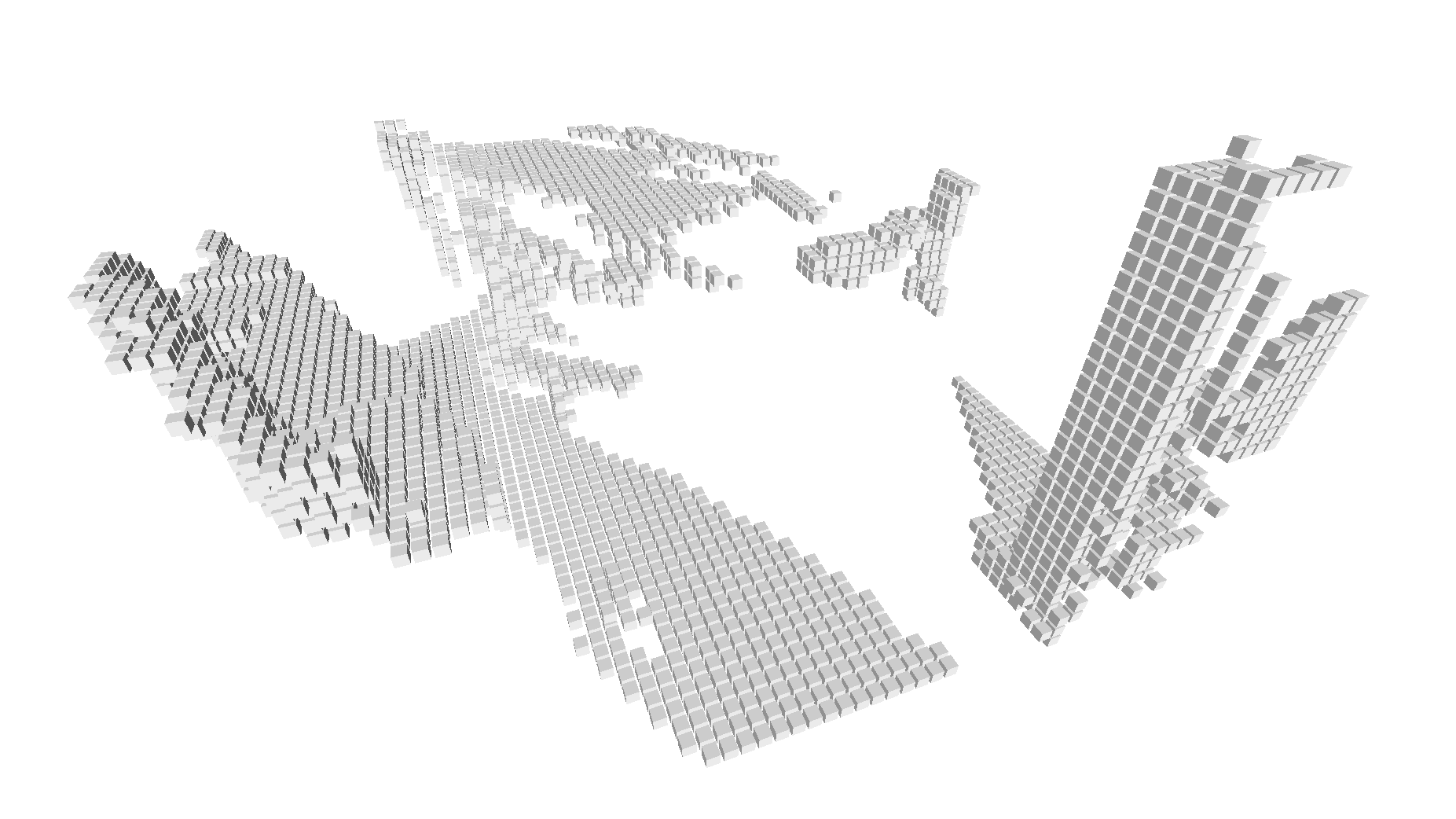}
  \includegraphics[width=\textwidth, height=0.1\textheight]{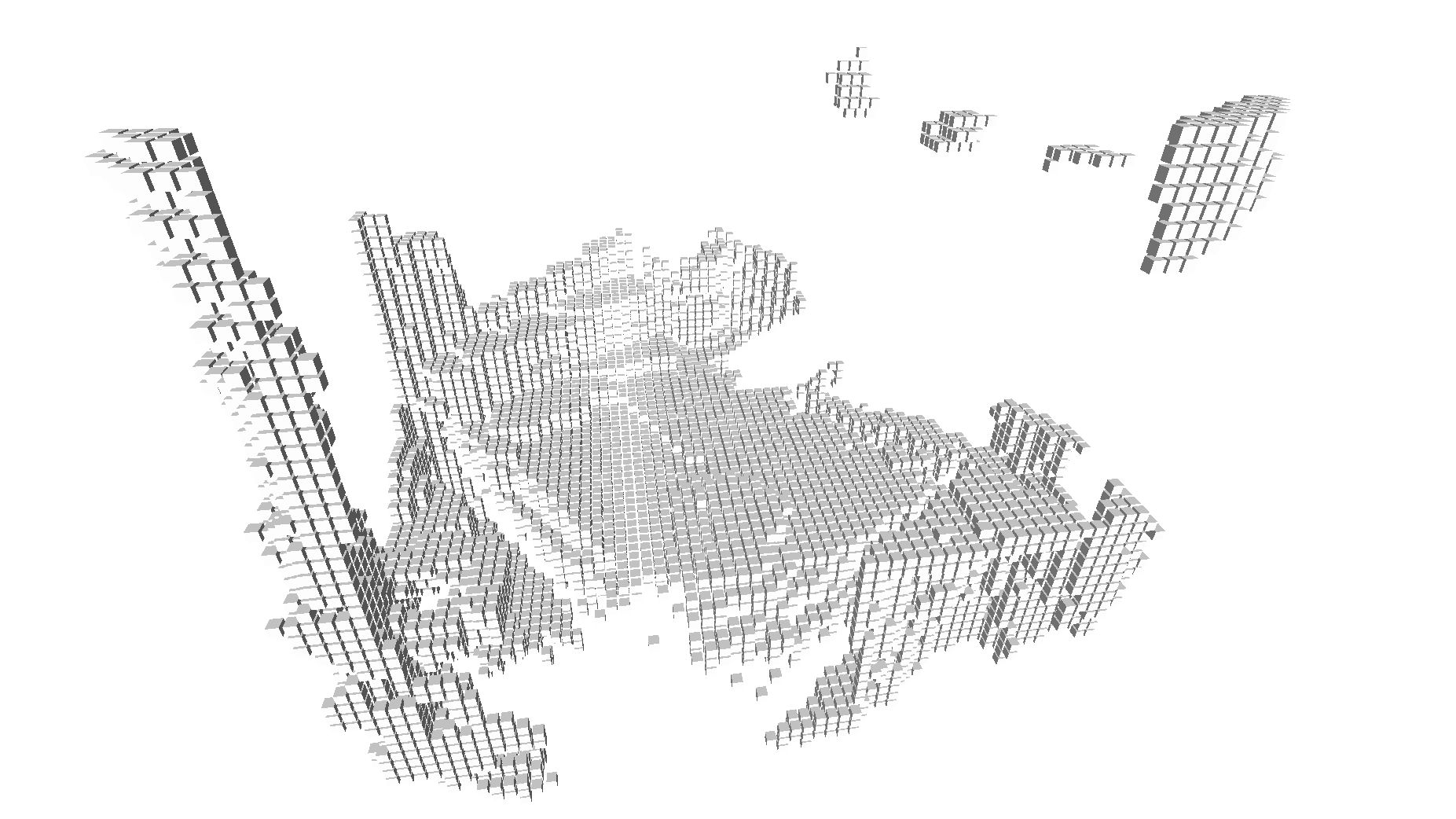}
\end{minipage}%
\begin{minipage}[b]{.19\textwidth}
  \centering
  base
  \includegraphics[width=\textwidth, height=0.1\textheight]{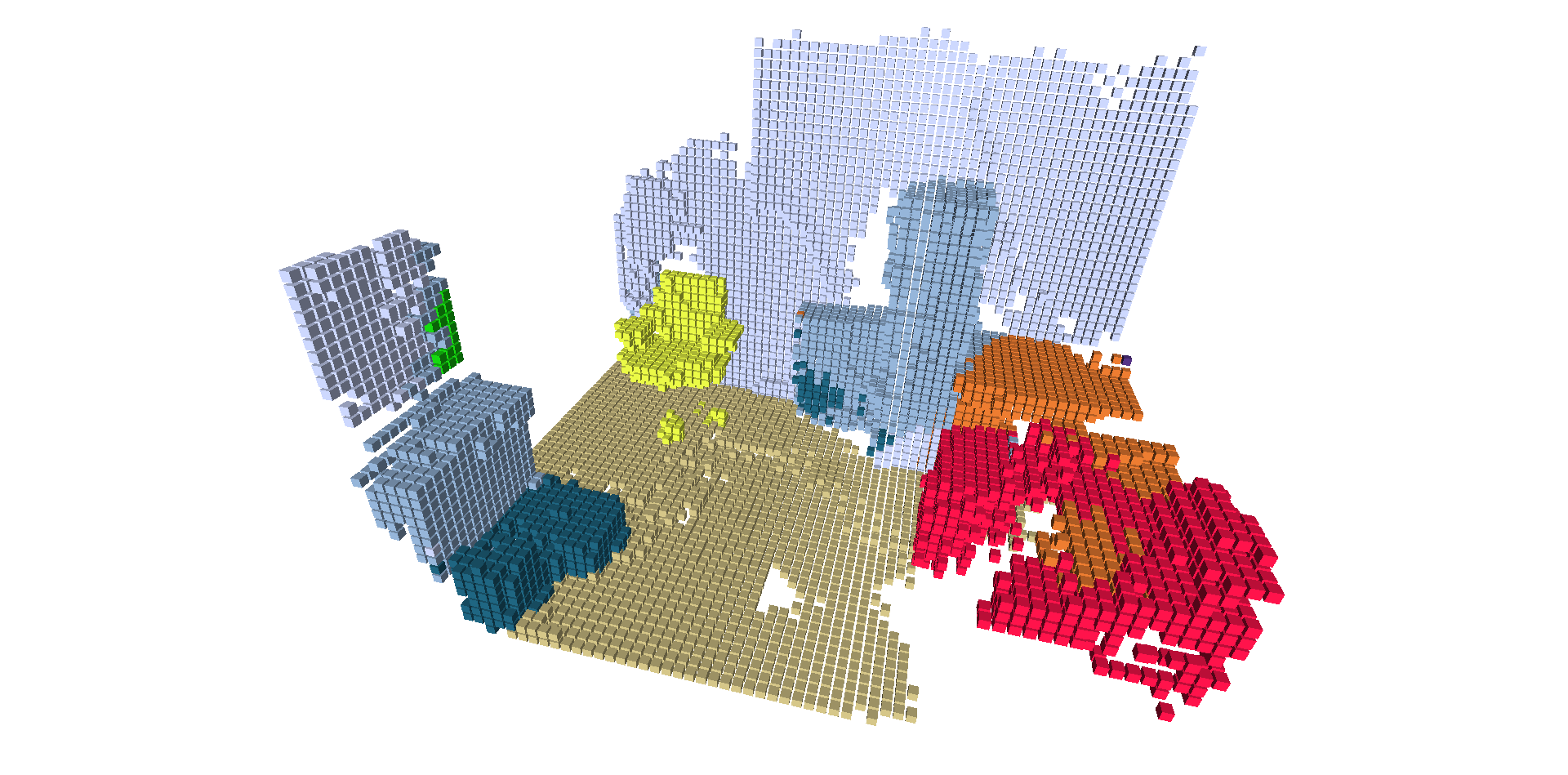}
  \includegraphics[width=\textwidth, height=0.1\textheight]{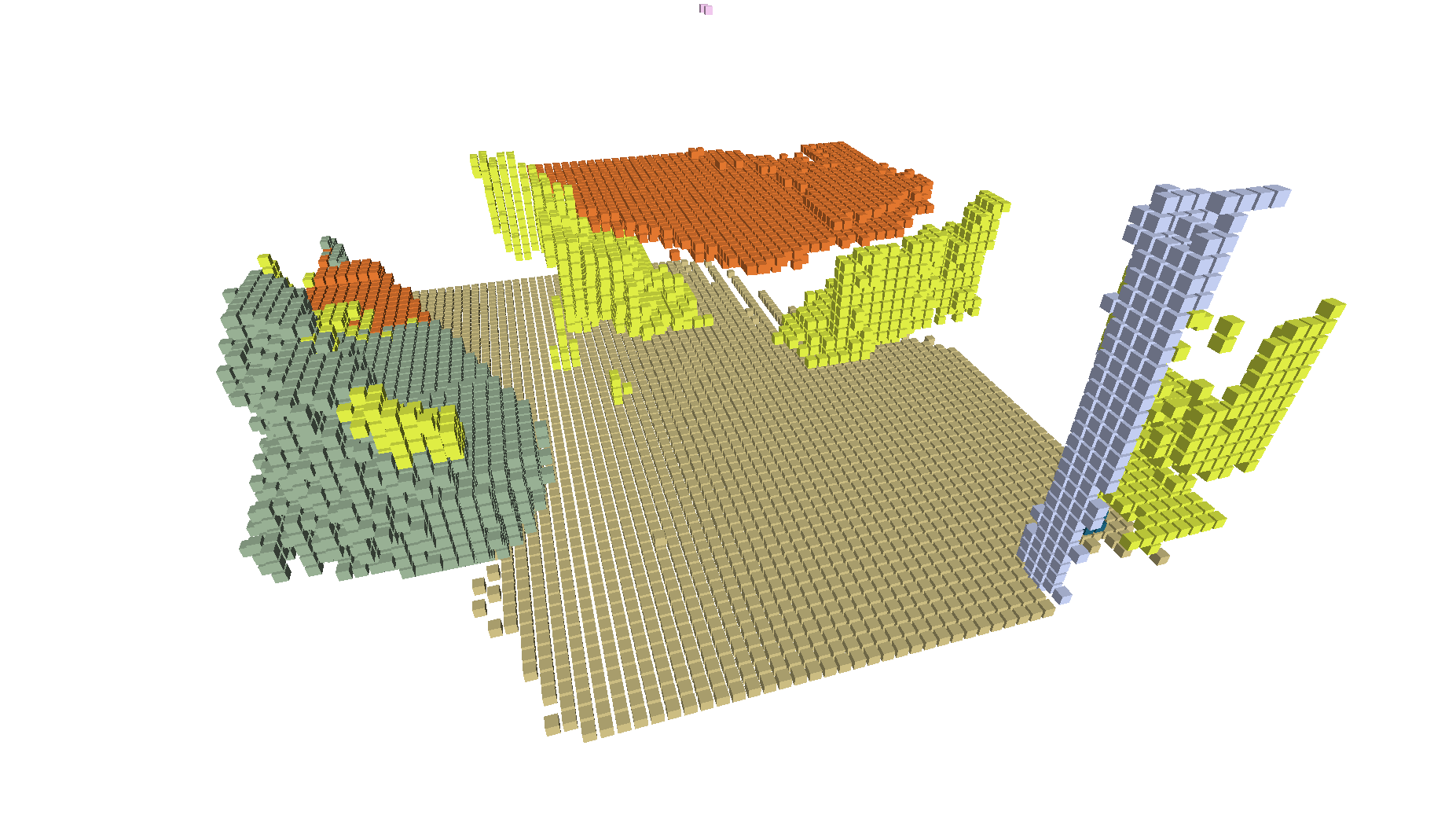}
  \includegraphics[width=\textwidth, height=0.1\textheight]{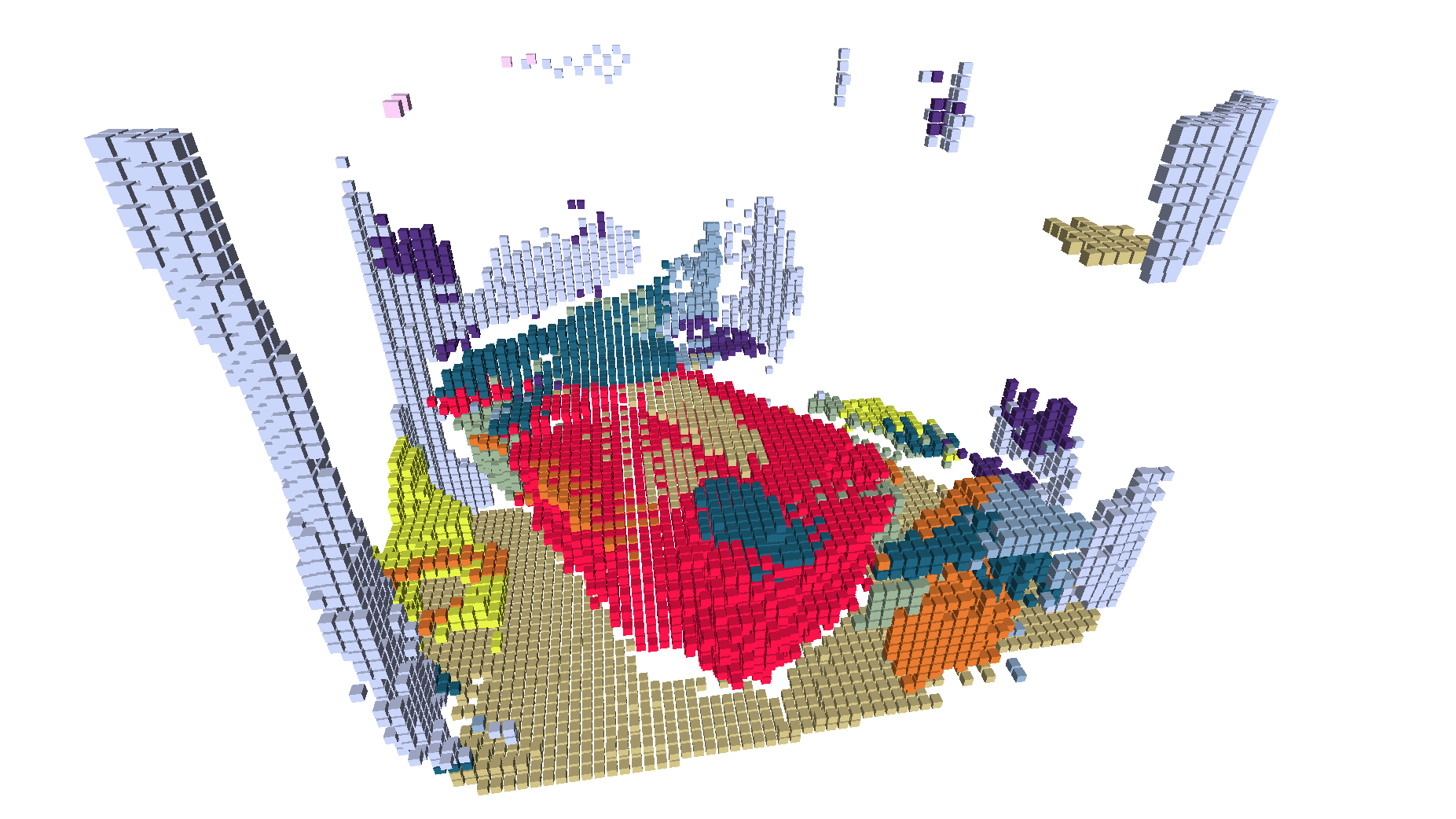}
\end{minipage}%
\begin{minipage}[b]{.19\textwidth}
  \centering
  base+G \cite{wang2019forknet}
  \includegraphics[width=\textwidth, height=0.1\textheight]{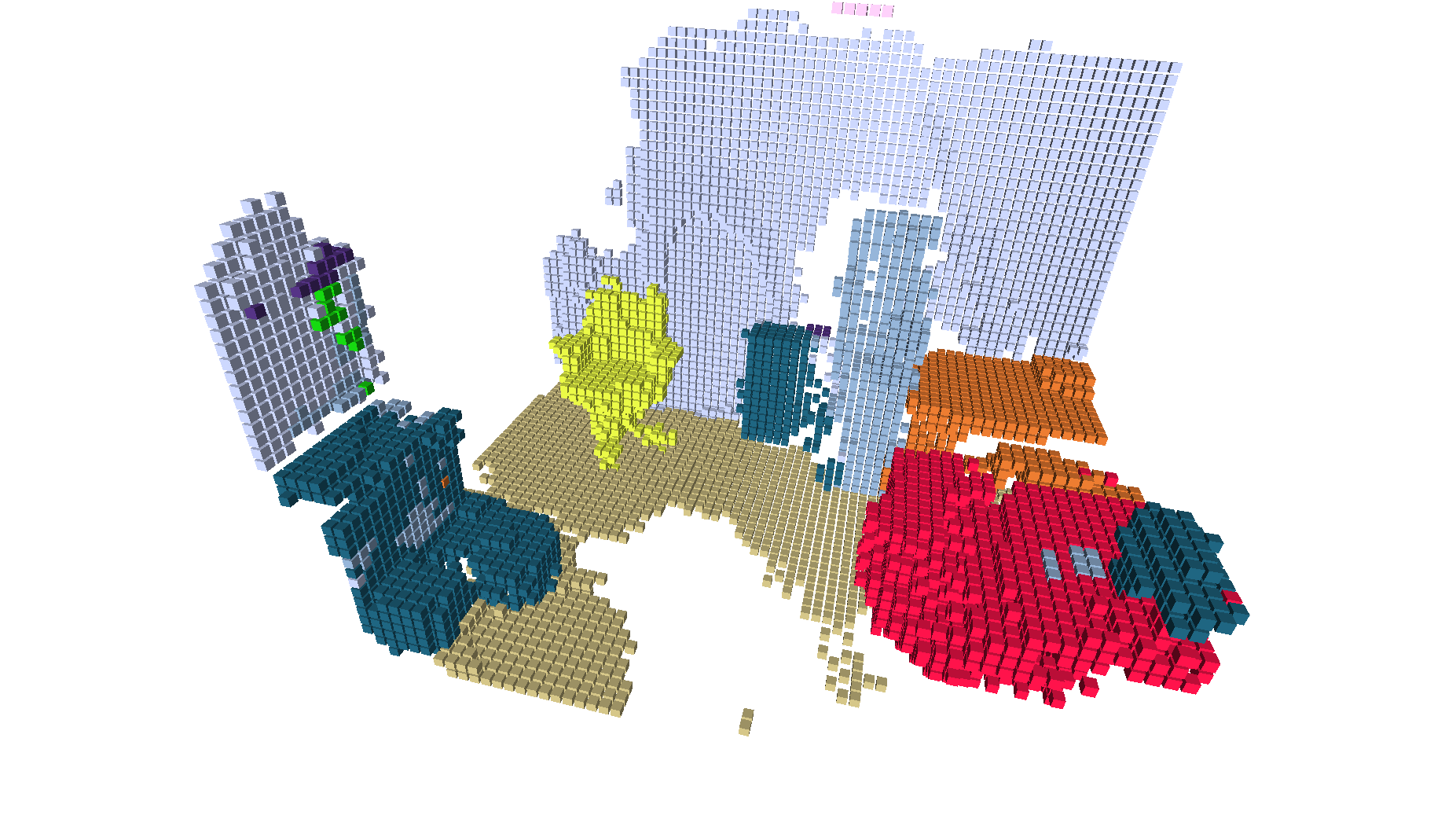}
  \includegraphics[width=\textwidth, height=0.1\textheight]{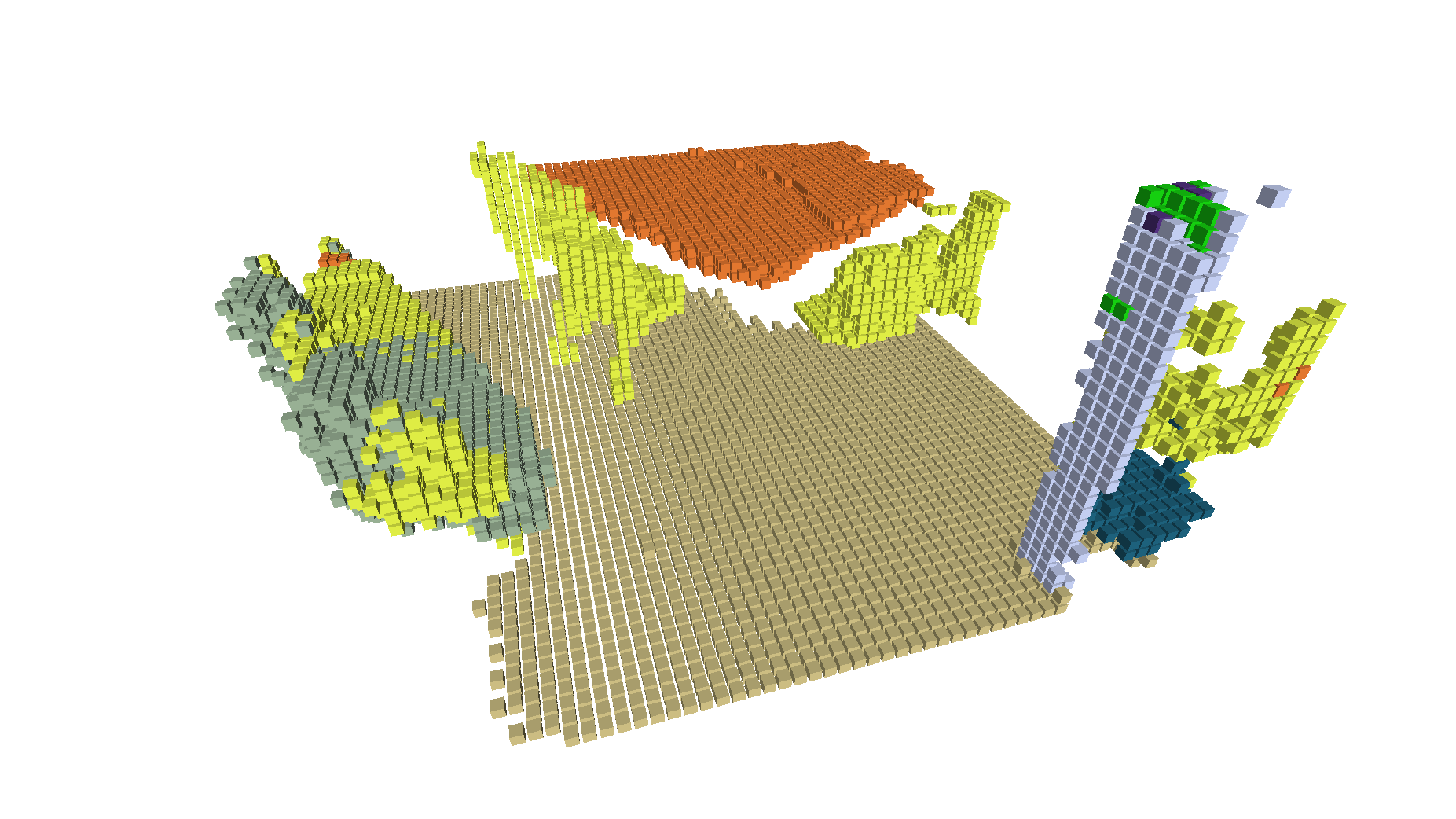}
  \includegraphics[width=\textwidth, height=0.1\textheight]{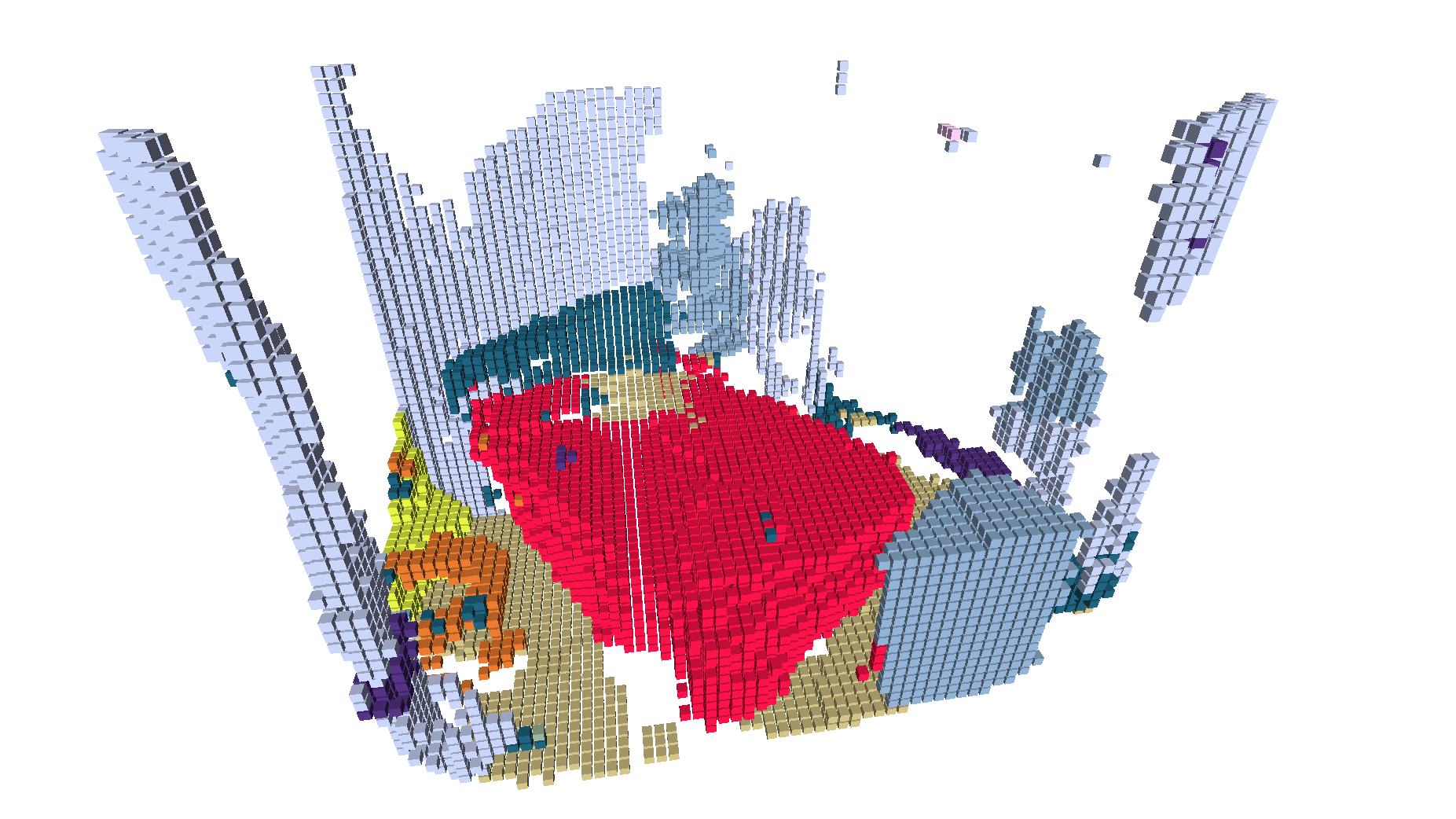}
\end{minipage}%
\begin{minipage}[b]{.19\textwidth}
  \centering
  Ours
  \includegraphics[width=\textwidth, height=0.1\textheight]{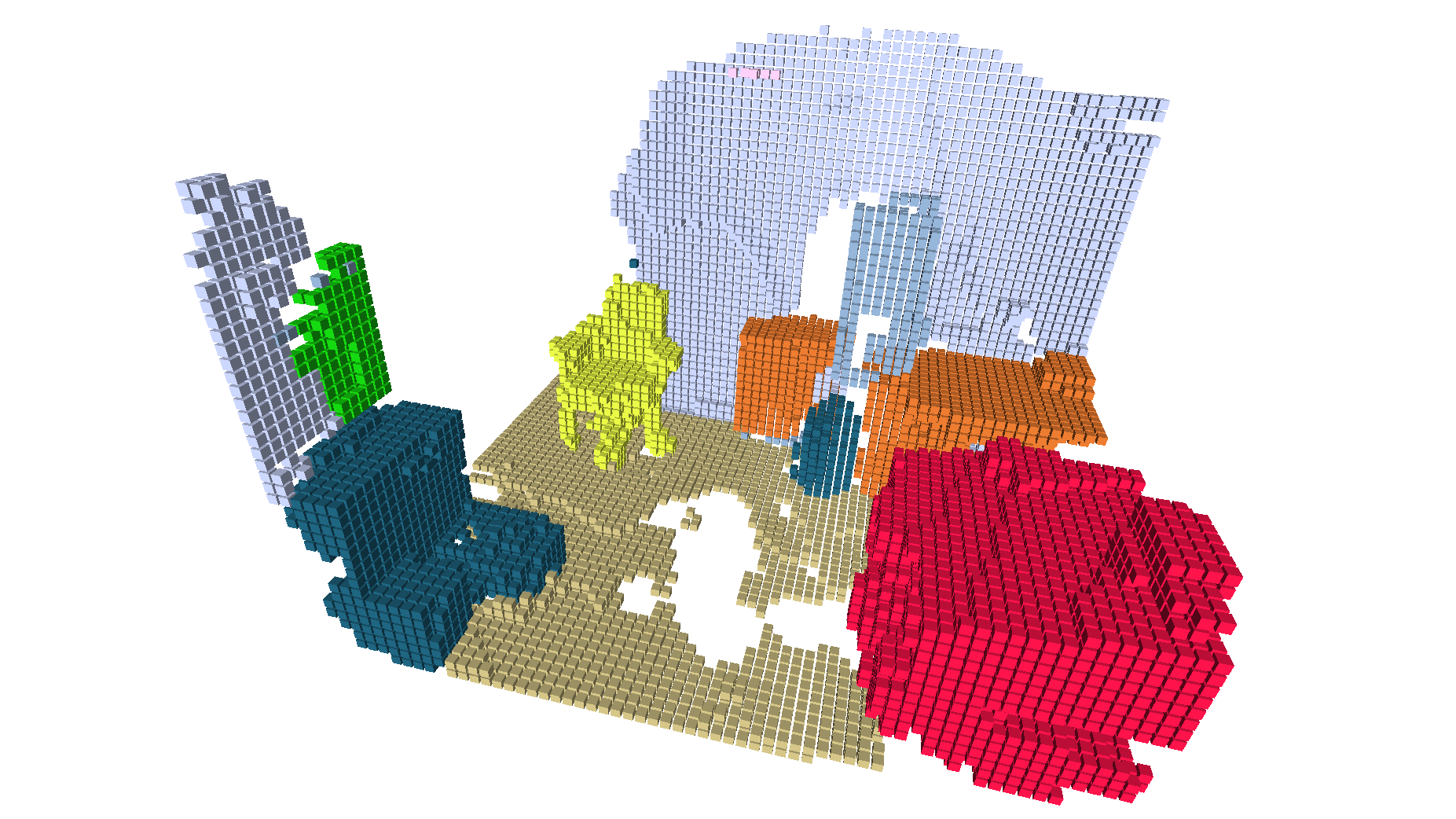}
  \includegraphics[width=\textwidth, height=0.1\textheight]{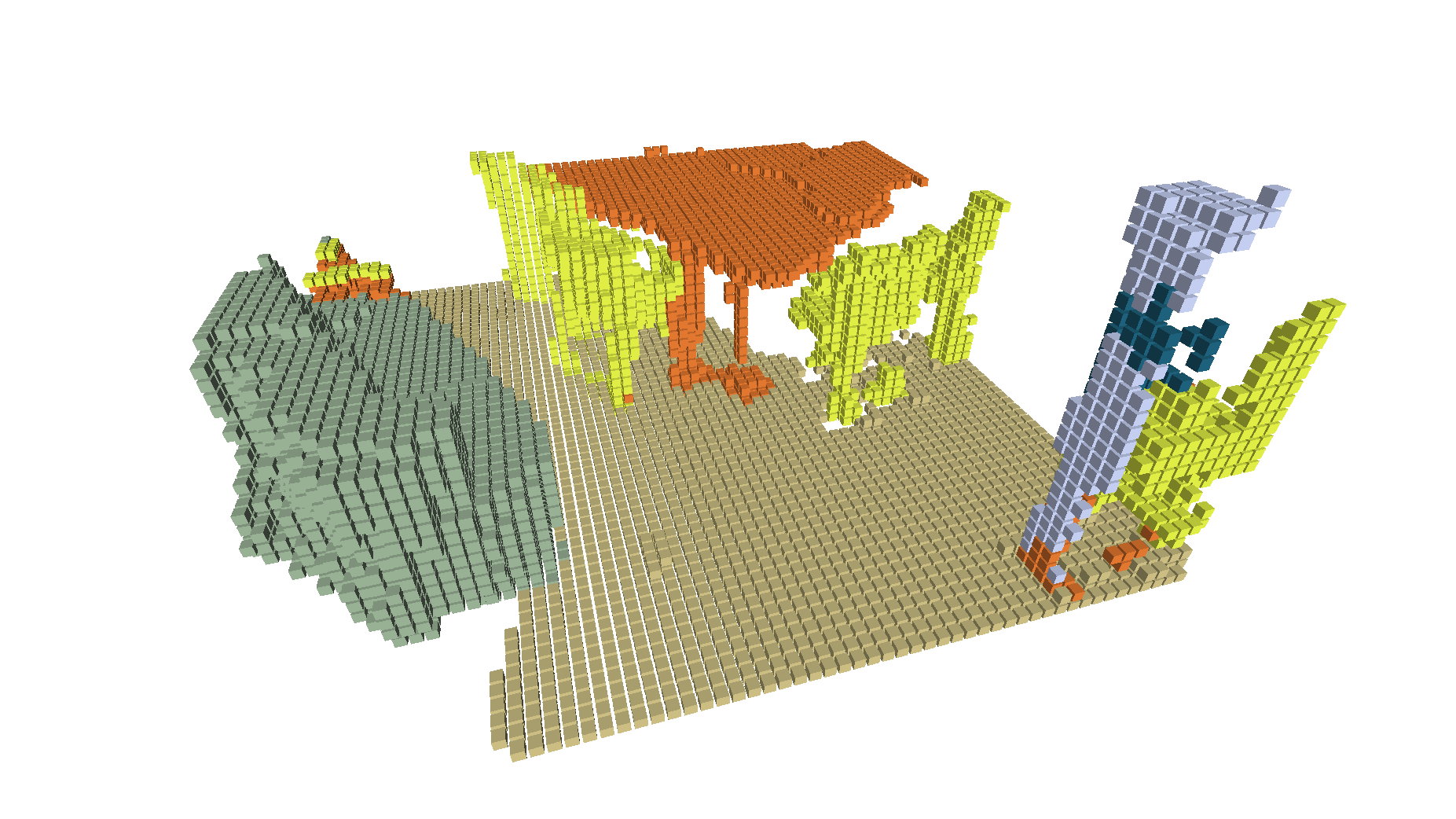}
  \includegraphics[width=\textwidth, height=0.1\textheight]{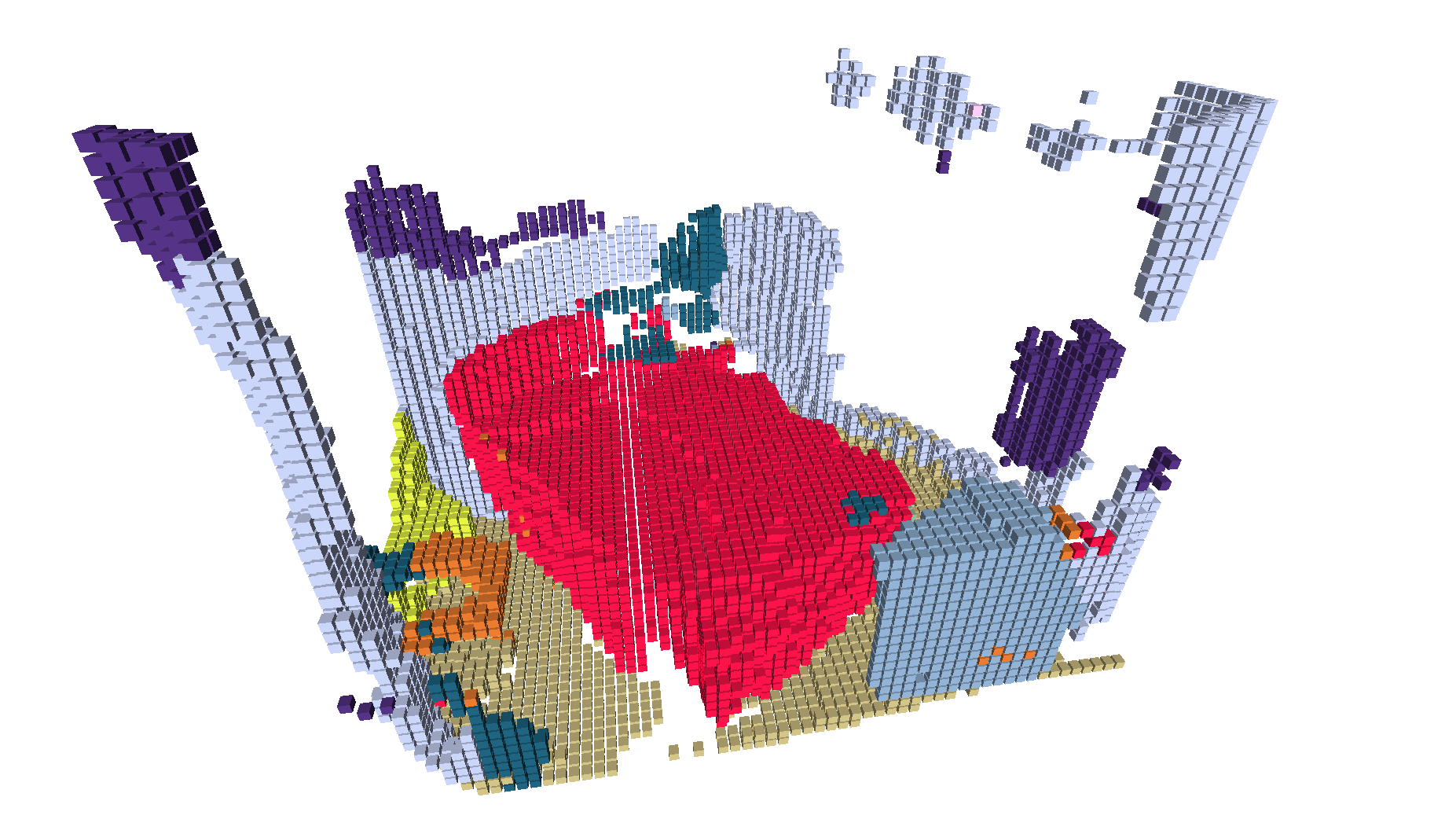}
\end{minipage}%
\begin{minipage}[b]{.19\textwidth}
  \centering
  Ground Truth
  \includegraphics[width=\textwidth, height=0.1\textheight]{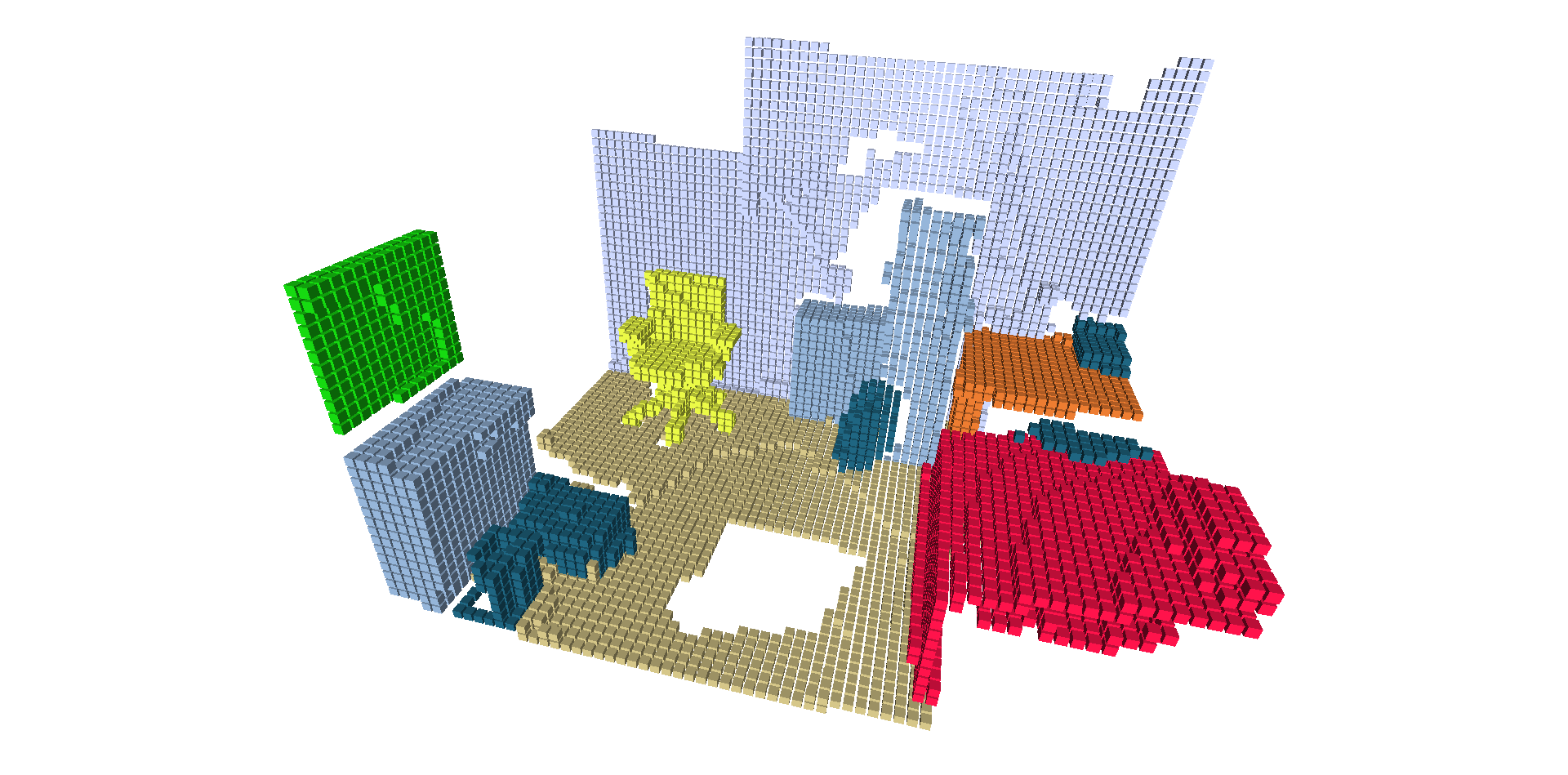}
  \includegraphics[width=\textwidth, height=0.1\textheight]{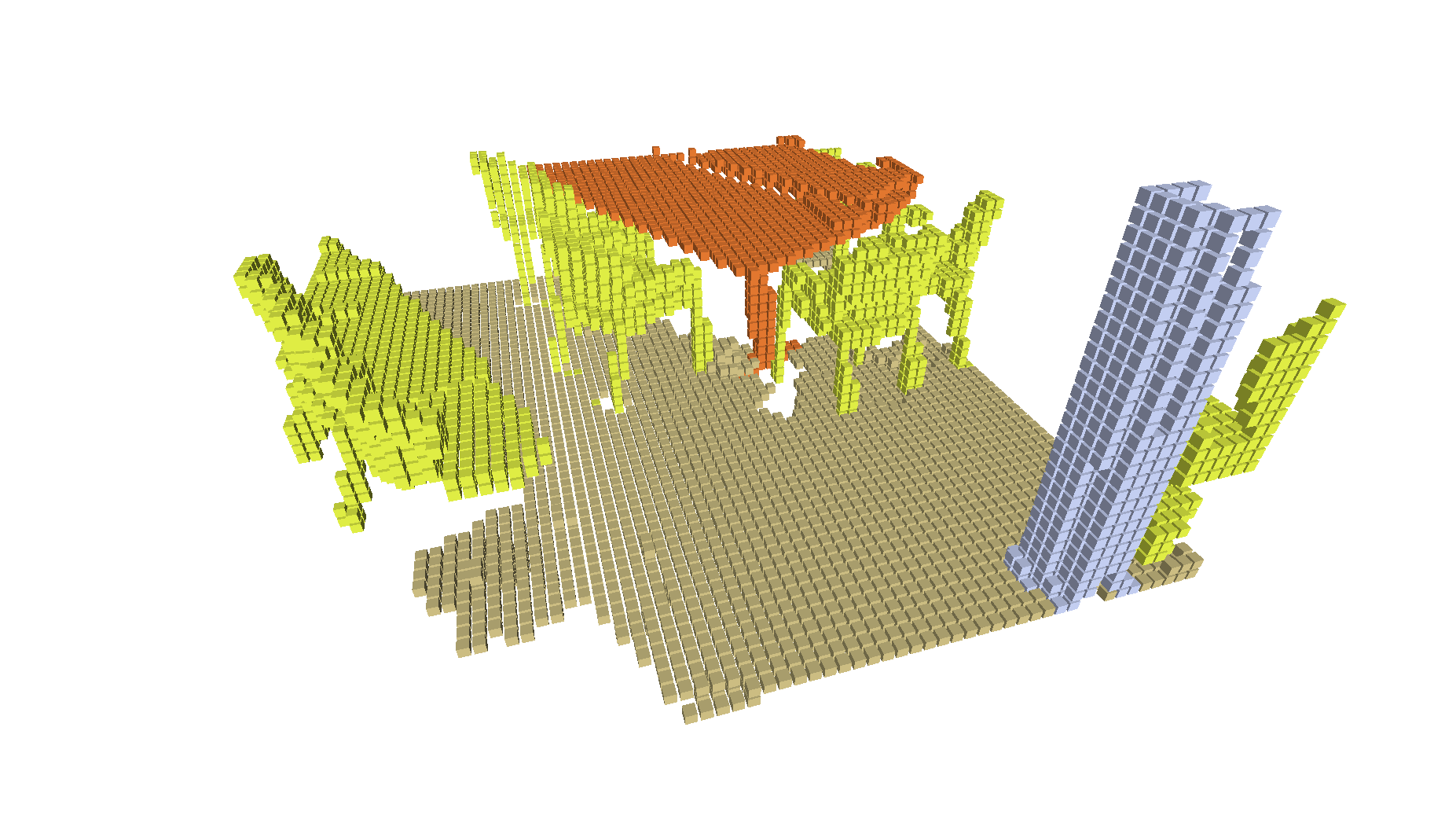}
  \includegraphics[width=\textwidth, height=0.1\textheight]{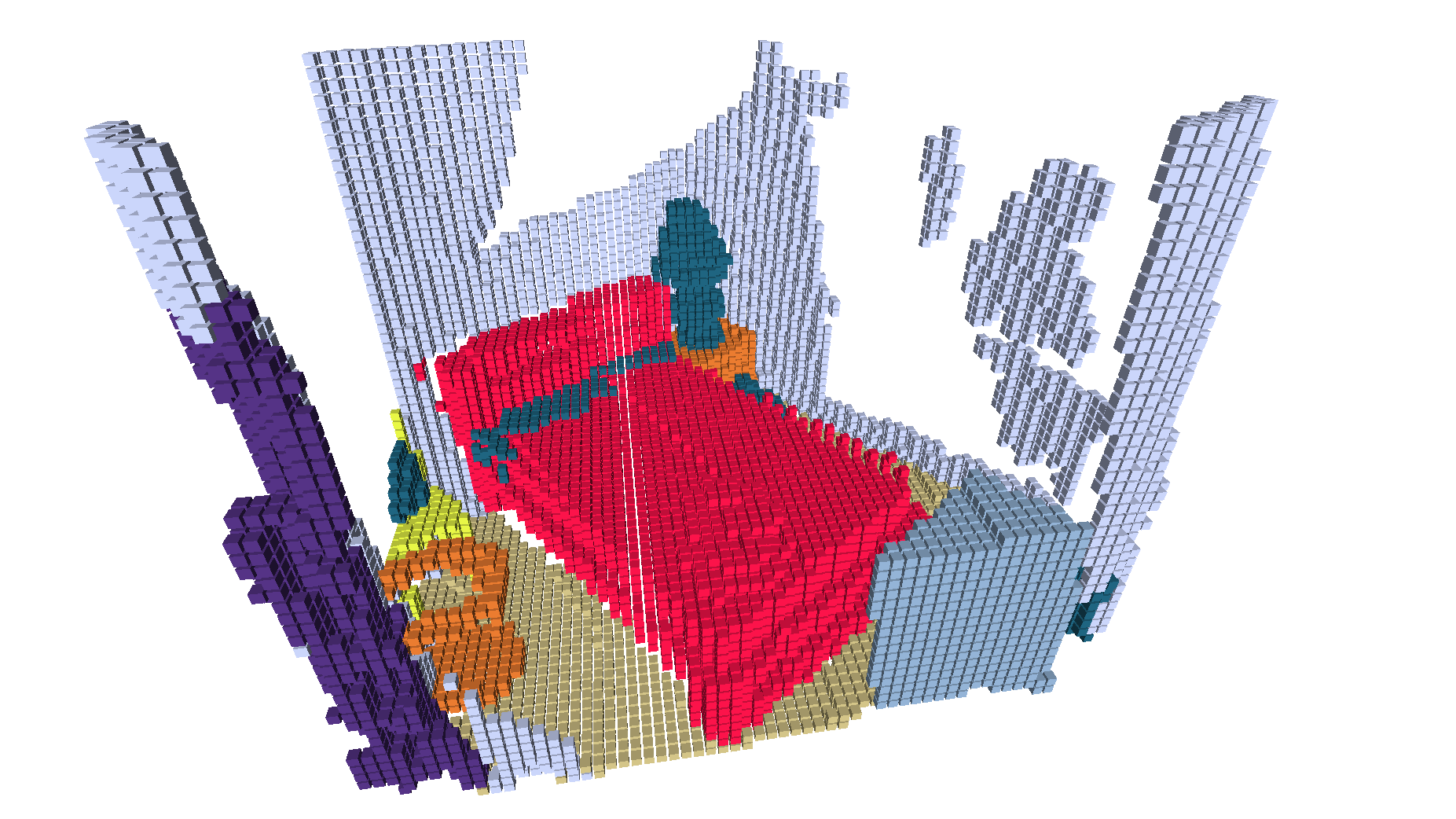}
\end{minipage}%
\\
\begin{minipage}[b]{\textwidth}
\centering
\includegraphics[width=1.0\textwidth]{images/SunCG11Labels2.png}
\end{minipage}
\caption{
%
The use of gated convolutional layers with an input mask increase the completion ability significantly. Plus the constraint provided from the discriminator, the network is able to predict more complete scenes more precisely.
}
\label{fig:abla_mask}
\end{figure*}

We compare our final design with the two other setups with the metrics we used to evaluate our network in the main paper. 
First, as a baseline, the semantic scene prediction branch from ForkNet\cite{wang2019forknet} with the replacement of the input format from the inverted truncated distance function to the occupancy probability (denote as \textbf{base}). 
Second, we replace all the convolutional layers with the gated convolutional layer and add a mask, as an additional input, which indicates the regions where a completion process may be needed (denote as \textbf{base+G}). The result is shown in table \ref{tab:abla_mask} and figure \ref{fig:abla_mask}. It can be seen that the use of gated convolutional layers and a mask improve the overall network performance in a margin. The effect of discriminator slightly improves the numerical result, while dramatically increases the prediction in fine details. 

\subsection{Limitations}
\begin{figure}[t]
    \centering
\includegraphics[width=1\linewidth]{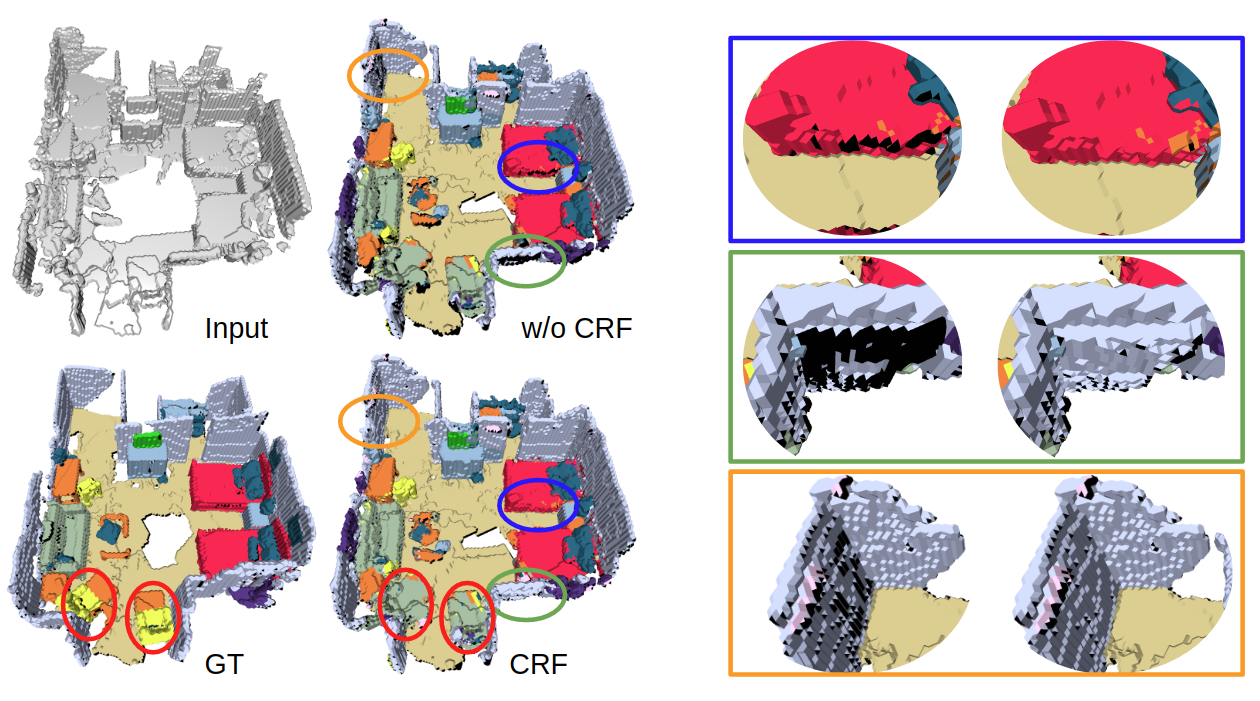}
    \caption{The effect of our regularization method and some failure cases of our SCFusion. The blue, green, and orange areas indicate the regions with noticeable improvement by our regularization method. The red circles show the failure case of misleading geometry of connecting chair and table, which result in wrongly label prediction. }
    \label{fig:abla_crf}
\end{figure}

Our method uses only geometry input which limits its ability to distinguish the objects with ambiguous geometry shape and the objects that only differ from colors. 
For instance, a small table is adjacent to a chair may be classified as a sofa (See figure \ref{fig:abla_crf} red circles). 

\begin{table*}[t]
\resizebox{\textwidth}{!}{%
    \begin{tabular}{c|c|c|c|c|c|c|c|c|c|c|c|c|c}
    Metric & Method & Ceiling & Floor & Wall & Window & Chair & Bed & Sofa & Table & TV & Furni & Object & Mean \\
    \hline\hline
\multirow{3}{*}{IoU}&base&          0.193 &          0.548 &          0.372 &          0.057 &          0.300 &          0.211 &          0.207 &          0.323 &          0.295 &          0.207 &          0.121 &          0.258\\ \cline{2-14} 
                    &base+G& \textbf{0.226} & \textbf{0.564} & \textbf{0.392} &          0.068 & \textbf{0.337} & \textbf{0.290} & \textbf{0.295} & \textbf{0.334} &          0.181 &          0.207 & \textbf{0.152} &          0.277\\ \cline{2-14} 
                    &Ours&          0.197 &          0.541 &          0.379 & \textbf{0.108} &          0.310 &          0.194 &          0.266 &          0.322 & \textbf{0.659} & \textbf{0.219} &          0.148 & \textbf{0.304}\\
    \hline\hline
\multirow{2}{*}{Precision}&base& \textbf{0.561} &          0.691 & \textbf{0.654} &          0.189 &          0.501 &          0.398 &          0.319 & \textbf{0.582} & 0         .367 &          0.381 & \textbf{0.321} &          0.451\\ \cline{2-14} 
                          &base+G&          0.535 & \textbf{0.757} &          0.632 &          0.179 & \textbf{0.551} & \textbf{0.495} & \textbf{0.440} &          0.579 &          0.238 & \textbf{0.424} &          0.313 & \textbf{0.468}\\ \cline{2-14} 
                          &Ours&          0.432 &          0.680 &          0.591 & \textbf{0.248} &          0.528 &          0.304 &          0.378 &          0.505 & \textbf{0.771} &          0.410 &          0.300 & \textbf{0.468}\\ 
  \hline\hline
\multirow{2}{*}{Recall}&base&          0.248 &          0.704 & \textbf{0.536} &          0.082 &          0.447 &          0.260 &          0.251 &          0.471 &          0.310 &          0.314 &          0.202 &          0.348\\ \cline{2-14} 
                       &base+G&          0.333 &          0.687 &          0.511 & \textbf{0.397} & \textbf{0.484} &          0.613 &          0.684 &          0.459 & \textbf{0.850} &          0.315 & \textbf{0.257} &          0.508\\ \cline{2-14} 
                       &Ours& \textbf{0.343} & \textbf{0.722} &          0.517 &          0.391 &          0.450 & \textbf{0.667} & \textbf{0.714} & \textbf{0.497} &          0.842 & \textbf{0.347} &          0.253 & \textbf{0.522}\\ 
   \hline\hline
    \end{tabular}
}
\caption{Ablation study of the network design. We compare our method (\textbf{ours}) against the baseline ForkNet \cite{wang2019forknet} semantic branch (\textbf{base}) and the baseline plus an input mask with gated convolutions (\textbf{base+G}).  }
    \label{tab:abla_mask}
\end{table*}

\end{document}